\documentclass{article}

\usepackage{microtype}
\usepackage{graphicx}
\usepackage{subcaption} %
\usepackage{booktabs} %

\usepackage{hyperref}

\usepackage[accepted]{icml2021}

\usepackage{amsmath,amsfonts,bm}

\def\eqref#1{equation~\ref{#1}}

\def\1{\bm{1}}

\def\vp{{\bm{p}}}

\def\vx{{\bm{x}}}
\def\vy{{\bm{y}}}

\DeclareMathAlphabet{\mathsfit}{\encodingdefault}{\sfdefault}{m}{sl}
\SetMathAlphabet{\mathsfit}{bold}{\encodingdefault}{\sfdefault}{bx}{n}

\newcommand{\E}{\mathbb{E}}

\usepackage{hyperref}       %
\usepackage{url}            %

\usepackage{amsmath}
\usepackage{amssymb}
\usepackage{amsfonts}       %
\usepackage{amsthm}
\usepackage{nicefrac}       %
\usepackage{microtype}      %

\usepackage[utf8]{inputenc} %
\usepackage[T1]{fontenc}    %
\usepackage{bbold}
\usepackage{stmaryrd}
\usepackage{bm}

\usepackage{float}
\usepackage[inline]{enumitem}
\usepackage{caption} 
\usepackage{wrapfig}
\usepackage{booktabs}
\usepackage{makecell}
\usepackage[table]{colortbl}
\usepackage{xspace}
\usepackage{multirow}

\usepackage{subfiles}

\icmltitlerunning{Evaluating Robustness of Predictive Uncertainty Estimation: Are Dirichlet-based Models Reliable?}

\DeclareMathOperator{\DNormal}{\mathcal{N}}

\newcommand\idata{i\xspace}
\newcommand\dataix{^{(\idata)}}
\newcommand\iclass{c\xspace}
\newcommand\nclass{C\xspace}
\newcommand\PriorNet{PriorNet\xspace}
\newcommand\RevPriorNet{RevPriorNet\xspace}
\newcommand\EvNet{EvNet\xspace}
\newcommand\DDNet{DDNet\xspace}
\newcommand\PostNet{PostNet\xspace}

\begin{document}

\twocolumn[
\icmltitle{Evaluating Robustness of Predictive Uncertainty Estimation: \\ Are Dirichlet-based Models Reliable?}

\icmlsetsymbol{equal}{*}

\begin{icmlauthorlist}
\icmlauthor{Anna-Kathrin Kopetzki}{equal,tum}
\icmlauthor{Bertrand Charpentier}{equal,tum}
\icmlauthor{Daniel Z\"ugner}{tum}
\icmlauthor{Sandhya Giri}{tum}
\icmlauthor{Stephan G\"unnemann}{tum}
\end{icmlauthorlist}

  \hypersetup{ %
    pdftitle={Evaluating Robustness of Predictive Uncertainty Estimation: Are Dirichlet-based Models Reliable?},
    pdfauthor={Anna-Kathrin Kopetzki, Bertrand Charpentier, Daniel Z\"ugner, Sandhya Giri, Stephan G\"unnemann },
    pdfsubject={Proceedings of the International Conference on Machine Learning 2021},
    pdfkeywords={Uncertainty Estimation, Robustness},
}

\icmlaffiliation{tum}{Technical University of Munich, Germany; Department of Informatics}

\icmlcorrespondingauthor{Anna-Kathrin Kopetzki}{kopetzki@in.tum.de}
\icmlcorrespondingauthor{Bertrand Charpentier}{charpent@in.tum.de}

\icmlkeywords{Machine Learning, ICML}

\vskip 0.3in
]

\printAffiliationsAndNotice{\icmlEqualContribution} %

\begin{abstract}
Dirichlet-based uncertainty (DBU) models are a recent and promising class of uncertainty-aware models. DBU models predict the parameters of a Dirichlet distribution to provide fast, high-quality uncertainty estimates alongside with class predictions. 
In this work, we present the first large-scale, in-depth study of the robustness of DBU models under adversarial attacks. Our results suggest that uncertainty estimates of DBU models are not robust w.r.t.\ three important tasks:
\textbf{(1)} indicating correctly and wrongly classified samples;
\textbf{ (2)} detecting adversarial examples; and 
\textbf{(3)} distinguishing between in-distribution (ID) and out-of-distribution (OOD) data.
Additionally, we explore the first approaches to make DBU models more robust. While adversarial training has a minor effect, our median smoothing based approach significantly increases robustness of DBU models. 

\end{abstract}

\section{Introduction}
\label{sec:intro}

Neural networks achieve high predictive accuracy in many tasks, but they are known to have two substantial weaknesses: First, neural networks are not robust against adversarial perturbations, i.e., semantically meaningless input changes that lead to wrong predictions \citep{szegedy2014, goodfellow2014}. 
Second, standard neural networks are unable to identify samples that are different from the ones they were trained on and tend to make over-confident predictions at test time \citep{ensemble_simple}. These weaknesses make them impracticable in sensitive domains like financial, autonomous driving or medical areas which require trust in predictions.

To increase trust in neural networks, models that provide predictions along with the corresponding uncertainty have been proposed. 
An important, quickly growing family of models is the Dirichlet-based uncertainty (DBU) family \citep{malini2018, malinin2019, sensoy2018, malinin2019ensemble, charpentier2020, graph_uncertainty, max_gap_id_ood, multifaceted_uncertainty, uncertainty-generative-classifier}. 
Contrary to other approaches such as Bayesian Neural Networks \citep{blundell2015, osawa2019, wesley2019}, drop out \citep{drop_out} or ensembles \citep{ensemble_simple}, DBU models provide efficient uncertainty estimates at test time in a single forward pass by directly predicting the parameters of a Dirichlet distribution over categorical probability distributions.
\begin{figure}[t]
\centering
\includegraphics[width=0.36\textwidth]{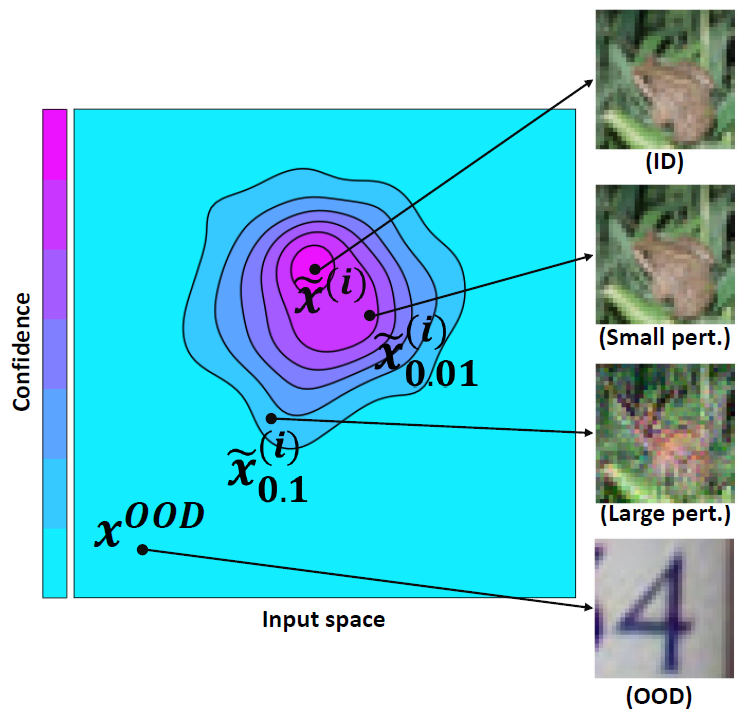}
\caption{Visualization of the desired uncertainty estimates. 
}
\label{fig:uncertainty_attack_diagram}
\end{figure}
DBU models have the advantage that they provide both, aleatoric uncertainty estimates resulting from irreducible uncertainty (e.g. class overlap or noise) and epistemic uncertainty estimates resulting from the lack of knowledge about unseen data (e.g. an unknown object is presented to the model). Both uncertainty types can be quantified from Dirichlet distributions using different uncertainty measures such as differential entropy, mutual information, or pseudo-counts. These uncertainty measures show outstanding performance in, e.g., the detection of OOD samples and thus are superior to softmax based confidence \citep{malini2018, malinin2019, charpentier2020}.

Neural networks from the families outlined above are expected to \emph{know what they don't know}, i.e. they are supposed to notice when they are unsure about a prediction. 
This raises questions with regards to adversarial examples: should uncertainty estimation methods \emph{detect} these corrupted samples by indicating a high uncertainty on them and abstain from making a prediction? Or should uncertainty estimation be \emph{robust} to adversarial examples and assign the correct label even under perturbations? We argue that being robust to adversarial perturbations is the best option (see Figure~\ref{fig:uncertainty_attack_diagram}) for two reasons. First, in image classification a human is usually not able to observe any difference between an adversarial example and an unperturbed image. Second, the size of the perturbation corresponding to a good adversarial example is typically small w.r.t.\ the $L_p$-norm and thus assumed to be semantically meaningless. 
Importantly, robustness should not only be required for the class predictions, but also for the uncertainty estimates. This means that DBU models should be able to distinguish robustly between ID and OOD data even if those are perturbed. 

In this work, we focus on DBU models and analyze their robustness capacity w.r.t. class predictions as well as uncertainty predictions. In doing so, we go beyond simple softmax output confidence by investigating advanced uncertainty measures such as differential entropy.
Specifically, we study the following questions: 
\begin{enumerate}
    \item \emph{Is low uncertainty a reliable indicator of correct predictions?}
    \item \emph{Can we use uncertainty estimates to detect label attacks on the class prediction?}
    \item \emph{Are uncertainty estimates such as differential entropy a robust feature for OOD detection?}
\end{enumerate}

In addressing these questions we place particular focus on adversarial perturbations of the input to evaluate the \emph{worst case} performance of the models on increasing complex data sets and attacks.
We evaluate robustness of DBU models w.r.t. to these three questions by comparing their performance on unperturbed and perturbed inputs. Perturbed inputs are obtained by computing \emph{label attacks} and \emph{uncertainty attacks}, which are a new type of attacks we propose.  While label attacks aim at changing the class prediction, uncertainty attacks aim at changing the uncertainty estimate such that ID data is marked as OOD data and vice versa.
In total, we performed more than $138,800$ attack settings to explore the robustness landscape of DBU models. Those settings cover different data sets, attack types, attack losses, attack radii, DBU model types and initialisation seeds.
Finally, we propose and evaluate median smoothing and adversarial training based on label attacks and uncertainty attacks to make DBU models more robust. Our median smoothing approach provides certificates on epistemic uncertainty measures such as differential entropy and allows to certify uncertainty estimation.  The code and further supplementary material is available online (\url{www.daml.in.tum.de/dbu-robustness}).

\section{Related work}
\label{sec:related_work}

The existence of adversarial examples is a problematic property of neural networks \citep{szegedy2014, goodfellow2014}. Previous works have study this phenomena by proposing adversarial attacks \citep{carlini2016, brendel2018,DBLP:conf/kdd/ZugnerAG18}, defenses \citep{cisse2017, gu2015} and verification techniques  \citep{wong2017, singh2019krelu, cohen2019, DBLP:conf/icml/BojchevskiKG20, kopetzki2021}. This includes the study of different settings such as i.i.d. inputs, sequential inputs and graphs \citep{zheng2016, DBLP:conf/nips/BojchevskiG19, cheng2020, schuchardt2021}.

In the context of uncertainty estimation, robustness of the class prediction has been studied in previous works for Bayesian Neural Networks \citep{blundell2015, osawa2019, wesley2019}, drop out \citep{drop_out} or ensembles \citep{ensemble_simple} focusing on data set shifts \cite{snoek2019} or adversarial attacks \cite{robustness_bnn, cardelli2019, wicker2020}. Despite their efficient and high quality uncertainty estimates, the robustness of DBU models has not been investigated in detail yet --- indeed only for one single DBU model, \citep{malinin2019} has briefly performed attacks aiming to change the label. 
In contrast, our work focuses on a large variety of DBU models and analyzes two robustness properties: robustness of the class prediction w.r.t. adversarial perturbations and robustness of uncertainty estimation w.r.t.\ our newly proposed attacks against uncertainty measures.

This so called \emph{uncertainty attack} directly targets uncertainty estimation and are different from traditional \emph{label attacks}, which target the class prediction \citep{madry2018, raphael2020}. They allow us to jointly evaluate robustness of the class prediction and robustness of uncertainty estimation. This goes beyond previous attack defenses that were either focused on evaluating \emph{robustness w.r.t.\ class predictions} \citep{carlini2016, clever_robustness} or detecting attacks against the class prediction \citep{bypassing_attack_detection}.

Different models have been proposed to account for uncertainty while being robust.  \citep{smith2018} and \citep{simple_ood_adv_detection} have tried to improve label attack detection based on uncertainty using drop-out or density estimation. In addition to improving label attack detection for large unseen perturbations, \citep{stutz2020} aimed at improving robustness w.r.t. class label predictions on small input perturbations. They used adversarial training and soft labels for adversarial samples further from the original input. \citep{qin2020} suggested a similar adversarial training procedure, that softens labels depending on the input robustness. These previous works consider the aleatoric uncertainty that is contained in the predicted categorical probabilities, but in contrast to DBU models they are not capable of taking epistemic uncertainty into account.

Recently, four studies tried to obtain certificates on aleatoric uncertainty estimates. \citep{single_model_quantile} and \citep{confidence_certificate_rs} compute confidence intervals and certificates on softmax predictions. \citep{bitterwolf2020} uses interval bound propagation to compute bounds on softmax predictions within the $L_{\infty}$-ball around an OOD sample using ReLU networks. \citep{meinke2020} focuses on obtaining certifiably low confidence for OOD data. These four studies estimate confidence based on softmax predictions, which accounts for aleatoric uncertainty only. In this paper, we provide certificates which apply for all uncertainty measures. In particular, we use our certificates on epistemic uncertainty measures such as differential entropy which are well suited for OOD detection.

\section{Dirichlet-based uncertainty models}
\label{sec:dirichlet_models}
Standard (softmax) neural networks predict the parameters of a categorical distribution \smash{$\vp\dataix = [p\dataix_1, \ldots, p\dataix_\nclass]$} for a given input \smash{$\vx\dataix \in \mathbb{R}^{d}$}, where $\nclass$ is the number of classes. 
Given the parameters of a categorical distribution, the \emph{ aleatoric uncertainty} can be evaluated. The aleatoric uncertainty is the uncertainty on the class label prediction \smash{$y\dataix \in \{1, \ldots, \nclass\}$}. For example if we predict the outcome of an unbiased coin flip, the model is expected to have high aleatoric uncertainty and predict $p(\text{head})=0.5$.

\begin{table*}[ht]
	\centering
	\caption{Summary of DBU models. Further details on the loss functions are provided in the appendix.}
	\label{tab:dirichlet_models}
	\resizebox{.9 \textwidth}{!}{%
		\begin{tabular}{lllllll}
			\toprule
			{} &  \textbf{$\alpha\dataix$-parametrization} & \textbf{Loss} & \textbf{OOD training data} & \textbf{Ensemble training} & \textbf{Density estimation}\\
			\midrule
			\textbf{\PostNet} & $f_{\theta}(\vx\dataix) = \mathbf{1} + \bm{\alpha}\dataix$ & Bayesian loss & No & No & Yes \\
			\textbf{\PriorNet} & $f_{\theta}(\vx\dataix) = \bm{\alpha}\dataix$ & Reverse KL & Yes &  No & No \\
			\textbf{\DDNet} & $f_{\theta}(\vx\dataix) = \bm{\alpha}\dataix$ & Dir. Likelihood & No & Yes & No \\
			\textbf{\EvNet} & $f_{\theta}(\vx\dataix) =  \mathbf{1} + \bm{\alpha}\dataix$ & Expected MSE & No &  No & No \\
			\bottomrule
		\end{tabular}
	}
\end{table*}

In contrast to standard (softmax) neural networks, DBU models predict the parameters of a Dirichlet distribution -- the natural prior of categorical distributions -- given input~$\vx \dataix$ (i.e. \smash{$q\dataix = \text{Dir}(\bm{\alpha}\dataix)$} where \smash{$f_{\theta}(\vx\dataix) = \bm{\alpha} \dataix \in \mathbb{R}_+^\nclass$}). Hence, the \emph{epistemic distribution} \smash{$q\dataix$} expresses the \emph{epistemic} uncertainty on $\vx \dataix$, i.e. the uncertainty on the categorical distribution prediction \smash{$\vp\dataix$}. From the epistemic distribution, follows an estimate of the \emph{aleatoric distribution} of the class label prediction $\text{Cat}(\bar{\vp}\dataix)$ where \smash{$\E_{q\dataix}[\vp\dataix] = \bar{\vp}\dataix$}.
An advantage of DBU models is that one pass through the neural network is sufficient to compute epistemic distribution, aleatoric distribution, and predict the class label:
\begin{equation}
\begin{aligned}
    q^{(\idata)}           = \text{Dir}(\bm{\alpha}\dataix), \hspace{5pt}
    \bar{p}_\iclass\dataix  = \frac{\alpha_\iclass\dataix}{\alpha_0\dataix}, \hspace{5pt}
    y^{(\idata)}           = \arg \max_{\iclass} \;[\bar{p}_\iclass\dataix]
\end{aligned}
\end{equation}
where $\alpha_0\dataix = \sum^{\nclass}_{\iclass=1} \alpha_\iclass\dataix$. This parametrization allows to compute classic uncertainty measures in closed-form such as the total pseudo-count $m\dataix_{\alpha_0} = \sum_\iclass \alpha\dataix_\iclass$, the differential entropy of the Dirichlet distribution $m\dataix_\text{diffE} = h(\text{Dir}(\bm{\alpha}\dataix))$ or the mutual information $m\dataix_\text{MI} = I(y\dataix, \vp\dataix)$ (App. \ref{subsec:appendix_measurecomp}, \citep{malini2018}). 
Hence, these measure can efficiently be used to assign high uncertainty to unknown data, which makes DBU models specifically suited for detection of OOD samples.

Several recently proposed models for uncertainty estimations belong to the family of DBU models, such as \PriorNet, \EvNet, \DDNet and \PostNet. These models differ in terms of their parametrization of the Dirichlet distribution, the training, and density estimation. An overview of theses differences is provided in Table \ref{tab:dirichlet_models}. In our study we evaluate all recent versions of these models.

Contrary to the other models, Prior Networks \textbf{(\PriorNet)} \citep{malini2018, malinin2019}  require OOD data for training to ``teach'' the neural network the difference between ID and OOD data. \PriorNet is trained with a loss function consisting of two KL-divergence terms. The fist term is designed to learn Dirichlet parameters for ID data, while the second one is used to learn a flat Dirichlet distribution %
for OOD data: 
\begin{equation}
\begin{aligned}
    L_{\mathrm{\PriorNet}} &= \frac{1}{N} \left[\sum_{\vx\dataix \in \text{ID data}}  [\mathrm{KL} [\mathrm{Dir} (\alpha^{\mathrm{ID}}) || q\dataix]]  \right. \\
                           &+ \left.\sum_{\vx\dataix \in OOD data} [\mathrm{KL} [\mathrm{Dir} (\alpha^{\mathrm{OOD}}) || q\dataix]]\right] \\
\end{aligned}
\end{equation}
where $\alpha^{\mathrm{ID}}$ and $\alpha^{\mathrm{OOD}}$ are hyper-parameters. Usually $\alpha^{\mathrm{ID}}$ is set to $1e^{1}$ for the correct class and $1$ for all other classes, while $\alpha^{\mathrm{OOD}}$ is set to $\mathbf{1}$ for all classes.
There a two variants of \PriorNet. The first one is trained based on reverse KL-divergence \citep{malinin2019}, while the second one is trained with KL-divergence \citep{malini2018}. In our experiments, we include the most recent reverse version of \PriorNet, as it shows superior performance \citep{malinin2019}. 

Evidential Networks \textbf{(\EvNet)} \citep{sensoy2018} are trained with a loss that computes the sum of squares between the on-hot encoded true label $\vy*\dataix$ and the predicted categorical $\vp\dataix$ under the Dirichlet distribution:
\begin{equation}
\begin{aligned}
    L_{\mathrm{\EvNet}} &= \frac{1}{N} \sum_i \E_{\vp\dataix \sim \text{Dir}(\bm{\alpha}\dataix)}||\vy*\dataix - \vp\dataix||^2 \\
\end{aligned}
\end{equation}

Ensemble Distribution Distillation \textbf{(\DDNet)} \citep{malinin2019ensemble} is trained in two steps. First, an ensemble of $M$ classic neural networks needs to be trained. 
Then, the soft-labels \smash{$\{\vp_{m}\dataix\}_{m=1}^{M}$} provided by the ensemble of networks are distilled into a Dirichlet-based network by fitting them with the maximum likelihood under the Dirichlet distribution: 
\begin{equation}
\begin{aligned}
    L_{\mathrm{\DDNet}} &= - \frac{1}{N}  \sum_i \sum_{m=1}^{M} [\ln q\dataix(\pi^{im})] \\
\end{aligned}
\end{equation}
where $\pi^{im}$ denotes the soft-label of $m$th neural network. 

Posterior Network \textbf{(\PostNet)} \citep{charpentier2020} performs density estimation for ID data with normalizing flows and uses a Bayesian loss formulation: %
\begin{equation}
\begin{aligned}
    L_{\mathrm{\PostNet}} &= \frac{1}{N} \sum_i \E_{q(p\dataix)}  [\mathrm{CE} (p\dataix, y\dataix)] - H(q\dataix)
\end{aligned}
\end{equation}
where $\mathrm{CE}$ denotes the cross-entropy.
All loss functions can be computed in closed-form. For more details please have a look at the original paper on \PriorNet \citep{malini2018}, \PostNet \citep{charpentier2020}, \DDNet \citep{malinin2019} and \EvNet \citep{sensoy2018}.
Note that \EvNet and \PostNet model the Dirichlet parameters as \smash{$f_{\theta}(\vx\dataix) = 1 + \bm{\alpha}\dataix$} while \PriorNet, \RevPriorNet and \DDNet compute them as \smash{$f_{\theta}(\vx\dataix) = \bm{\alpha}\dataix$}.

\section{Robustness of Dirichlet-based uncertainty models}
\label{sec:attack_dirichlet_model}

We analyze robustness of DBU models on tasks in connection with uncertainty estimation w.r.t.\ the following four aspects: \emph{accuracy}, \emph{confidence calibration}, \emph{label attack detection} and \emph{OOD detection}. Uncertainty is quantified by differential entropy, mutual information or pseudo counts. 
A formal definition of all uncertainty estimation measures is provided in the appendix (see Section~\ref{subsec:appendix_measurecomp}).  

Robustness of Dirichlet-based uncertainty models is evaluated based on \emph{label attacks} and a newly proposed type of attacks called \emph{uncertainty attacks}. 
While label attacks aim at changing the predicted class, uncertainty attacks aim at changing the uncertainty assigned to a prediction. 
All previous works are based on label attacks and focus on robustness w.r.t. the class prediction. Thus, we are the first to propose attacks targeting uncertainty estimates such as differential entropy and analyze desirable robustness properties of DBU models beyond the class prediction. 
Label attacks and uncertainty attacks both compute a perturbed input $\tilde{\vx}\dataix$ close to the original input~$\vx\dataix$ i.e. $|| \vx\dataix - \tilde{\vx}\dataix ||_2 < r$ where $r$ is the attack radius. This perturbed input is obtained by optimizing a loss function $l(\vx)$ using Fast Gradient Sign Method (FGSM) or Projected Gradient Descent (PGD). Furthermore, we include a black box attack setting (Noise) which generates 10 noise samples from a Gaussian distribution, which is centered at the original input. From these 10 perturbed samples we choose the one with the greatest effect on the loss function and use it as attack. 
To complement attacks, we compute certificates on uncertainty estimates using median smoothing \cite{median_smoothing}.

The following questions we address by our experiments have a common assessment metric and can be treated as binary classification problems: distinguishing between correctly and wrongly classified samples, discriminating between non-attacked input and attacked inputs or differentiating between ID data and OOD data. To quantify the performance of the models on these binary classification problems, we compute the area under the precision recall curve (AUC-PR).

Experiments are performed on two image data sets (MNIST \citep{mnist} and CIFAR10 \citep{cifar10}), which contain bounded inputs and two tabular data sets (Segment \citep{uci_datasets} and Sensorless drive \citep{uci_datasets}), consisting of unbounded inputs. Note that unbounded inputs are challenging since it is impossible to describe the infinitely large OOD distribution. As PriorNet requires OOD training data, we use two further image data sets (FashionMNIST \citep{fashionmnist} and CIFAR100 \citep{cifar10}) for training on MNIST and CIFAR10, respectively. All other models are trained without OOD data. To obtain OOD data for the tabular data sets, we remove classes from the ID data set (class window for the Segment data set and class 9 for Sensorless drive) and use them as the OOD data. Further details on the experimental setup are provided in the appendix (see Section~\ref{subsec:exp_setup}).

\subsection{Uncertainty estimation under label attacks}
\label{subsec:label_attacks}
Label attacks aim at changing the predicted class. To obtain a perturbed input with a different label, we maximize the cross-entropy loss $\tilde{\vx}\dataix \approx \arg\max_\vx l(\vx) = \text{CE}(\vp\dataix, \vy\dataix)$ under the radius constraint. For the sake of completeness we additionally analyze label attacks w.r.t. to their performance of changing the class prediction and the accuracy of the neural network under label attacks constraint by different radii (see Appendix, Table~\ref{tab:acc_label_attack}). As expected and partially shown by previous works, none of the DBU models is robust against label attacks. %
However, we note that \PriorNet is slightly more robust than the other DBU models. This might be explained by the use of OOD data during training, which can be seen as some kind of robust training. 
From now on, we switch to the core focus of this work and analyze robustness properties of uncertainty estimation.

\begin{table*}[ht]
	\centering
	\caption{Distinguishing between correctly predicted and wrongly predicted labels based on the differential entropy under PGD label attacks (metric: AUC-PR).}
	\resizebox{0.8\textwidth}{!}{
		\begin{tabular}{@{}rrrrrrrc|crrrrrr@{}}
			\toprule
			& \multicolumn{6}{c}{CIFAR10} & & & \multicolumn{6}{c}{Sensorless} \\
			\cmidrule{2-7}  \cmidrule{10-15}
			Att. Rad. & 0.0 & 0.1 & 0.2 & 0.5 & 1.0 & 2.0 & & & 0.0 & 0.1 & 0.2 & 0.5 & 1.0 & 2.0  \\
			\midrule
			\PostNet  &  \bf{98.7} &  88.6 &  56.2 &   7.8 &   1.2 &   0.4 &  & %
			&  99.7 &   8.3 &   3.9 &  3.6 &  \bf{7.0} &  \bf{9.8} \\
			\PriorNet &  92.9 &  77.7 &  60.5 &  \bf{37.6} &  \bf{24.9} &  \bf{11.3} &  & %
			&  99.8 &  10.5 &   3.2 &  0.7 &  0.2 &  0.2 \\
			\DDNet    &  97.6 &  \bf{91.8} &  \bf{78.3} &  18.1 &   0.8 &   0.0 &  &%
			&  99.7 &  11.9 &   1.6 &  0.4 &  0.2 &  0.1 \\
			\EvNet    &  97.9 &  85.9 &  57.2 &  10.2 &   4.0 &   2.4 &  &%
			&  \bf{99.9} &  \bf{22.9} &  \bf{13.0} &  \bf{6.0} &  3.7 &  3.2 \\
			\bottomrule
		\end{tabular}}
	\label{tab:conf_label_attack}
\end{table*}

\textbf{Is low uncertainty a reliable indicator of correct predictions?} \\
\underline{\emph{Expected behavior:}}  Predictions with low uncertainty are more likely to be correct than high uncertainty predictions. 
\underline{\emph{Assessment metric:}} We distinguish between correctly classified samples (label 0) and wrongly classified ones (label 1) based on the differential entropy scores produced by the DBU models \citep{malini2018}. Correctly classified samples are expected to have low differential entropy, reflecting the model's confidence, and analogously wrongly predicted samples are expected to have higher differential entropy. 
\underline{\emph{Observed behavior:}} Note that the positive and negative class are not balanced, thus, the use of AUC-PR scores \citep{imbalance_apr} are important to enable meaningful measures. While uncertainty estimates are indeed an indicator of correctly classified samples on unperturbed data, none of the models maintains its high performance on perturbed data computed by PGD, FGSM or Noise label attacks (see. Table~\ref{tab:conf_label_attack}, \ref{tab:conf_label_attack_fgsm} and \ref{tab:conf_label_attack_noise_attack}). Thus, using uncertainty estimates as indicator for correctly labeled inputs is not robust to adversarial perturbations. This result is notable, since the used attacks do not target uncertainty.

\vspace{1em}
\begin{table*}[ht]
	\centering
	\caption{Label Attack-Detection by normally trained DBU models based on differential entropy under PGD label attacks (AUC-PR).}
	\resizebox{0.8\textwidth}{!}{
		\begin{tabular}{@{}rrrrrrc|crrrrr@{}}
			\toprule
			& \multicolumn{5}{c}{CIFAR10} & & & \multicolumn{5}{c}{Sensorless} \\
			\cmidrule{2-6}  \cmidrule{9-13}
			Att. Rad. & 0.1 & 0.2 & 0.5 & 1.0 & 2.0 & & & 0.1 & 0.2 & 0.5 & 1.0 & 2.0 \\
			\midrule
			\PostNet  &  \bf{63.4} &  \bf{66.9} &  42.1 &  32.9 &  31.6 &  &%
			&  47.7 &  42.3 &  36.9 &  \bf{48.5} &  \bf{85.0} \\
			\PriorNet &  53.3 &  56.0 &  55.6 &  \bf{49.2} &  42.2 & &%
			&  38.8 &  33.6 &  31.4 &  33.1 &  40.9 \\
			\DDNet    &  55.8 &  60.5 &  \bf{57.3} &  38.7 &  32.3 & &%
			&  \bf{53.5} &  42.2 &  35.0 &  32.8 &  32.6 \\
			\EvNet    &  48.4 &  46.9 &  46.3 &  46.3 &  \bf{44.5} & &%
			&  48.2 &  \bf{42.6} &  \bf{38.2} &  36.0 &  37.2 \\
			\bottomrule 			
		\end{tabular}}
	\label{tab:label_attack_detect}
\end{table*}

\textbf{Can uncertainty estimates be used to detect label attacks against the class prediction?}\\
\underline{\emph{Expected behavior:}} Adversarial examples are not from the natural data distribution. Therefore, DBU models are expected to detect them as OOD data by assigning them a higher uncertainty. We expect that perturbations computed based on a bigger attack radius~$r$ are easier to detect as their distance from the data distribution is larger. 
\underline{\emph{Assessment metric:}} The goal of attack-detection is to distinguish between unperturbed samples (label 0) and perturbed samples (label 1). Uncertainty on samples is quantified by the differential uncertainty \citep{malini2018}. Unperturbed samples are expected to have low differential entropy, because they are from the same distribution as the training data, while perturbed samples are expected to have a high differential entropy. 
\underline{\emph{Observed behavior:}} Table~\ref{tab:acc_label_attack} shows that the accuracy of all models decreases significantly under PGD label attacks, but none of the models is able to provide an equivalently increasing attack detection rate (see Table~\ref{tab:label_attack_detect}). Even larger perturbations are hard to detect for DBU models. 

\begin{figure}[ht]
	\centering
	\resizebox{0.42\textwidth}{!}{\includegraphics[width=\textwidth]{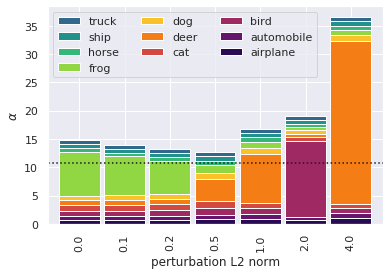}}
	\resizebox{0.42\textwidth}{!}{\includegraphics[width=\textwidth]{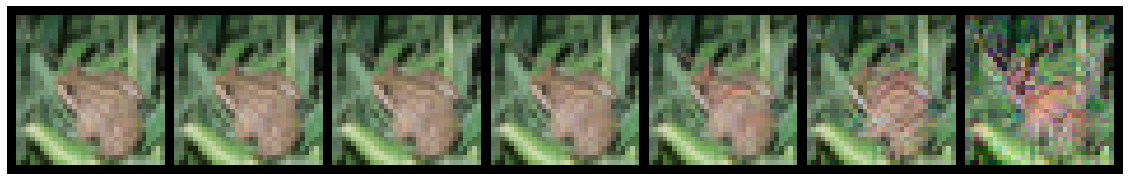}}
	\caption{Input and predicted Dirichlet-parameters under label attacks (dotted line: threshold to distinguish ID and OOD data). %
	}
	\label{fig:attaked_samples_labels}
\end{figure}

Similar results are obtained when we use mutual information or the precision~$\alpha_0$ to quantify uncertainty (see appendix Table~\ref{tab:conf_label_attack_mi} and~\ref{tab:conf_label_attack_alpha}).
Although PGD label attacks do not explicitly consider uncertainty, they seem to generate adversarial examples with similar uncertainty as the original input. 
Such high-certainty adversarial examples are illustrated in Figure~\ref{fig:attaked_samples_labels}, where certainty is visualized based on the precision~$\alpha_0$, which is supposed to be high for ID data and low for OOD data. While the original input (perturbation size $0.0$) is correctly classified as frog and ID data, there exist adversarial examples that are classified as deer or bird. The certainty ($\alpha_0$-score) on the prediction of these adversarial examples has a similar or even higher value than on the prediction of the original input. Using the differential entropy to distinguish between ID and OOD data results in the same ID/OOD assignment since the differential entropy of the three right-most adversarial examples is similar or even smaller than on the unperturbed input.

Under the less powerful FGSM and Noise attacks (see Appendix), DBU models achieve mostly higher attack detection rates than under PGD attacks. This suggests that uncertainty estimation is able to detect weak attacks, which is consistent with the observations in \citep{malinin2018_adetect} but fails under stronger PGD attacks. 

On tabular data sets, \PostNet shows a better label attack detection rate for large perturbations. This observation might be explained by the fact that the density estimation of the ID samples has been shown to work better for tabular data sets \citep{charpentier2020}. 
Overall, none of the DBU models provides a reliable indicator for adversarial inputs that target the class prediction.

\vspace{.4mm}
\begin{table*}[ht]
	\centering
	\caption{OOD detection based on differential entropy under PGD uncertainty attacks against differential entropy computed on ID data and OOD data (metric: AUC-PR).}
	\resizebox{\textwidth}{!}{
		\begin{tabular}{@{}rrrrrrrc|crrrrrr@{}}
			\toprule
			& \multicolumn{6}{c}{ID-Attack (non-attacked OOD)} &  & &  \multicolumn{6}{c}{OOD-Attack (non-attacked ID)} \\
			\cmidrule{2-7}  \cmidrule{10-15}
			Att. Rad. & 0.0 & 0.1 & 0.2 & 0.5 & 1.0 & 2.0 & & &
			0.0 & 0.1 & 0.2 & 0.5 & 1.0 & 2.0  \\
			\midrule
			& \multicolumn{14}{c}{\textbf{CIFAR10 -- SVHN}} \\
			\PostNet  &  81.8 &  64.3 &  47.2 &  22.4 &  17.6 &  \bf{16.9} &  &%
			&  81.8 &  60.5 &  40.7 &  23.3 &  21.8 &  19.8 \\
			\PriorNet &  54.4 &  40.1 &  30.0 &  17.9 &  15.6 &  15.4 &  &%
			&  54.4 &  40.7 &  30.7 &  19.5 &  16.5 &  15.7 \\
			\DDNet    &  \bf{82.8} &  \bf{71.4} &  \bf{59.2} &  \bf{28.9} &  16.0 &  15.4 & &%
			&  \bf{82.8} &  \bf{72.0} &  \bf{57.2} &  20.8 &  15.6 &  15.4 \\
			\EvNet    &  80.3 &  62.4 &  45.4 &  21.7 &  \bf{17.9} &  16.5 &  &%
			&  80.3 &  58.2 &  46.5 &  \bf{34.6} &  \bf{28.0} &  \bf{23.9} \\
			\midrule
			& \multicolumn{14}{c}{\textbf{Sens. -- Sens. class 10, 11}} \\
			\PostNet  &  \bf{74.5} &  \bf{39.8} &  \bf{36.1} &  \bf{36.0} &  \bf{45.9} &  \bf{46.0} & &%
			&  \bf{74.5} &  \bf{43.3} &  \bf{42.0} &  \bf{32.1} &  \bf{35.1} &  \bf{82.6} \\
			\PriorNet &  32.3 &  26.6 &  26.5 &  26.5 &  26.6 &  28.3 & &%
			&  32.3 &  26.7 &  26.6 &  26.6 &  27.0 &  30.4 \\
			\DDNet    &  31.7 &  26.8 &  26.6 &  26.5 &  26.6 &  27.1 & &%
			&  31.7 &  27.1 &  26.7 &  26.7 &  26.8 &  26.9 \\
			\EvNet    &  66.5 &  30.5 &  28.2 &  27.1 &  28.1 &  31.8 & &%
			&  66.5 &  38.7 &  36.1 &  30.2 &  28.2 &  28.8 \\
			\bottomrule
		\end{tabular}}
	\label{tab:id_ood_attacks}
\end{table*}

\subsection{Attacking uncertainty estimation}
\label{subsec:uncertainty_attacks}

DBU models are designed to provide sophisticated uncertainty estimates (beyond softmax scores) alongside predictions and use them to detect OOD samples. In this section, we propose and analyze a new attack type that targets these uncertainty estimates. 
DBU models enable us to compute uncertainty measures i.e. differential entropy, mutual information and precision~$\alpha_0$ in closed from (see \citep{malini2018} for a derivation). Uncertainty attacks use this closed form solution as loss function for PGD, FGSM or Noise attacks. 
Since differential entropy is the most widely used metric for ID-OOD-differentiation, we present results based on the differential entropy loss function $\tilde{\vx}\dataix \approx \arg\max_\vx l(\vx) = \text{Diff-E}(\text{Dir}(\mathbf{\alpha}\dataix))$: 
\begin{equation}
\begin{aligned}
	\text{Diff-E}(\text{Dir}(\mathbf{\alpha}\dataix))  = &\sum_c^K \ln \Gamma (\alpha_c^{(i)}) - \ln \Gamma (\alpha_0^{(i)}) \\
	&- \sum_c^K (\alpha_c^{(i)} -1) \cdot (\Psi (\alpha_c^{(i)}) - \Psi (\alpha_0^{(i)}))
\end{aligned}
\end{equation}
where $\alpha_0^{(i)} = \sum_c \alpha_c^{(i)}$. 
Result based on further uncertainty measures, loss functions and more details on attacks are provided in the appendix.

We analyze the performance of DBU models under uncertainty attacks w.r.t.\ two tasks. First, uncertainty attacks are computed on ID data aiming to indicate it as OOD data, while OOD data is left non-attacked. Second, we attack OOD data aiming to indicate it as ID data, while ID data is not attacked. Hence, uncertainty attacks target at posing ID data as OOD data and vice versa.

\begin{figure}[ht!]
    \centering
        \begin{subfigure}[t]{0.49\columnwidth}
        \centering
        \includegraphics[width=0.99 \textwidth]{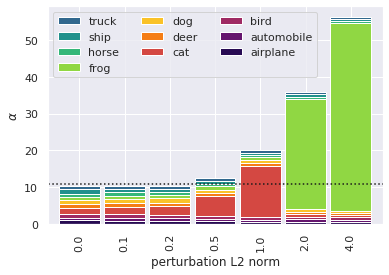}
    \end{subfigure}%
    \begin{subfigure}[t]{0.49\columnwidth}
        \centering
        \includegraphics[width=0.99 \textwidth]{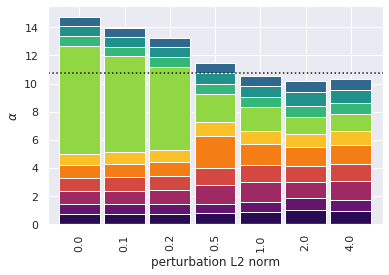}
    \end{subfigure}%

    \begin{subfigure}[t]{0.49 \columnwidth}
        \centering
        \includegraphics[width=0.99 \textwidth]{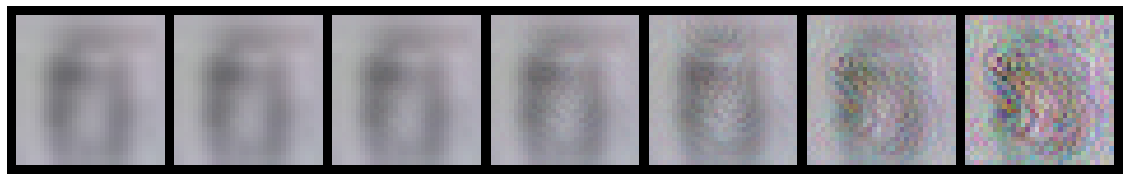}
        \caption{OOD uncertainty attack
        }
        \label{fig:attaked_samples_idood_a}
    \end{subfigure}%
        \begin{subfigure}[t]{0.49 \columnwidth}
        \centering
        \includegraphics[width=0.99 \textwidth]{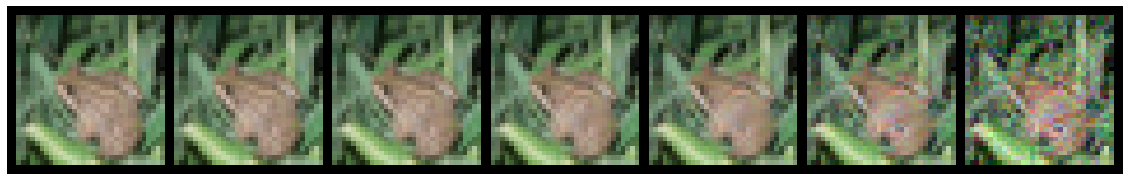}
        \caption{ID uncertainty attack
        }
        \label{fig:attaked_samples_idood_b}
    \end{subfigure}%
    \caption{ID and OOD input with corresponding Dirichlet-parameters under uncertainty attacks (dotted line: threshold to distinguish ID and OOD).}
    \label{fig:attaked_samples_idood}
\end{figure}

\textbf{Are uncertainty estimates a robust feature for OOD detection?}\\
\underline{\emph{Expected behavior:}} We expect DBU models to be able to distinguish between ID and OOD data by providing reliable uncertainty estimates, even under small perturbations. Thus, we expect uncertainty estimates of DBU models to be robust under attacks. 
\underline{\emph{Assessment metric:}} We distinguish between ID data (label 0) and OOD data (label 1) based on the differential entropy as uncertainty scoring function \citep{malini2018}. Differential entropy is expected to be small on ID samples and high on OOD samples. Experiments on further uncertainty measure and results on the AUROC metric are provided in the appendix. 
\underline{\emph{Observed behavior:}} OOD samples are perturbed as illustrated in  Figure~\ref{fig:attaked_samples_idood}. Part (a) of the figure illustrates an OOD-samples, that is correctly identified as OOD. Adding adversarial perturbations $\geq 0.5$ changes the Dirichlet parameters such that the resulting images are identified as ID, based on precision or differential entropy as uncertainty measure. Perturbing an ID sample (part (b)) results in images that are marked as OOD samples. 
OOD detection performance of all DBU models rapidly decreases with the size of the perturbation, regardless of whether attacks are computed on ID or OOD data (see Table~\ref{tab:id_ood_attacks}). This performance decrease is also observed with AUROC as metric, attacks based on FGSM, Noise, when we use mutual information or precision~$\alpha_0$ to distinguish between ID samples and OOD samples (see appendix Table~\ref{tab:id_ood_attacks_part2} - \ref{tab:id_ood_attacks_measure_diffE_aupr_noise}). 
Thus, using uncertainty estimation to distinguish between ID and OOD data is not robust.

\begin{table*}[ht!]
	\centering
	\caption{Distinguishing between correctly and wrongly predicted labels based on differential entropy under PGD label attacks. Smoothed DBU models on CIFAR10. Column format: guaranteed lowest performance $\cdot$ empirical performance $\cdot$ guaranteed highest performance (blue: normally/adversarially trained smooth classifier is more robust than the base model).}
	\label{tab:cifar10_smooth_confidence_2}
	\resizebox{\textwidth}{!}{ %
    \begin{tabular}{clccccccc}
    \toprule
    & \textbf{Att. Rad.} & 0.0 &   0.1 &  0.2 &  0.5 &  1.0 &  2.0 \\
    \midrule
      \multirow{4}{1.2cm}{Smoothed models} & \textbf{PostNet} &  $80.5\cdot\bm{{\color{black}91.5}}\cdot94.5$ &  
  $52.8\cdot\bm{{\color{black}71.6}}\cdot95.2$ &  
  $31.9\cdot\bm{{\color{black}51.0}}\cdot96.8$ &  
  $\phantom{0}5.6\cdot\bm{{\color{blue}11.7}}\cdot100.0$ &  
  $\phantom{0}0.3\cdot\bm{{\color{black}\phantom{0}0.6}}\cdot100.0$ &  
  $\phantom{0}0.0\cdot\bm{{\color{black}\phantom{0}0.0}}\cdot100.0$ \\
 & \textbf{PriorNet} &  
 $81.9\cdot\bm{{\color{black}86.8}}\cdot88.0$ &     
 $69.6\cdot\bm{{\color{blue}78.0}}\cdot90.1$ &     
 $50.9\cdot\bm{{\color{blue}65.8}}\cdot89.4$ & 
 $36.5\cdot\bm{{\color{blue}59.9}}\cdot\phantom{0}97.0$ &   
 $24.3\cdot\bm{{\color{blue}39.3}}\cdot100.0$ &    
 $\phantom{0}9.2\cdot\bm{{\color{blue}17.9}}\cdot100.0$ \\
  &   \textbf{DDNet} &  
  $65.9\cdot\bm{{\color{black}81.2}}\cdot83.0$ &  
  $55.8\cdot\bm{{\color{black}70.5}}\cdot87.2$ &  
  $37.8\cdot\bm{{\color{black}56.8}}\cdot88.1$ &  
  $10.1\cdot\bm{{\color{blue}21.9}}\cdot\phantom{0}94.3$ &      
  $\phantom{0}0.9\cdot\phantom{0}\bm{{\color{blue}1.6}}\cdot\phantom{0}99.6$ &                   
  $\phantom{0}0.0\cdot\phantom{0}\bm{0.0}\cdot100.0$ \\
  &   \textbf{EvNet} &  
  $76.3\cdot\bm{{\color{black}90.2}}\cdot91.7$ &  
  $54.7\cdot\bm{{\color{black}74.3}}\cdot95.7$ &  
  $31.6\cdot\bm{{\color{black}51.5}}\cdot94.5$ &   
  $\phantom{0}5.8\cdot\bm{{\color{blue}11.9}}\cdot\phantom{0}86.9$ &     
  $\phantom{0}1.9\cdot\bm{{\color{blue}\phantom{0}7.0}}\cdot100.0$ &     
  $\phantom{0}1.1\cdot\bm{{\color{blue}\phantom{0}4.0}}\cdot100.0$ \\

    \midrule
      \multirow{4}{1.2cm}{Smoothed + adv. w. label attacks} &  
  \textbf{PostNet} &  - &  
  $52.1\cdot\bm{{\color{black}71.8}}\cdot95.6$ &  
  $31.2\cdot\bm{{\color{black}47.9}}\cdot96.1$ &  
  $\phantom{0}7.8\cdot\bm{{\color{blue}14.7}}\cdot\phantom{0}98.6$ &    
  $\phantom{0}1.8\cdot\phantom{0}\bm{{\color{blue}4.4}}\cdot100.0$ &  
  $\phantom{0}0.3\cdot\phantom{0}\bm{{\color{black}0.5}}\cdot100.0$ \\
 & \textbf{PriorNet} &  - &  
 $57.6\cdot\bm{{\color{black}71.7}}\cdot88.9$ &     
 $46.1\cdot\bm{{\color{blue}64.5}}\cdot90.1$ &  
 $38.1\cdot\bm{{\color{blue}59.3}}\cdot\phantom{0}99.5$ &  
 $32.3\cdot\bm{{\color{blue}51.7}}\cdot100.0$ &    
 $22.1\cdot\bm{{\color{blue}41.6}}\cdot\phantom{0}97.4$ \\
   & \textbf{DDNet} &  - &  
   $58.6\cdot\bm{{\color{black}78.4}}\cdot92.2$ &  
   $49.4\cdot\bm{{\color{black}66.0}}\cdot90.5$ &  
   $12.0\cdot\bm{{\color{blue}21.4}}\cdot\phantom{0}98.1$ &     
   $\phantom{0}0.8\cdot\phantom{0}\bm{{\color{blue}1.0}}\cdot\phantom{0}96.6$ &                   
   $\phantom{0}0.0\cdot\phantom{0}\bm{0.0}\cdot100.0$ \\
    & \textbf{EvNet} &  - &  
    $24.3\cdot\bm{{\color{black}34.2}}\cdot51.8$ &  
    $32.6\cdot\bm{{\color{black}49.5}}\cdot95.5$ &  
    $\phantom{0}5.9\cdot\bm{{\color{blue}13.0}}\cdot100.0$ &  
    $\phantom{0}2.6\cdot\phantom{0}\bm{{\color{black}5.2}}\cdot\phantom{0}99.9$ &     
    $\phantom{0}2.9\cdot\phantom{0}\bm{{\color{blue}5.9}}\cdot100.0$ \\

    \midrule
    \multirow{4}{1.3cm}{Smoothed + adv.\ w. uncert. attacks} &   
\textbf{PostNet} &  - &  
$52.8\cdot\bm{{\color{black}74.2}}\cdot94.6$ &  
$33.0\cdot\bm{{\color{black}49.4}}\cdot87.5$ &   
$\phantom{0}7.7\cdot\bm{{\color{blue}14.2}}\cdot\phantom{0}99.0$ &  
$\phantom{0}0.6\cdot\phantom{0}\bm{{\color{black}1.2}}\cdot100.0$ &    
$\phantom{0}0.7\cdot\phantom{0}\bm{{\color{blue}1.1}}\cdot100.0$ \\
 & \textbf{PriorNet} &  - &  
 $50.6\cdot\bm{{\color{black}68.1}}\cdot88.6$ &     
 $44.4\cdot\bm{{\color{blue}66.1}}\cdot96.0$ &  
 $35.1\cdot\bm{{\color{blue}57.4}}\cdot\phantom{0}98.4$ &   
 $18.4\cdot\bm{{\color{blue}32.2}}\cdot100.0$ &  
 $15.2\cdot\bm{{\color{blue}29.3}}\cdot100.0$ \\
   & \textbf{DDNet} &  - &  
   $68.8\cdot\bm{{\color{black}84.4}}\cdot93.2$ &  
   $45.1\cdot\bm{{\color{black}60.8}}\cdot86.8$ &  
   $12.3\cdot\bm{{\color{blue}22.0}}\cdot\phantom{0}91.0$ &      
   $\phantom{0}0.8\cdot\phantom{0}\bm{{\color{blue}1.7}}\cdot\phantom{0}87.0$ &                  
   $\phantom{0}0.0\cdot\phantom{0}\bm{0.0}\cdot100.0$ \\
&    \textbf{EvNet} &  - &  
$54.2\cdot\bm{{\color{black}73.7}}\cdot96.1$ &  
$30.5\cdot\bm{{\color{black}50.0}}\cdot99.5$ &  
$\phantom{0}7.1\cdot\bm{{\color{blue}13.9}}\cdot100.0$ &      
$\phantom{0}3.7\cdot\phantom{0}\bm{{\color{blue}8.7}}\cdot\phantom{0}75.2$ &    
$\phantom{0}3.3\cdot\phantom{0}\bm{{\color{blue}5.8}}\cdot100.0$ \\

    \bottomrule
    \end{tabular}}
\end{table*}

\begin{table*}[ht!]
	\centering
	\caption{Attack detection (PGD label attacks) based on differential entropy. Smoothed DBU models on CIFAR10. Column format: guaranteed lowest performance $\cdot$ empirical performance $\cdot$ guaranteed highest performance (blue: normally/adversarially trained smooth classifier is more robust than the base model).}
	\label{tab:cifar10_smooth_attackdetection_2}
	\resizebox{\textwidth}{!}{
		\begin{tabular}{clcccccc}
			\toprule
			& \textbf{Att. Rad.} &   0.1 &  0.2 &  0.5 &  1.0 &  2.0 \\
			\midrule
			  \multirow{4}{1.2cm}{Smoothed models}  & \textbf{PostNet} &  %
  $33.1\cdot\bm{{\color{black}50.4}}\cdot89.9$ &  
  $31.0\cdot\bm{{\color{black}50.2}}\cdot96.9$ &    
  $30.7\cdot\bm{{\color{blue}50.2}}\cdot100.0$ &    
  $30.7\cdot\bm{{\color{blue}50.0}}\cdot100.0$ &  
  $30.7\cdot\bm{{\color{blue}50.2}}\cdot100.0$ \\
 & \textbf{PriorNet} &  
 $35.9\cdot\bm{{\color{black}50.6}}\cdot74.5$ &  
 $33.0\cdot\bm{{\color{black}50.3}}\cdot82.8$ &  
 $31.2\cdot\bm{{\color{black}50.0}}\cdot\phantom{0}95.7$ &  
 $30.7\cdot\bm{{\color{black}50.4}}\cdot\phantom{0}99.9$ &  
 $30.7\cdot\bm{{\color{blue}50.4}}\cdot100.0$ \\
   & \textbf{DDNet} &  
   $36.3\cdot\bm{{\color{black}50.3}}\cdot76.4$ &  
   $32.8\cdot\bm{{\color{black}49.9}}\cdot84.6$ &  
   $30.8\cdot\bm{{\color{black}50.1}}\cdot\phantom{0}98.0$ &    
   $30.7\cdot\bm{{\color{blue}50.2}}\cdot100.0$ &  
   $30.7\cdot\bm{{\color{blue}50.2}}\cdot100.0$ \\
&    \textbf{EvNet} &  
$32.9\cdot\bm{{\color{black}50.4}}\cdot89.8$ &  
$31.4\cdot\bm{{\color{black}50.1}}\cdot94.0$ &     
$30.8\cdot\bm{{\color{blue}50.0}}\cdot\phantom{0}98.0$ &    
$30.7\cdot\bm{{\color{blue}50.3}}\cdot100.0$ &  
$30.7\cdot\bm{{\color{blue}49.6}}\cdot100.0$ \\

			\midrule
			\multirow{4}{1.3cm}{Smoothed + adv.\ w. label attacks}  & 
\textbf{PostNet} &   
$32.7\cdot\bm{{\color{black}50.1}}\cdot90.4$ &  
$31.1\cdot\bm{{\color{black}50.2}}\cdot96.5$ &     
$30.7\cdot\bm{{\color{blue}50.2}}\cdot\phantom{0}99.7$ &     
$30.7\cdot\bm{{\color{blue}50.3}}\cdot100.0$ &  
$30.7\cdot\bm{{\color{blue}50.2}}\cdot100.0$ \\
 & \textbf{PriorNet} & %
 $35.2\cdot\bm{{\color{black}51.8}}\cdot78.6$ &  
 $32.8\cdot\bm{{\color{black}51.1}}\cdot84.4$ &  
 $30.8\cdot\bm{{\color{black}50.2}}\cdot\phantom{0}98.7$ &  
 $30.7\cdot\bm{{\color{black}50.5}}\cdot100.0$ &   
 $30.8\cdot\bm{{\color{blue}50.1}}\cdot\phantom{0}98.2$ \\
   & \textbf{DDNet} &  %
   $35.5\cdot\bm{{\color{black}50.6}}\cdot79.2$ &  
   $33.4\cdot\bm{{\color{black}50.3}}\cdot84.1$ &  
   $30.8\cdot\bm{{\color{black}50.1}}\cdot\phantom{0}99.2$ &     
   $30.7\cdot\bm{{\color{blue}50.0}}\cdot100.0$ & 
   $30.7\cdot\bm{{\color{blue}50.5}}\cdot100.0$ \\
&    \textbf{EvNet} &  %
$40.3\cdot\bm{{\color{black}50.4}}\cdot66.8$ &  
$31.4\cdot\bm{{\color{black}50.3}}\cdot95.8$ &    
$30.7\cdot\bm{{\color{blue}50.3}}\cdot100.0$ &     
$30.7\cdot\bm{{\color{blue}50.1}}\cdot100.0$ &  
$30.7\cdot\bm{{\color{blue}50.0}}\cdot100.0$ \\
			\midrule
			\multirow{4}{1.3cm}{Smoothed + adv.\ w. uncert. attacks} & 
\textbf{PostNet} &  %
$33.3\cdot\bm{{\color{black}50.6}}\cdot88.7$ &  
$32.5\cdot\bm{{\color{black}50.1}}\cdot87.9$ &     
$30.7\cdot\bm{{\color{blue}49.9}}\cdot\phantom{0}99.8$ &     
$30.7\cdot\bm{{\color{blue}50.1}}\cdot100.0$ &  
$30.7\cdot\bm{{\color{blue}50.0}}\cdot100.0$ \\
& \textbf{PriorNet} &  %
$34.5\cdot\bm{{\color{black}51.0}}\cdot80.1$ &  
$31.4\cdot\bm{{\color{black}50.6}}\cdot92.8$ &  
$30.9\cdot\bm{{\color{black}50.0}}\cdot\phantom{0}97.7$ &  
$30.7\cdot\bm{{\color{black}50.1}}\cdot100.0$ &  
$30.7\cdot\bm{{\color{blue}50.0}}\cdot100.0$ \\
 &   \textbf{DDNet} &  %
 $37.4\cdot\bm{{\color{black}50.8}}\cdot74.5$ &  
 $33.4\cdot\bm{{\color{black}50.2}}\cdot83.0$ &  
 $30.9\cdot\bm{{\color{black}50.1}}\cdot\phantom{0}96.8$ &      
 $30.8\cdot\bm{{\color{blue}49.9}}\cdot\phantom{0}98.1$ &  
 $30.7\cdot\bm{{\color{blue}49.9}}\cdot100.0$ \\
  &  \textbf{EvNet} &  %
  $32.8\cdot\bm{{\color{black}50.1}}\cdot92.0$ &  
  $30.8\cdot\bm{{\color{black}50.0}}\cdot99.6$ &    
  $30.7\cdot\bm{{\color{blue}50.1}}\cdot100.0$ &      
  $31.2\cdot\bm{{\color{blue}50.2}}\cdot\phantom{0}96.1$ &  
  $31.0\cdot\bm{{\color{blue}50.0}}\cdot100.0$ \\

			\bottomrule
		\end{tabular}}
\end{table*}

\begin{table*}[ht!]
	\centering
	\caption{OOD detection based on differential entropy under PGD uncertainty attacks against differential entropy on ID data and OOD data. Smoothed DBU models on CIFAR10. Column format: guaranteed lowest performance $\cdot$ empirical performance $\cdot$ guaranteed highest performance (blue: normally/adversarially trained smooth classifier is more robust than the base model).}
	\label{tab:cifar10_smooth_ooddetection_2}
	\resizebox{\textwidth}{!}{
		\begin{tabular}{clccccccc}
			\toprule
			& \textbf{Att. Rad.} & 0.0 &   0.1 &  0.2 &  0.5 &  1.0 &  2.0 \\
			\midrule
			& & \multicolumn{6}{c}{\textbf{ID-Attack}} \\
			  \multirow{4}{1.2cm}{Smoothed models} &  
  \textbf{PostNet} &     
  $72.1\cdot\bm{{\color{blue}82.7}}\cdot88.0$ &  
  $35.0\cdot\bm{{\color{black}56.6}}\cdot97.4$ &     
  $31.9\cdot\bm{{\color{blue}65.6}}\cdot99.8$ &  
  $30.7\cdot\bm{{\color{blue}50.6}}\cdot100.0$ &  
  $30.7\cdot\bm{{\color{blue}46.9}}\cdot100.0$ &  
  $30.7\cdot\bm{{\color{blue}51.6}}\cdot100.0$ \\
& \textbf{PriorNet} &  
$50.2\cdot\bm{{\color{black}53.1}}\cdot55.9$ &     
$33.5\cdot\bm{{\color{blue}43.3}}\cdot65.3$ &     
$31.3\cdot\bm{{\color{blue}39.7}}\cdot69.1$ &   
$31.3\cdot\bm{{\color{blue}48.3}}\cdot\phantom{0}98.2$ &   
$30.7\cdot\bm{{\color{blue}44.4}}\cdot\phantom{0}99.9$ &  
$30.7\cdot\bm{{\color{blue}45.4}}\cdot100.0$ \\
 &   \textbf{DDNet} &  
 $72.0\cdot\bm{{\color{black}75.8}}\cdot79.8$ &  
 $35.6\cdot\bm{{\color{black}46.2}}\cdot69.8$ &  
 $32.9\cdot\bm{{\color{black}50.3}}\cdot87.1$ &   
 $31.1\cdot\bm{{\color{blue}58.7}}\cdot\phantom{0}98.6$ & 
 $30.7\cdot\bm{{\color{blue}59.3}}\cdot100.0$ &  
 $30.7\cdot\bm{{\color{blue}44.5}}\cdot100.0$ \\
  &  \textbf{EvNet} &     
  $79.5\cdot\bm{{\color{blue}87.1}}\cdot92.8$ &  
  $34.1\cdot\bm{{\color{black}58.6}}\cdot95.1$ &     
  $32.5\cdot\bm{{\color{blue}61.2}}\cdot96.9$ &   
  $31.7\cdot\bm{{\color{blue}60.6}}\cdot\phantom{0}98.7$ &  
  $30.7\cdot\bm{{\color{blue}62.4}}\cdot100.0$ &  
  $30.7\cdot\bm{{\color{blue}57.3}}\cdot100.0$ \\

			\midrule
			\multirow{4}{1.3cm}{Smoothed + adv.\ w. label attacks} &  
\textbf{PostNet} &  - & 
$35.0\cdot\bm{{\color{black}58.5}}\cdot97.7$ &  
$31.2\cdot\bm{{\color{black}46.6}}\cdot97.4$ &   
$30.8\cdot\bm{{\color{blue}57.7}}\cdot\phantom{0}99.7$ &  
$30.7\cdot\bm{{\color{blue}49.8}}\cdot100.0$ &  
$30.7\cdot\bm{{\color{blue}50.9}}\cdot100.0$ \\
& \textbf{PriorNet} &  - &  
$31.5\cdot\bm{{\color{black}36.7}}\cdot57.2$ &     
$33.1\cdot\bm{{\color{blue}51.8}}\cdot84.8$ &   
$30.7\cdot\bm{{\color{blue}57.7}}\cdot\phantom{0}98.7$ &   
$30.7\cdot\bm{{\color{blue}40.0}}\cdot\phantom{0}99.9$ &   
$30.9\cdot\bm{{\color{blue}53.6}}\cdot\phantom{0}96.7$ \\
 &   \textbf{DDNet} &  - &  
 $36.2\cdot\bm{{\color{black}50.0}}\cdot78.6$ &  
 $32.1\cdot\bm{{\color{black}41.3}}\cdot70.2$ &  
 $30.8\cdot\bm{{\color{blue}56.4}}\cdot100.0$ &  
 $30.7\cdot\bm{{\color{blue}49.4}}\cdot100.0$ &  
 $30.7\cdot\bm{{\color{blue}54.8}}\cdot100.0$ \\
  &  \textbf{EvNet} &  - &  
  $46.8\cdot\bm{{\color{black}61.0}}\cdot79.7$ &     
  $32.3\cdot\bm{{\color{blue}58.9}}\cdot99.1$ &  
  $30.7\cdot\bm{{\color{blue}45.0}}\cdot100.0$ &  
  $30.7\cdot\bm{{\color{blue}63.3}}\cdot100.0$ &  
  $30.8\cdot\bm{{\color{blue}38.1}}\cdot100.0$ \\

			\midrule
			\multirow{4}{1.3cm}{Smoothed + adv.\ w. uncert. attacks} &  
\textbf{PostNet} &  - &  
$35.2\cdot\bm{{\color{black}55.9}}\cdot96.0$ &     
$34.5\cdot\bm{{\color{blue}59.2}}\cdot94.9$ &  
$30.7\cdot\bm{{\color{blue}47.0}}\cdot100.0$ &  
$30.7\cdot\bm{{\color{blue}58.2}}\cdot100.0$ &  
$30.7\cdot\bm{{\color{blue}42.9}}\cdot100.0$ \\
 & \textbf{PriorNet} &  - &  
 $31.8\cdot\bm{{\color{black}38.9}}\cdot64.1$ &     
 $31.0\cdot\bm{{\color{blue}41.8}}\cdot87.9$ &  
 $30.7\cdot\bm{{\color{blue}42.9}}\cdot\phantom{0}99.2$ & 
 $30.7\cdot\bm{{\color{blue}48.6}}\cdot100.0$ & 
 $30.7\cdot\bm{{\color{blue}46.6}}\cdot100.0$ \\
   & \textbf{DDNet} &  - & 
   $39.7\cdot\bm{{\color{black}52.1}}\cdot75.7$ &  
   $36.4\cdot\bm{{\color{black}56.8}}\cdot83.8$ &   
   $31.0\cdot\bm{{\color{blue}51.5}}\cdot\phantom{0}97.4$ &  
   $31.0\cdot\bm{{\color{blue}56.8}}\cdot\phantom{0}97.8$ &  
   $30.7\cdot\bm{{\color{blue}49.1}}\cdot100.0$ \\
&    \textbf{EvNet} &  - &     
$34.8\cdot\bm{{\color{blue}64.9}}\cdot99.6$ &     
$30.8\cdot\bm{{\color{blue}48.9}}\cdot99.8$ &  
$30.7\cdot\bm{{\color{blue}66.8}}\cdot100.0$ &  
$30.9\cdot\bm{{\color{blue}41.5}}\cdot\phantom{0}93.8$ &  
$31.1\cdot\bm{{\color{blue}55.1}}\cdot100.0$ \\

			\midrule
			\midrule
			& & \multicolumn{6}{c}{\textbf{OOD-Attack}} \\
			 \multirow{4}{1.2cm}{Smoothed models} &   
 \textbf{PostNet} &     
 $72.0\cdot\bm{{\color{blue}82.7}}\cdot88.0$ &  
 $35.1\cdot\bm{{\color{black}56.8}}\cdot97.3$ &     
 $32.0\cdot\bm{{\color{blue}65.8}}\cdot99.8$ &  
 $30.7\cdot\bm{{\color{blue}50.7}}\cdot100.0$ &  
 $30.7\cdot\bm{{\color{blue}46.5}}\cdot100.0$ &  
 $30.7\cdot\bm{{\color{blue}51.7}}\cdot100.0$ \\
 & \textbf{PriorNet} &  
 $50.3\cdot\bm{{\color{black}53.1}}\cdot55.9$ &     
 $33.6\cdot\bm{{\color{blue}43.7}}\cdot65.9$ &     
 $31.3\cdot\bm{{\color{blue}39.8}}\cdot69.4$ &   
 $31.3\cdot\bm{{\color{blue}48.3}}\cdot\phantom{0}98.2$ &   
 $30.7\cdot\bm{{\color{blue}44.5}}\cdot\phantom{0}99.9$ &  
 $30.7\cdot\bm{{\color{blue}46.4}}\cdot100.0$ \\
   & \textbf{DDNet} &  
   $72.0\cdot\bm{{\color{black}75.8}}\cdot79.8$ &  
   $35.6\cdot\bm{{\color{black}46.2}}\cdot70.0$ &  
   $32.9\cdot\bm{{\color{black}50.1}}\cdot86.7$ &   
   $31.1\cdot\bm{{\color{blue}58.8}}\cdot\phantom{0}98.6$ &  
   $30.7\cdot\bm{{\color{blue}59.3}}\cdot100.0$ &  
   $30.7\cdot\bm{{\color{blue}44.6}}\cdot100.0$ \\
&    \textbf{EvNet} &     
$79.5\cdot\bm{{\color{blue}87.1}}\cdot92.8$ &     
$34.1\cdot\bm{{\color{blue}58.8}}\cdot95.2$ &     
$32.6\cdot\bm{{\color{blue}61.2}}\cdot96.9$ &   
$31.7\cdot\bm{{\color{blue}60.5}}\cdot\phantom{0}98.7$ &  
$30.7\cdot\bm{{\color{blue}62.4}}\cdot100.0$ &  
$30.7\cdot\bm{{\color{blue}57.6}}\cdot100.0$ \\

			\midrule
			\multirow{4}{1.3cm}{Smoothed + adv.\ w. label attacks} &  
\textbf{PostNet} &  - &  
$35.0\cdot\bm{{\color{black}58.5}}\cdot97.8$ &     
$31.2\cdot\bm{{\color{blue}46.6}}\cdot97.2$ &   
$30.8\cdot\bm{{\color{blue}57.7}}\cdot\phantom{0}99.7$ &  
$30.7\cdot\bm{{\color{blue}50.2}}\cdot100.0$ & 
$30.7\cdot\bm{{\color{blue}51.5}}\cdot100.0$ \\
 & \textbf{PriorNet} &  - & 
 $31.6\cdot\bm{{\color{black}37.3}}\cdot59.3$ &    
 $33.2\cdot\bm{{\color{blue}52.7}}\cdot85.8$ & 
 $30.7\cdot\bm{{\color{blue}57.8}}\cdot\phantom{0}98.7$ & 
 $30.7\cdot\bm{{\color{blue}40.1}}\cdot\phantom{0}99.9$ & 
 $30.9\cdot\bm{{\color{blue}53.8}}\cdot\phantom{0}96.8$ \\
   & \textbf{DDNet} &  - & 
   $36.4\cdot\bm{{\color{black}50.2}}\cdot78.9$ &  
   $32.1\cdot\bm{{\color{black}41.5}}\cdot70.4$ & 
   $30.9\cdot\bm{{\color{blue}56.2}}\cdot100.0$ & 
   $30.7\cdot\bm{{\color{blue}49.3}}\cdot100.0$ &
   $30.7\cdot\bm{{\color{blue}55.1}}\cdot100.0$ \\
&    \textbf{EvNet} &  - &    
$47.2\cdot\bm{{\color{blue}61.1}}\cdot80.0$ &   
$32.4\cdot\bm{{\color{blue}59.1}}\cdot99.1$ &  
$30.7\cdot\bm{{\color{blue}45.0}}\cdot100.0$ &  
$30.7\cdot\bm{{\color{blue}63.2}}\cdot100.0$ & 
$30.8\cdot\bm{{\color{blue}38.0}}\cdot100.0$ \\

			\midrule
			\multirow{4}{1.3cm}{Smoothed + adv.\ w. uncert. attacks} &
\textbf{PostNet} &  - &  
$35.3\cdot\bm{{\color{black}56.4}}\cdot96.1$ &     
$34.5\cdot\bm{{\color{blue}59.0}}\cdot94.9$ &  
$30.7\cdot\bm{{\color{blue}46.8}}\cdot100.0$ &  
$30.7\cdot\bm{{\color{blue}57.8}}\cdot100.0$ &  
$30.7\cdot\bm{{\color{blue}43.2}}\cdot100.0$ \\
& \textbf{PriorNet} &  - &  
$31.9\cdot\bm{{\color{black}39.4}}\cdot65.5$ &     
$31.0\cdot\bm{{\color{blue}42.0}}\cdot88.6$ &  
$30.7\cdot\bm{{\color{blue}42.9}}\cdot\phantom{0}99.2$ & 
$30.7\cdot\bm{{\color{blue}48.4}}\cdot100.0$ & 
$30.7\cdot\bm{{\color{blue}47.1}}\cdot100.0$ \\
 &   \textbf{DDNet} &  - & 
 $40.2\cdot\bm{{\color{black}52.9}}\cdot76.5$ & 
 $36.4\cdot\bm{{\color{black}56.9}}\cdot83.9$ & 
 $31.1\cdot\bm{{\color{blue}51.5}}\cdot\phantom{0}97.3$ &  
 $31.0\cdot\bm{{\color{blue}57.0}}\cdot\phantom{0}97.8$ & 
 $30.7\cdot\bm{{\color{blue}49.1}}\cdot100.0$ \\
  &  \textbf{EvNet} &  - &     
  $34.9\cdot\bm{{\color{blue}64.8}}\cdot99.6$ &   
  $30.8\cdot\bm{{\color{blue}48.8}}\cdot99.8$ & 
  $30.7\cdot\bm{{\color{blue}66.1}}\cdot100.0$ & 
  $30.9\cdot\bm{{\color{blue}41.6}}\cdot\phantom{0}93.6$ &
  $31.1\cdot\bm{{\color{blue}54.7}}\cdot100.0$ \\

			\bottomrule
		\end{tabular}}
\end{table*}

\subsection{How to make DBU models more robust?}

Our robustness analysis based on label attacks and uncertainty attacks shows that predictions, uncertainty estimation and the differentiation between ID and OOD data are not robust. Next, we explore approaches to improve robustness properties of DBU models w.r.t.\ these tasks based on randomized smoothing and adversarial training. 

\textbf{Randomized smoothing} was originally proposed for certification of classifiers \cite{cohen2019}.
The core idea is to draw multiple samples $\vx\dataix_s \sim \DNormal(\vx\dataix, \sigma)$ around the input data $\vx\dataix$, to feed all these samples through the neural network, and to aggregate the resulting set of predictions (e.g. by taking their mean), to get a smoothed prediction. Besides allowing certification, as a side effect, the smoothed model is more robust. Our idea is to use randomized smoothing to improve robustness of DBU models, particularly w.r.t.\ uncertainty estimation. In contrast to discrete class predictions, however, certifying uncertainty estimates such as differential entropy scores requires a smoothing approach that is able to handle continuous values as in regression tasks. So far, only few works for randomized smoothing for regression models have been proposed \citep{confidence_certificate_rs,median_smoothing}. We choose median smoothing \citep{median_smoothing}, because it is applicable to unbounded domains as required for the uncertainty estimates covered in this work. In simple words: The set of uncertainty scores obtained from the $\vx\dataix_s \sim \DNormal(\vx\dataix, \sigma)$ is aggregated by taking their median. 

In the following experiments we focus on differential entropy as the uncertainty score. We denote the resulting smoothed differential entropy, i.e. the median output, as $m(\vx\dataix)$.
Intuitively, we expect that the random sampling around a data point as well as the outlier-insensitivity of the median to improve the robustness of the uncertainty estimates w.r.t.\ adversarial examples.

To measure the performance and robustness of our smoothed DBU models, we apply median smoothing on the same tasks as in the previous sections, i.e., distinguishing between correctly and wrongly labeled inputs, attack detection, OOD detection and compute the corresponding AUC-PR score under label attacks and uncertainty attacks. 
The bold, middle part of the columns in Tables~\ref{tab:cifar10_smooth_confidence_2}, \ref{tab:cifar10_smooth_attackdetection_2}, and~\ref{tab:cifar10_smooth_ooddetection_2} show the AUC-PR scores on CIFAR10, which we call \emph{empirical performance} of the smoothed models. To facilitate the comparison with the base model of Section~\ref{sec:attack_dirichlet_model}, we highlight the AUC-PR scores in blue in cases where the smooth model is more robust. The highlighting clearly shows that randomized smoothing increases the robustness of the empirical performance on OOD detection. 
OOD detection under strong PGD attacks (attack radius $\geq 0.5$) performs comparable to random guessing (i.e. AUC-PR scores around $50\%$ whith $50\%$ ID and $50\%$ OOD data). This shows that DBU models are not reliably efficient w.r.t. this task.
In attack detection and distinguishing between correctly and wrongly predicted labels the smoothed DBU model are mostly more robust than the base models for attack radii $\geq 0.5$.

\textbf{Certified performance.} Using the median based on smoothing improves the empirical robustness, but it does not provide formal guarantees how low/high the performance might actually get under perturbed data (since any attack is only a heuristic). 
Here, we propose novel guarantees by exploiting the individual certificates we obtain via randomized smoothing.
 Note that the certification procedure \citep{median_smoothing} enables us to derive lower and upper bounds $\underline{m}(\vx\dataix) \leq m(\vx\dataix) \leq \overline{m}(\vx\dataix)$ which hold with high probability and indicate how much the median might change in the worst-case when $\vx\dataix$ gets perturbed subject to a specific (attack) radius.
 
These bounds allow us to compute certificates that bound the performance of the smooth models, which we refer to as the \emph{guaranteed lowest performance} and \emph{guaranteed highest performance}. More precisely, for the guaranteed lowest performance of the model we take the pessimistic view that all ID data points realize their individual upper bounds $\overline{m}(\vx\dataix)$, i.e.\ have their highest possible uncertainty (worst case). On the other hand, we assume all OOD samples realize their lower bounds $\underline{m}(\vx\dataix_s)$. Using these values as the uncertainty scores for all data points we obtain the guaranteed lowest performance of the model. 
A guaranteed lowest performance of e.g. $35.0$ means that even under the worst case conditions an attack is not able to decrease the performance below $35.0$. 
Analogously, we can take the optimistic view to obtain the guaranteed highest performance of the smoothed models. 
Tables~\ref{tab:cifar10_smooth_confidence_2}, \ref{tab:cifar10_smooth_attackdetection_2} and~\ref{tab:cifar10_smooth_ooddetection_2} show the guaranteed lowest/highest performance (non-bold, left/right of the empirical performance). 
Our results show that the difference between guaranteed highest and guaranteed lowest performance increases with the attack radius, which might be explained by the underlying lower/upper bounds on the median being tighter for smaller perturbations.

\textbf{Adversarial training.}
Randomized smoothing improves robustness of DBU models and allows us to compute performance guarantees. However, an open question is whether it is possible to increase robustness even further by combining it with adversarial training. To obtain adversarially trained models we augment the data set using perturbed samples that are computed by PGD attacks against the cross-entropy loss (label attacks) or the differential entropy (uncertainty attacks). These perturbed samples $\tilde{\vx}\dataix$ are computed during each epoch of the training based on inputs $\vx\dataix$ and added to the training data (with the label $y\dataix$ of the original input). 
Tables~\ref{tab:cifar10_smooth_confidence_2}, \ref{tab:cifar10_smooth_attackdetection_2}, and~\ref{tab:cifar10_smooth_ooddetection_2} illustrate the results. We choose the attack radius used during training and the $\sigma$ used for smoothing to be equal. %
To facilitate comparison, we highlight the empirical performance of the adversarially trained models in blue if it is better than the performance of the base model. Our results show that the additional use of adversarial training has a minor effect on the robustness and does not result in a significant further increase of the robustness. 

We conclude that median smoothing is a promising technique to increase robustness w.r.t.\ distinguishing between correctly labeled samples and wrongly labeled samples, attack detection and differentiation between in-distribution data and out-of-distribution data of all Dirichlet-based uncertainty models, while additional adversarial training has a minor positive effect on robustness.

\section{Conclusion}
\label{sec:conclusion}

This work analyzes robustness of uncertainty estimation by DBU models and answers multiple questions in this context. Our results show: (1) While uncertainty estimates are a good indicator to identify correctly classified samples on unperturbed data, performance decrease drastically on perturbed data-points. (2) None of the Dirichlet-based uncertainty models is able to detect PGD label attacks against the class prediction by uncertainty estimation, regardless of the used uncertainty measure. (3) Detecting OOD samples and distinguishing between ID-data and OOD-data is not robust. (4) Applying median smoothing to  uncertainty estimates increases robustness of DBU models w.r.t. all analyzed tasks, while adversarial training based on label or uncertainty attacks resulted in minor improvements.

\section*{Acknowledgments}

This research was supported by BMW AG.

\bibliographystyle{icml2021}
\bibliography{literature}

\newpage
~\newpage %
\section{Appendix}
\label{sec:appendix}

\subsection{Closed-form computation of uncertainty measures \& Uncertainty attacks}
\label{subsec:appendix_measurecomp}

Dirichlet-based uncertainty models allow to compute several uncertainty measures in closed form (see \citep{malini2018} for a derivation). As proposed by \cite{malini2018}, we use precision~$m_{\alpha_0}$, differential entropy~$m_{\mathrm{diffE}}$ and mutual information~$m_{\mathrm{MI}}$ to estimate uncertainty on predictions.

The differential entropy $m_{\mathrm{diffE}}$ of a DBU model reaches its maximum value for equally probable categorical distributions and thus, a on flat Dirichlet distribution. It is a measure for distributional uncertainty and expected to be low on ID data, but high on OOD data. 
\begin{equation}
\begin{aligned}
	m_{\mathrm{diffE}}  = &\sum_c^K \ln \Gamma (\alpha_c) - \ln \Gamma (\alpha_0) \\
	&- \sum_c^K (\alpha_c -1) \cdot (\Psi (\alpha_c) - \Psi (\alpha_0))
\end{aligned}
\end{equation}
where $\alpha$ are the parameters of the Dirichlet-distribution, $\Gamma$ is the Gamma function and $\Psi$ is the Digamma function.

The mutual information $m_{\mathrm{MI}}$ is the difference between the total uncertainty (entropy of the expected distribution) and the expected uncertainty on the data (expected entropy of the distribution). This uncertainty is expected to be low on ID data and high on OOD data. 
\begin{equation}
\begin{aligned}
	m_{\mathrm{MI}}  &= - \sum_{c=1}^{K} \frac{\alpha_c}{\alpha_0} \left( \ln \frac{\alpha_c}{\alpha_0} - \Psi(\alpha_c +1) + \Psi (\alpha_0 +1) \right)
\end{aligned}
\end{equation}

Furthermore, we use the precision~$\alpha_0$ to measure uncertainty, which is expected to be high on ID data and low on OOD data.
\begin{equation}
\begin{aligned}
	m_{\alpha_0}        &= \alpha_0 = \sum_{c=1}^{K} \alpha_c 
\end{aligned}
\end{equation}

As these uncertainty measures are computed in closed form and it is possible to obtain their gradients, we use them (i.e. $m_{\mathrm{diffE}}$, $m_{\mathrm{MI}}$, $m_{\alpha_0}$) are target function of our uncertainty attacks. Changing the attacked target function allows us to use a wide range of gradient-based attacks such as FGSM attacks, PGD attacks, but also more sophisticated attacks such as Carlini-Wagner attacks.

\subsection{Details of the Experimental setup}
\label{subsec:exp_setup}

\textbf{Models.} We trained all models with a similar based architecture. We used namely 3 linear layers for vector data sets, 3 convolutional layers with size of 5 + 3 linear layers for MNIST and the VGG16 \cite{vgg} architecture with batch normalization for CIFAR10. All the implementation are performed using Pytorch \citep{pytorch}. We optimized all models using Adam optimizer. We performed early stopping by checking for loss improvement every 2 epochs and a patience of 10. The models were trained on GPUs (1 TB SSD).

We performed a grid-search for hyper-parameters for all models. The learning rate grid search was done in $[1e^{-5}, 1e^{-3}]$. For \PostNet, we used Radial Flows with a depth of 6 and a latent space equal to 6. Further, we performed a grid search for the regularizing factor in $[1e^{-7}, 1e^{-4}]$. For \PriorNet, we performed a grid search for the OOD loss weight in $[1, 10]$. For \DDNet, we distilled the knowledge of $5$ neural networks after a grid search in $[2, 5, 10, 20]$ neural networks. Note that it already implied a significant overhead at training compare to other models.

\textbf{Metrics.} For all experiments, we focused on using AUC-PR scores since it is well suited to imbalance tasks \citep{imbalance_apr} while bringing theoretically similar information than AUC-ROC scores \citep{apr_auroc}. We scaled all scores from $[0, 1]$ to $[0, 100]$. All results are average over 5 training runs using the best hyper-parameters found after the grid search.

\textbf{Data sets.} For vector data sets, we use 5 different random splits to train all models. We split the data in training, validation and test sets ($60\%$, $20\%$, $20\%$). 

We use the segment vector data set \cite{uci_datasets}, where the goal is to classify areas of images into $7$ classes (window, foliage, grass, brickface, path, cement, sky). We remove class window from ID training data to provide OOD training data to \PriorNet. Further, We remove the class 'sky' from training and instead use it as the OOD data set for OOD detection experiments. Each input is composed of $18$ attributes describing the image area. The data set contains $2,310$ samples in total.

We further use the Sensorless Drive vector data set \cite{uci_datasets}, where the goal is to classify extracted motor current measurements into $11$ different classes. We remove class 9 from ID training data to provide OOD training data to \PriorNet. We remove classes 10 and 11 from training and use them as the OOD dataset for OOD detection experiments. Each input is composed of $49$ attributes describing motor behaviour. The data set contains $58,509$ samples in total.

Additionally, we use the MNIST image data set \cite{mnist} where the goal is to classify pictures of hand-drawn digits into $10$ classes (from digit $0$ to digit $9$). Each input is composed of a $1 \times 28 \times 28$ tensor. The data set contains $70,000$ samples. For OOD detection experiments, we use FashionMNIST \cite{fashionmnist} and KMNIST \cite{kmnist} containing images of Japanese characters and images of clothes, respectively. FashionMNIST was used as training OOD for \PriorNet while KMNIST is used as OOD at test time.

Finally, we use the CIFAR10 image data set \cite{cifar10} where the goal is to classify a picture of objects into $10$ classes (airplane, automobile, bird, cat, deer, dog, frog, horse, ship, truck). Each input is a $3 \times 32 \times 32$ tensor. The data set contains $60,000$ samples. For OOD detection experiments, we use street view house numbers (SVHN) \cite{svhn}  and CIFAR100 \citep{cifar10} containing images of numbers and objects respectively. CIFAR100 was used as training OOD for \PriorNet while SVHN is used as OOD at test time.
 
\textbf{Perturbations.} For all label and uncertainty attacks, we used Fast Gradient Sign Methods and Project Gradient Descent. We tried 6 different attack radii $[0.0, 0.1, 0.2, 0.5, 1.0, 2.0, 4.0]$. These radii operate on the input space after data normalization. We bound perturbations by~$L_{\infty}$-norm or by~$L_2$-norm, with 
\begin{equation}
\begin{aligned}
	L_{\infty} (x) = \max_{i=1,\dots, D} \left|x_i\right| \mathrm{~~~~and~~~~}
	L_2 (x)        = (\sum_{i=1}^{D} x_i^2)^{0.5}.
\end{aligned}
\end{equation}
For $L_{\infty}$-norm it is obvious how to relate perturbation size~$\varepsilon$ with perturbed input images, because all inputs are standardized such that the values of their features are between~$0$ and~$1$.
A perturbation of size~$\varepsilon=0$ corresponds to the original input, while a perturbation of size~$\varepsilon=1$ corresponds to the whole input space and allows to change all features to any value. 

For~$L_2$-norm the relation between perturbation size~$\varepsilon$ and perturbed input images is less obvious. To justify our choice for~$\varepsilon$ w.r.t. this norm, we relate perturbations size~$\varepsilon_2$ corresponding to $L_2$-norm with perturbations size~$\varepsilon_{\infty}$ corresponding to $L_{\infty}$-norm. 
First, we compute~$\varepsilon_2$, such that the $L_2$-norm is the smallest super-set of the $L_{\infty}$-norm. Let us consider a perturbation of~$\varepsilon_{\infty}$. The largest~$L_2$-norm would be obtained if each feature is perturbed by~$\varepsilon_{\infty}$. Thus, perturbation~$\varepsilon_2$, such that $L_2$ encloses~$L_{\infty}$ is $\varepsilon_2 = (\sum_{i=1}^{D} \varepsilon_{\infty}^2)^{0.5} = \sqrt{D} \varepsilon_{\infty}$. For the MNIST-data set, with $D=28 \times 28$ input features $L_2$-norm with $\varepsilon_2=28$ encloses $L_{\infty}$-norm with~$\varepsilon_{\infty}=1$. 

Alternatively, $\varepsilon_2$ can be computes such that the volume spanned by~$L_2$-norm is equivalent to the one spanned by~$L_{\infty}$-norm. Using that the volume spanned by $L_{\infty}$-norm is $\varepsilon_{\infty}^D$ and the volume spanned by $L_2$-norm is 
$\frac{\pi^{0.5 D} \varepsilon_2^D}{\Gamma(0.5 D +1)}$ (where $\Gamma$ is the Gamma-function), we obtain volume equivalence if 
$\varepsilon_2 = \Gamma(0.5 D +1)^{\frac{1}{D}} \sqrt{\pi} \varepsilon_{\infty}$. For the MNIST-data set, with $D=28 \times 28$ input features $L_2$-norm with $\varepsilon_2 \approx 21.39$ is volume equivalent to $L_{\infty}$-norm with~$\varepsilon_{\infty}=1$.

\newpage 
\subsection{Additional Experiments}

Table~\ref{tab:acc_label_attack} and~\ref{tab:acc_label_attack_fgsm} illustrate that no DBU model maintains high accuracy under gradient-based label attacks. Accuracy under PGD attacks decreases more than under FGSM attacks, since PGD is stronger.  Interestingly Noise attacks achieve also good performances with increasing Noise standard deviation. Note that the attack is not constraint to be with a given radius for Noise attacks.

\begin{table*}[htbp!]
 	\centering
 	\caption{Accuracy under PGD label attacks.}
 	\begin{small}
 		\begin{tabular}{@{}rrrrrrrrc|crrrrrrr@{}}
 			\toprule
  			Att. Rad. & 0.0 & 0.1 & 0.2 & 0.5 & 1.0 & 2.0 & 4.0 & & & 0.0 & 0.1 & 0.2 & 0.5 & 1.0 & 2.0 & 4.0 \\
 			\midrule
 			& \multicolumn{7}{c}{MNIST} & & & \multicolumn{7}{c}{CIFAR10} \\
 			\PostNet  &  \bf{99.4} &  \bf{99.2} &  \bf{98.8} &  96.8 &  89.6 &  53.8 &  13.0 & &
 			          &  89.5 &  73.5 &  51.7 &  13.2 &   2.2 &   0.8 &  0.3 \\
 			\PriorNet &  99.3 &  99.1 &  \bf{98.8} &  97.4 &  \bf{93.9} &  \bf{75.3} &   4.8 & &
 			          &  88.2 &  \bf{77.8} &  \bf{68.4} &  \bf{54.0} &  \bf{37.9} &  \bf{17.5} &  \bf{5.1} \\
 		    \DDNet    &  \bf{99.4} &  99.1 &  \bf{98.8} &  \bf{97.5} &  91.6 &  48.8 &   0.2 & &
 		              &  86.1 &  73.9 &  59.1 &  20.5 &   1.5 &   0.0 &  0.0 \\
 		    \EvNet    &  99.2 &  98.9 &  98.4 &  96.8 &  92.4 &  73.1 &  \bf{40.9} & &
 		              &  \bf{89.8} &  71.7 &  48.8 &  11.5 &   2.7 &   1.5 &  0.4 \\
 		    \midrule
 		 & \multicolumn{7}{c}{Sensorless} & & & \multicolumn{7}{c}{Segment} \\
 			\PostNet  &  98.3 &  13.1 &   6.4 &   4.0 &  \bf{7.0} &  \bf{9.8} &  \bf{11.3} & &
 			          &  98.9 &  82.8 &  \bf{50.1} &  \bf{19.2} &  \bf{8.8} &  \bf{5.1} &  \bf{8.6}   \\
 			\PriorNet &  \bf{99.3} &  16.5 &   5.6 &   1.2 &  0.4 &  0.2 &   1.6 & &
 			          &  \bf{99.5} &  90.7 &  47.6 &   7.8 &  0.2 &  0.0 &  0.4 \\
 		    \DDNet    &  \bf{99.3} &  12.4 &   2.4 &   0.6 &  0.3 &  0.1 &   0.1 & &
 		              &  99.2 &  \bf{90.8} &  45.7 &   6.9 &  0.0 &  0.0 &  0.0 \\
 		    \EvNet    &  99.0 &  \bf{35.3} &  \bf{22.3} &  \bf{11.2} &  \bf{7.0} &  5.2 &   4.0 & &
 		              &  99.3 &  91.8 &  54.0 &  10.3 &  0.8 &  0.5 &  0.6 \\
 			\bottomrule
 		\end{tabular}
 	\end{small}
 	\label{tab:acc_label_attack}
\end{table*}

\begin{table*}[htbp!]
 	\centering
 	\caption{Accuracy under FGSM label attacks.}
 	\begin{small}
 		\begin{tabular}{@{}rrrrrrrrc|crrrrrrr@{}}
 			\toprule
 			Att. Rad. & 0.0 & 0.1 & 0.2 & 0.5 & 1.0 & 2.0 & 4.0 & & & 0.0 & 0.1 & 0.2 & 0.5 & 1.0 & 2.0 & 4.0 \\
 			\midrule
 			& \multicolumn{7}{c}{MNIST} & & & \multicolumn{7}{c}{CIFAR10} \\
 			\PostNet  & \bf{99.4} &  \bf{99.2} &  \bf{98.9} &  97.7 &  95.2 &  \bf{90.1} &  \bf{79.2} & &
 			          & 89.5 &  72.3 &  54.9 &  31.2 &  21.0 &  16.8 &  15.6 \\
 			\PriorNet & 99.3 &  99.1 &  \bf{98.9} &  97.7 &  \bf{95.8} &  93.2 &  76.7 & &
 			          & 88.2 &  \bf{77.3} &  \bf{70.1} &  \bf{59.4} &  \bf{52.3} &  \bf{48.5} &  \bf{46.8} \\
 		    \DDNet    & \bf{99.4} &  \bf{99.2} &  \bf{98.9} &  \bf{97.8} &  94.7 &  79.2 &  25.2 & &
 			          & 86.1 &  73.0 &  60.2 &  32.5 &  14.6 &   7.1 &   6.0 \\
 		    \EvNet    & 99.2 &  98.9 &  98.6 &  97.6 &  \bf{95.8} &  \bf{90.1} &  74.4 & &
 			          & \bf{89.8} &  71.4 &  54.5 &  29.6 &  18.1 &  14.4 &  13.4 \\
 		    \midrule
 		     & \multicolumn{7}{c}{Sensorless} & & & \multicolumn{7}{c}{Segment} \\
 			\PostNet  & 98.3 &  19.6 &  10.9 &  10.9 &  11.9 &  12.4 &  12.5 & &
 			          & 98.9 &  79.6 &  \bf{57.3} &  \bf{31.5} &  \bf{18.4} &  \bf{20.6} &  \bf{19.9} \\
 			\PriorNet & \bf{99.3} &  24.7 &  11.8 &   8.6 &   8.5 &   8.1 &   8.3 & &
 			          & \bf{99.5} &  85.5 &  40.5 &   8.9 &   0.4 &   0.3 &   0.2 \\
 		    \DDNet    & \bf{99.3} &  18.0 &   8.2 &   6.5 &   5.4 &   6.7 &   7.8 & &
 			          & 99.2 &  86.4 &  36.2 &  11.9 &   0.9 &   0.0 &   0.0 \\
 		    \EvNet    & 99.0 &  \bf{42.0} &  \bf{28.0} &  \bf{17.5} &  \bf{13.7} &  \bf{13.6} &  \bf{14.9} & &
 			          & 99.3 &  \bf{90.6} &  55.2 &  14.2 &   2.4 &   0.5 &   0.1 \\
 			\bottomrule
 		\end{tabular}
 	\end{small}
 	\label{tab:acc_label_attack_fgsm}
\end{table*}

\begin{table*}[htbp!]
 	\centering
 	\caption{Accuracy under Noise label attacks.}
 	\begin{small}
 		\begin{tabular}{@{}rrrrrrrrc|crrrrrrr@{}}
 			\toprule
 			Noise Std & 0.0 & 0.1 & 0.2 & 0.5 & 1.0 & 2.0 & 4.0 & & & 0.0 & 0.1 & 0.2 & 0.5 & 1.0 & 2.0 & 4.0 \\
 			\midrule
 			& \multicolumn{7}{c}{MNIST} & & & \multicolumn{7}{c}{CIFAR10} \\
 			\PostNet  & \bf{99.4} &  \bf{98.6} &  91.8 &  \bf{14.9} &  \bf{1.3} &  \bf{0.1} &  0.0 & &
 			          & \bf{91.7} &  21.5 &  10.1 &   0.1 &   1.2 &  0.0 &  1.9 \\
 			\PriorNet & 99.3 &  98.5 &  \bf{95.7} &  14.4 &  0.0 &  0.0 &  0.0 & &
 			          & 87.7 &  \bf{28.1} &  \bf{11.2} &   9.7 &   5.0 &  \bf{8.5} &  \bf{9.0}\\
 		    \DDNet    & \bf{99.4} &  \bf{98.6} &  92.4 &  13.3 &  0.7 &  0.0 &  0.0 & &
 			          & 81.7 &  23.0 &  \bf{11.2} &  \bf{11.2} &  \bf{11.0} &  7.8 &  6.7 \\
 		    \EvNet    & 99.3 &  96.9 &  81.6 &  11.7 &  0.5 &  0.0 &  0.0 & &
 			          & 89.5 &  20.7 &  11.1 &   5.2 &   0.5 &  2.3 &  3.9 \\
 		    \midrule
 		     & \multicolumn{7}{c}{Sensorless} & & & \multicolumn{7}{c}{Segment} \\
 			\PostNet  & 98.1 &  0.1 &  \bf{3.7} &  \bf{11.7} &  \bf{11.7} &  \bf{11.7} &  \bf{11.7} & &
 			          & 98.5 &  39.4 &   3.9 &  \bf{1.8} &  \bf{12.1} &  \bf{20.3} &  \bf{22.1} \\
 			\PriorNet & \bf{99.3} &  0.2 &  0.0 &   0.0 &   0.0 &   0.3 &   2.4 & &
 			          & \bf{99.4} &  47.9 &   8.8 &  0.0 &   0.0 &   0.0 &   0.0 \\
 		    \DDNet    & 99.0 &  \bf{0.4} &  0.1 &   0.0 &   0.0 &   0.0 &   0.0 & &
 			          & 99.1 &  50.0 &  \bf{10.3} &  0.0 &   0.0 &   0.3 &   0.0 \\
 		    \EvNet    & 98.6 &  0.2 &  0.0 &   0.1 &   1.4 &   4.6 &   8.8 & &
 			          & 99.1 &  \bf{50.3} &  \bf{10.3} &  1.2 &   0.3 &   0.0 &   1.5 \\
 			\bottomrule
 		\end{tabular}
 	\end{small}
 	\label{tab:acc_label_attack_noise_attack}
\end{table*}

\clearpage
\subsubsection{Uncertainty estimation under label attacks}

\textbf{Is low uncertainty a reliable indicator of correct predictions?}

On non-perturbed data uncertainty estimates are an indicator of correctly classified samples, but if the input data is perturbed none of the DBU models maintains its high performance. Thus, uncertainty estimates are not a robust indicator of correctly labeled inputs. 

\begin{table*}[htbp!]
 	\centering
 	\caption{Distinguishing between correctly and wrongly predicted labels based on the differential entropy under PGD label attacks (AUC-PR).}
 	\begin{small}
 		\begin{tabular}{@{}rrrrrrrrc|crrrrrrr@{}}
 			\toprule
 			& \multicolumn{7}{c}{MNIST} & & & \multicolumn{7}{c}{Segment} \\
 			\cmidrule{2-8}  \cmidrule{11-16}
 			Att. Rad. & 0.0 & 0.1 & 0.2 & 0.5 & 1.0 & 2.0 & 4.0 & & & 0.0 & 0.1 & 0.2 & 0.5 & 1.0 & 2.0 & 4.0 \\
 			\midrule
 			\PostNet  &  99.9 &   99.9 &  99.8 &  98.7 &  89.5 &  43.5 &   9.0 & & 
 			          &  99.9 &  77.6 &  31.6 &  \bf{11.1} &  \bf{5.3} &  \bf{4.4} &   8.7 \\
 			\PriorNet &  99.9 &   99.8 &  99.6 &  97.7 &  90.5 &  \bf{69.1} &   6.4 & & 
 			          &  \bf{100.0} &  \bf{96.8} &  44.5 &   4.5 &  0.4 &  0.0 &  \bf{15.2} \\
 		    \DDNet    &  \bf{100.0} &  \bf{100.0} &  \bf{99.9} &  \bf{99.7} &  \bf{97.6} &  50.2 &   0.1 & &
 		              &  \bf{100.0} &  \bf{96.8} &  \bf{54.0} &   4.3 &  0.0 &  0.0 &   0.0 \\
 		    \EvNet    &  99.6 &   99.3 &  98.7 &  96.1 &  88.8 &  63.1 &  \bf{31.7} & &
 		              &  \bf{100.0} &  95.9 &  44.3 &   5.9 &  0.8 &  0.6 &   0.7 \\
 			\bottomrule
 		\end{tabular}
 	\end{small}
 	\label{tab:conf_label_attack_2}
\end{table*}

\begin{table*}[htbp!]
 	\centering
 	\caption{Distinguishing between correctly and wrongly predicted labels based on the precision~$\alpha_0$ under PGD label attacks (AUC-PR).}
 	\begin{small}
 		\begin{tabular}{@{}rrrrrrrrc|crrrrrrr@{}}
 			\toprule
 			Att. Rad. & 0.0 & 0.1 & 0.2 & 0.5 & 1.0 & 2.0 & 4.0 & & & 0.0 & 0.1 & 0.2 & 0.5 & 1.0 & 2.0 & 4.0 \\
 			\midrule
 			& \multicolumn{7}{c}{MNIST} & & & \multicolumn{7}{c}{CIFAR10} \\
            \PostNet  & \bf{100.0} &   99.9 &   99.7 &  98.2 &  87.9 &  39.1 &   6.9 & &
                      & \bf{98.7} &  88.6 &  56.2 &   7.8 &   1.2 &   0.4 &  0.3  \\
            \PriorNet &  99.9 &   99.8 &   99.6 &  97.7 &  90.4 & \bf{69.1} &   6.6  & &
                      &  92.9 &  77.7 &  60.5 & \bf{37.6} & \bf{24.9} & \bf{11.3} & \bf{3.0} \\
            \DDNet    & \bf{100.0} & \bf{100.0} & \bf{100.0} & \bf{99.8} & \bf{98.2} &  51.1 &   0.1  & &
                      &  97.6 & \bf{91.8} & \bf{78.3} &  18.1 &   0.8 &   0.0 &  0.0  \\
            \EvNet    &  99.6 &   99.2 &   98.6 &  95.7 &  88.6 &  63.6 & \bf{32.6} & &
                                  &  97.9 &  85.9 &  57.2 &  10.2 &   4.0 &   2.4 &  0.3  \\
 		    \midrule
 		     & \multicolumn{7}{c}{Sensorless} & & & \multicolumn{7}{c}{Segment} \\
            \PostNet  &  99.6 &   7.0 &   3.3 &  3.1 & \bf{6.9} & \bf{9.8} & \bf{11.3} & &
                      &  99.9 &  74.2 &  31.6 & \bf{11.1} & \bf{5.0} & \bf{4.2} & \bf{8.6} \\
            \PriorNet &  99.8 &  10.5 &   3.2 &  0.6 &  0.2 &  0.2 &   1.8 & &
                     & \bf{100.0} &  96.9 & \bf{45.2} &   4.4 &  0.4 &  0.0 &  1.2  \\
            \DDNet    &  99.8 &   8.7 &   1.3 &  0.3 &  0.2 &  0.1 &   0.2 & &
                    & \bf{100.0} & \bf{97.1} &  45.0 &   4.1 &  0.0 &  0.0 &  0.0 \\
            \EvNet    & \bf{99.9} & \bf{23.2} & \bf{13.2} & \bf{6.0} &  3.7 &  2.7 &   2.1 & &
                    & \bf{100.0} &  95.7 &  44.5 &   5.9 &  0.8 &  0.6 &  0.7 \\
 			\bottomrule
 		\end{tabular}
 	\end{small}
 	\label{tab:conf_label_attack_alpha}
\end{table*}

\begin{table*}[htbp!]
 	\centering
 	\caption{Distinguishing between correctly and wrongly predicted labels based on the mutual information under PGD label attacks (AUC-PR).}
 	\begin{small}
 		\begin{tabular}{@{}rrrrrrrrc|crrrrrrr@{}}
 			\toprule
 			Att. Rad. & 0.0 & 0.1 & 0.2 & 0.5 & 1.0 & 2.0 & 4.0 & & & 0.0 & 0.1 & 0.2 & 0.5 & 1.0 & 2.0 & 4.0 \\
 			\midrule
 			& \multicolumn{7}{c}{MNIST} & & & \multicolumn{7}{c}{CIFAR10} \\
            \PostNet  & 99.7 &  99.7 &  99.6 &  99.2 &  92.4 &  40.0 &   6.9 & &
                      & \bf{97.3} &  84.5 &  56.2 &  12.2 &   2.4 &   0.7 &  0.3  \\
            \PriorNet &  99.9 &  99.8 &  99.6 &  97.7 &  90.3 & \bf{68.9} &   6.4  & &
                      &  82.7 &  65.6 &  51.4 & \bf{35.5} & \bf{24.4} & \bf{11.0} & \bf{2.9} \\
            \DDNet    & \bf{100.0} & \bf{99.9} & \bf{99.9} & \bf{99.7} & \bf{97.4} &  50.2 &   0.1  & &
                      &  96.9 & \bf{90.8} & \bf{77.2} &  18.8 &   0.8 &   0.0 &  0.0  \\
            \EvNet    &  97.8 &  97.0 &  95.7 &  92.6 &  86.1 &  62.3 & \bf{28.9} & &
                      &  91.3 &  72.4 &  47.9 &  11.4 &   1.6 &   0.9 &  1.6  \\
 		    \midrule
 		     & \multicolumn{7}{c}{Sensorless} & & & \multicolumn{7}{c}{Segment} \\
            \PostNet  &  99.3 &   7.0 &   3.3 &  3.3 & \bf{7.0} & \bf{9.8} &  11.3 & &
                      &  99.9 &  73.2 &  31.5 & \bf{11.1} & \bf{5.0} & \bf{4.3} & \bf{8.7} \\
            \PriorNet & \bf{99.8} &  10.5 &   3.2 &  0.6 &  0.2 &  0.1 & \bf{11.8} & &
                      & \bf{100.0} & \bf{96.6} & \bf{45.2} &   4.5 &  0.4 &  0.0 &  1.1  \\
            \DDNet    &  99.6 &   8.6 &   1.3 &  0.3 &  0.2 &  0.1 &   0.1 & &
                      & \bf{100.0} &  96.5 &  42.4 &   4.1 &  0.0 &  0.0 &  0.0 \\
            \EvNet    &  99.1 & \bf{22.0} & \bf{12.6} & \bf{5.9} &  3.7 &  2.7 &   2.2 & &
                      & \bf{100.0} &  90.5 &  41.0 &   5.9 &  0.8 &  0.6 &  0.7 \\
 			\bottomrule
 		\end{tabular}
 	\end{small}
 	\label{tab:conf_label_attack_mi}
\end{table*}

\begin{table*}[htbp!]
 	\centering
 	\caption{Distinguishing between correctly and wrongly predicted labels based on the differential entropy under FGSM label attacks (AUC-PR).}
 	\begin{small}
 		\begin{tabular}{@{}rrrrrrrrc|crrrrrrr@{}}
 			\toprule
 			Att. Rad. & 0.0 & 0.1 & 0.2 & 0.5 & 1.0 & 2.0 & 4.0 & & & 0.0 & 0.1 & 0.2 & 0.5 & 1.0 & 2.0 & 4.0 \\
 			\midrule
 			& \multicolumn{7}{c}{MNIST} & & & \multicolumn{7}{c}{CIFAR10} \\
             \PostNet  & 99.9 &   99.9 &  99.8 &  99.4 &  97.8 & \bf{92.1} & \bf{83.2} & &
                      & \bf{98.5} &  88.7 &  68.9 &  31.0 &  18.6 &  15.5 &  16.7 \\
            \PriorNet & 99.9 &   99.9 &  99.7 &  98.3 &  94.1 &  88.5 &  78.6  & &
                      & 90.1 &  73.6 &  61.6 & \bf{46.1} & \bf{38.5} & \bf{35.6} & \bf{37.3} \\
            \DDNet    & \bf{100.0} & \bf{100.0} & \bf{99.9} & \bf{99.8} & \bf{98.7} &  86.4 &  23.0 & &
                      & 97.3 & \bf{90.6} & \bf{78.7} &  39.4 &  13.7 &   6.0 &   5.1 \\
            \EvNet    & 99.6 &   99.4 &  99.1 &  97.8 &  95.8 &  90.4 &  76.8 & &
                      & 98.0 &  86.2 &  67.4 &  32.7 &  19.9 &  18.2 &  19.7 \\		
 		    \midrule
 		     & \multicolumn{7}{c}{Sensorless} & & & \multicolumn{7}{c}{Segment} \\
             \PostNet  & 99.7 &  11.7 &   7.3 &   9.3 &  11.8 &  12.5 &  12.5 & &
                      & 99.9 &  73.6 &  40.6 & \bf{23.7} & \bf{17.2} & \bf{19.8} & \bf{20.2} \\
            \PriorNet & 99.8 &  21.4 &  10.4 &   8.5 &   9.0 &   9.2 &  10.3 & &
                      & \bf{100.0} &  93.7 &  37.7 &   5.8 &   1.1 &   0.9 &   0.8 \\
            \DDNet    & 99.7 &  18.5 &   5.4 &   4.3 &   4.2 &   5.7 &   7.9 & &
                      & \bf{100.0} & \bf{94.1} &  42.9 &   7.2 &   1.0 &   0.0 &   0.0 \\
            \EvNet    & \bf{99.9} & \bf{44.8} & \bf{29.2} & \bf{18.2} & \bf{15.1} & \bf{14.9} & \bf{15.5} & &
                      & \bf{100.0} &  93.7 & \bf{48.7} &   8.7 &   2.4 &   1.6 &   0.5 \\
 			\bottomrule
 		\end{tabular}
 	\end{small}
 	\label{tab:conf_label_attack_fgsm}
\end{table*}

\begin{table*}[htbp!]
 	\centering
 	\caption{Distinguishing between correctly and wrongly predicted labels based on the differential entropy under Noise label attacks (AUC-PR).}
 	\begin{small}
 		\begin{tabular}{@{}rrrrrrrrc|crrrrrrr@{}}
 			\toprule
 			Noise Std & 0.0 & 0.1 & 0.2 & 0.5 & 1.0 & 2.0 & 4.0 & & & 0.0 & 0.1 & 0.2 & 0.5 & 1.0 & 2.0 & 4.0 \\
 			\midrule
 			& \multicolumn{7}{c}{MNIST} & & & \multicolumn{7}{c}{CIFAR10} \\
             \PostNet  & 99.9 &  99.8 &  99.6 &  \bf{74.2} &  \bf{7.4} &  \bf{0.2} &  0.0 & &
                      & \bf{98.}7 &  \bf{76.}3 &  24.3 &   0.4 &   4.9 &  0.0 &  1.7 \\
            \PriorNet & 99.9 &  99.9 &  \bf{99.8} &  73.4 &  0.0 &  0.0 &  0.0  & &
                      & 85.0 &  27.8 &  15.9 &  \bf{20.}4 &   7.0 &  \bf{7.}7 &  \bf{8.3} \\
            \DDNet    & \bf{100.0} &  \bf{99.9} &  99.4 &  51.1 &  0.6 &  0.1 &  0.0 & &
                      & 96.1 &  61.0 &  \bf{39.}8 &  14.2 &  \bf{11.}3 &  6.9 &  6.9 \\
            \EvNet    & 99.5 &  98.4 &  88.5 &  20.2 &  0.9 &  0.0 &  0.0  & &
                      & 97.5 &  66.1 &  21.4 &   7.7 &   2.3 &  3.0 &  3.8 \\		
 		    \midrule
 		     & \multicolumn{7}{c}{Sensorless} & & & \multicolumn{7}{c}{Segment} \\
             \PostNet  & 99.7 &  0.3 &  \bf{3.2} &  \bf{13.3} &  \bf{12.0} &  \bf{11.7} &  \bf{11.7} & &
                      & 99.9 &  53.9 &   4.8 &  1.8 &  \bf{11.2} &  \bf{21.7} &  \bf{21.6} \\
            \PriorNet & \bf{100.0} &  0.3 &  0.0 &   0.0 &   0.0 &  7.8 &  11.5 & &
                      & \bf{100.0} &  \bf{84.5} &  15.6 &  0.0 &   0.0 &   0.0 &   0.0 \\
            \DDNet    & 99.7 &  \bf{0.9} &  0.6 &   0.0 &   0.0 &   0.0 &   0.0 & &
                      & \bf{100.0} &  82.7 &  \bf{23.9} &  0.0 &   0.0 &   0.6 &   0.0 \\
            \EvNet    & 99.8 &  0.3 &  0.0 &   0.1 &   1.7 &   5.5 &  10.0 & &
                      & \bf{100.0} &  78.3 &  19.0 &  \bf{3.5} &   0.5 &   0.0 &   1.7 \\
 			\bottomrule
 		\end{tabular}
 	\end{small}
 	\label{tab:conf_label_attack_noise_attack}
\end{table*}

Table~\ref{tab:conf_label_attack}, \ref{tab:conf_label_attack_2}, \ref{tab:conf_label_attack_alpha}, and~\ref{tab:conf_label_attack_mi} illustrate that neither differential entropy nor precision, nor mutual information are a reliable indicator of correct predictions under PGD attacks. 
DBU-models achieve significantly better results when they are attacked by FGSM-attacks (Table~\ref{tab:conf_label_attack_fgsm}), but as FGSM attacks provide much weaker adversarial examples than PGD attacks, this cannot be seen as real advantage.

\clearpage
\textbf{Can we use uncertainty estimates to detect attacks against the class prediction?}

PGD attacks do not explicitly consider uncertainty during the computation of adversarial examples, but they seem to provide perturbed inputs with similar uncertainty as the original input.

\begin{table*}[htbp!]
 	\centering
 	\caption{Attack-Detection based on differential entropy under PGD label attacks (AUC-PR).}
 	\begin{small}
\begin{tabular}{@{}rrrrrrrc|crrrrrr@{}}
 			\toprule
 		     & \multicolumn{6}{c}{MNIST} & & & \multicolumn{6}{c}{Segment} \\
 			\cmidrule{2-7}  \cmidrule{10-15}
 			Att. Rad. & 0.1 & 0.2 & 0.5 & 1.0 & 2.0 & 4.0 & & & 0.1 & 0.2 & 0.5 & 1.0 & 2.0 & 4.0 \\
 			\midrule
            \PostNet  &  57.7 &  66.3 &  83.4 &  90.5 &  79.0 &  50.1 & & 
                      & \bf{95.6} &  73.5 & \bf{47.0} & \bf{42.3} & \bf{53.4} & \bf{82.7} \\
            \PriorNet & \bf{67.7} & \bf{83.2} & \bf{97.1} & \bf{96.7} &  92.1 &  82.9 & & 
                      &  86.7 &  83.3 &  38.0 &  31.3 &  30.8 &  31.5 \\
            \DDNet    &  53.4 &  57.1 &  68.5 &  83.9 & \bf{96.0} & \bf{86.3} & & 
                      &  76.1 & \bf{83.5} &  45.4 &  32.4 &  30.8 &  30.8 \\
            \EvNet    &  54.8 &  59.0 &  68.5 &  75.9 &  72.6 &  59.8 & & 
                      &  94.9 &  80.9 &  41.5 &  32.5 &  31.1 &  31.1 \\
 			\bottomrule 			
 		\end{tabular}
 	\end{small}
 	\label{tab:label_attack_detect_2}
\end{table*}

\begin{table*}[htbp!]
 	\centering
 	\caption{Attack-Detection based on precision $\alpha_0$ under PGD label attacks (AUC-PR).}
 	\begin{small}
 		\begin{tabular}{@{}rrrrrrrc|crrrrrr@{}}
 			\toprule
 			Att. Rad. & 0.1 & 0.2 & 0.5 & 1.0 & 2.0 & 4.0 & & & 0.1 & 0.2 & 0.5 & 1.0 & 2.0 & 4.0 \\
 			\midrule
 		     & \multicolumn{6}{c}{MNIST} & & & \multicolumn{6}{c}{CIFAR10} \\
            \PostNet  &  63.3 &  75.7 &  92.6 &  95.1 &  75.3 &  39.5 & & 
                      & \bf{63.4} & \bf{66.9} &  42.1 &  32.9 &  31.6 &  31.2 \\
            \PriorNet & \bf{67.6} & \bf{83.2} & \bf{97.1} & \bf{96.9} & \bf{92.7} & \bf{84.7} & & 
                      &  53.3 &  56.0 &  55.6 & \bf{49.2} &  42.2 &  35.4 \\
            \DDNet    &  52.7 &  55.7 &  64.7 &  78.4 &  91.9 &  80.9 & & 
                      &  55.8 &  60.5 & \bf{57.3} &  38.7 &  32.3 &  31.4  \\
            \EvNet    &  49.1 &  48.0 &  45.1 &  42.7 &  41.8 &  39.2 & & 
                      &  48.4 &  46.9 &  46.3 &  46.3 & \bf{44.5} & \bf{42.5} \\
 		    \midrule
 		  	& \multicolumn{6}{c}{Sensorless} & & & \multicolumn{6}{c}{Segment} \\
            \PostNet  &  39.8 &  35.8 &  35.4 & \bf{52.0} & \bf{88.2} & \bf{99.0} & & 
                      & \bf{94.6} &  70.3 & \bf{46.3} & \bf{42.6} & \bf{54.9} & \bf{84.0} \\
            \PriorNet &  40.9 &  35.1 &  32.0 &  31.1 &  30.7 &  30.7 & & 
                      &  82.7 &  82.6 &  39.4 &  31.6 &  30.8 &  30.8 \\
            \DDNet    & \bf{47.7} & \bf{40.3} &  35.3 &  32.8 &  31.3 &  30.8 & & 
                      &  80.0 & \bf{86.0} &  43.3 &  33.6 &  31.0 &  30.8 \\
            \EvNet    &  45.4 &  39.7 & \bf{36.1} &  34.8 &  34.7 &  36.0 & & 
                      &  90.9 &  72.4 &  40.4 &  32.4 &  31.1 &  31.1 \\
             			\bottomrule
 		\end{tabular}
 	\end{small}
 	\label{tab:label_attack_detect_auroc_1}
\end{table*}

\begin{table*}[htbp!]
 	\centering
 	\caption{Attack-Detection based on mutual information under PGD label attacks (AUC-PR).}
 	\begin{small}
 		\begin{tabular}{@{}rrrrrrrc|crrrrrr@{}}
 			\toprule
 			Att. Rad. & 0.1 & 0.2 & 0.5 & 1.0 & 2.0 & 4.0 & & & 0.1 & 0.2 & 0.5 & 1.0 & 2.0 & 4.0 \\
 			\midrule
 		     & \multicolumn{6}{c}{MNIST} & & & \multicolumn{6}{c}{CIFAR10} \\
 		    \PostNet  &   42.2 &  37.5 &  36.7 &  54.5 &  70.5 &  70.3 & & 
                        &   52.2 &  52.1 &  50.0 & \bf{65.9} & \bf{76.3} & \bf{80.7} \\
            \PriorNet & \bf{67.7} & \bf{83.3} & \bf{97.1} & \bf{96.9} &  92.6 & \bf{84.5} & & 
                      &   54.0 &  56.9 &  56.3 &  49.7 &  42.4 &  35.5  \\
            \DDNet    &   53.1 &  56.3 &  66.5 &  81.0 & \bf{94.0} &  82.9 & & 
                      & \bf{56.0} & \bf{60.8} & \bf{57.4} &  38.2 &  32.1 &  31.3  \\
            \EvNet    &   49.1 &  48.0 &  45.2 &  42.9 &  41.9 &  39.3 & & 
                      &   48.7 &  47.3 &  46.3 &  46.0 &  44.1 &  42.2 \\
 		    \midrule
 		  	& \multicolumn{6}{c}{Sensorless} & & & \multicolumn{6}{c}{Segment} \\
            \PostNet  & \bf{75.3} & \bf{76.6} & \bf{66.5} & \bf{57.7} & \bf{85.6} & \bf{98.7} & & 
                      & \bf{94.8} &  73.5 & \bf{55.9} & \bf{47.9} & \bf{58.0} & \bf{84.0} \\
            \PriorNet &  40.7 &  35.0 &  32.0 &  31.0 &  30.7 &  30.7 & & 
                      &  83.5 &  82.7 &  39.2 &  31.6 &  30.8 &  30.8 \\
            \DDNet    &  48.0 &  40.0 &  35.2 &  32.6 &  31.2 &  30.8 & & 
                      &  82.4 & \bf{88.1} &  43.4 &  33.4 &  30.9 &  30.8 \\
            \EvNet    &  45.5 &  39.7 &  36.1 &  34.8 &  34.7 &  36.0 & & 
                      &  91.7 &  72.9 &  40.5 &  32.4 &  31.1 &  31.1  \\ 			
 			\bottomrule
 		\end{tabular}
 	\end{small}
 	\label{tab:label_attack_detect_auroc_2}
\end{table*}

FGSM and Noise attacks are easier to detect, but also weaker thand PGD attacks. This suggests that DBU models are capable of detecting weak attacks by using uncertainty estimation.

\begin{table*}[htbp!]
 	\centering
 	\caption{Attack-Detection based on differential entropy under FGSM label attacks (AUC-PR).}
 	\begin{small}
 		\begin{tabular}{@{}rrrrrrrc|crrrrrr@{}}
 			\toprule
 			Att. Rad. & 0.1 & 0.2 & 0.5 & 1.0 & 2.0 & 4.0 & & & 0.1 & 0.2 & 0.5 & 1.0 & 2.0 & 4.0 \\
 			\midrule
 		     & \multicolumn{6}{c}{MNIST} & & & \multicolumn{6}{c}{CIFAR10} \\
            \PostNet  &   55.9 &  61.8 &  74.8 &  84.0 &  88.9 &  89.9 & & 
                      & \bf{62.1} & \bf{67.2} &  65.7 &  63.1 &  65.4 &  73.8  \\
            \PriorNet & \bf{67.4} & \bf{82.4} & \bf{96.9} & \bf{98.3} & \bf{98.9} & \bf{99.6} & & 
                      &   58.4 &  63.1 &  68.5 & \bf{70.1} &  68.5 &  62.5  \\
            \DDNet    &   53.6 &  57.3 &  68.3 &  82.6 &  95.6 &  98.7 & & 
                      &   57.2 &  62.9 & \bf{69.1} &  68.7 & \bf{69.7} & \bf{76.5} \\
            \EvNet    &   54.1 &  57.4 &  63.8 &  67.6 &  68.6 &  69.9 & & 
                      &   57.8 &  61.7 &  63.3 &  62.9 &  65.7 &  72.5 \\ 		    
 		    \midrule
 		  	& \multicolumn{6}{c}{Sensorless} & & & \multicolumn{6}{c}{Segment} \\
            \PostNet  & \bf{98.4} & \bf{99.8} & \bf{99.9} & \bf{99.9} & \bf{99.9} & \bf{99.9} & & 
                      & \bf{96.9} & \bf{93.9} & \bf{99.5} & \bf{99.9} & \bf{100.0} & \bf{100.0} \\
            \PriorNet &  48.7 &  38.6 &  32.7 &  32.9 &  38.6 &  44.3 & & 
                      &  89.0 &  80.8 &  46.7 &  37.2 &   33.7 &   32.4 \\
            \DDNet    &  61.5 &  47.8 &  37.1 &  33.1 &  32.4 &  33.2 & & 
                      &  79.6 &  86.2 &  60.2 &  47.5 &   36.6 &   31.6 \\
            \EvNet    &  67.3 &  65.5 &  72.3 &  73.4 &  75.3 &  79.1 & & 
                      &  95.7 &  87.2 &  59.3 &  51.7 &   51.1 &   53.5 \\
 			\bottomrule
 		\end{tabular}
 	\end{small}
 	\label{tab:label_attack_detect_auroc_3}
\end{table*}

\begin{table*}[htbp!]
 	\centering
 	\caption{Attack-Detection based on differential entropy under Noise label attacks (AUC-PR).}
 	\begin{small}
 		\begin{tabular}{@{}rrrrrrrc|crrrrrr@{}}
 			\toprule
 			Noise Std. & 0.1 & 0.2 & 0.5 & 1.0 & 2.0 & 4.0 & & & 0.1 & 0.2 & 0.5 & 1.0 & 2.0 & 4.0 \\
 			\midrule
 		     & \multicolumn{6}{c}{MNIST} & & & \multicolumn{6}{c}{CIFAR10} \\
            \PostNet  &   51.3 &  65.3 &  93.8 &   95.1 &  95.2 &  95.2 & & 
                      & \bf{80.8} &  \bf{84.5} &  \bf{97.6} &  \bf{99.5} &  99.3 &   98.2  \\
            \PriorNet & 32.5 &  36.8 &  88.9 &   99.6 &  99.7 &  92.7 & & 
                      &   34.7 &  32.3 &  34.3 &  60.3 &  95.5 &  \bf{100.0}  \\
            \DDNet    &   \bf{60.7} &  \bf{87.6} &  \bf{99.8} &  \bf{100.0} &  \bf{99.9} &  \bf{99.8} & & 
                      &   59.1 &  62.6 &  81.5 &  98.6 &  \bf{99.8} &   98.7 \\
            \EvNet    &   51.2 &  55.7 &  66.9 &   70.3 &  68.0 &  67.1 & & 
                      &   75.7 &  78.6 &  88.2 &  97.8 &  96.4 &   95.6 \\ 		    
 		    \midrule
 		  	& \multicolumn{6}{c}{Sensorless} & & & \multicolumn{6}{c}{Segment} \\
            \PostNet  & \bf{99.8} &  \bf{100.0} &  \bf{100.0} &  \bf{100.0} &  \bf{100.0} &  \bf{100.0} & & 
                      &  \bf{95.6} &  \bf{99.4} &  \bf{100.0} &  \bf{100.0} &  \bf{100.0} &  \bf{100.0} \\
            \PriorNet & 42.0 &   33.8 &   31.5 &   34.7 &   43.7 &   47.0 & & 
                      & 56.7 &  56.7 &   39.8 &   33.7 &   31.9 &   33.7 \\
            \DDNet    &  53.4 &   43.5 &   34.3 &   31.6 &   32.5 &   36.1 & & 
                      &  57.0 &  58.9 &   43.1 &   33.7 &   31.5 &   31.3 \\
            \EvNet    &  67.1 &   78.8 &   88.3 &   95.4 &   96.9 &   97.8 & & 
                      &  60.8 &  63.5 &   61.2 &   64.8 &   73.7 &   85.2 \\
 			\bottomrule
 		\end{tabular}
 	\end{small}
 	\label{tab:label_attack_detect_auroc_4}
\end{table*}

\clearpage
\subsubsection{Attacking uncertainty estimation}

\textbf{Are uncertainty estimates a robust feature for OOD detection?}

Using uncertainty estimation to distinguish between ID and OOD data is not robust as shown in the following tables.

 \begin{table*}[htbp!]
 	\centering
 	\caption{OOD detection based on differential entropy under PGD uncertainty attacks against differential entropy on ID data and OOD data (AUC-PR).}
 	\begin{small}
 		\begin{tabular}{@{}rrrrrrrrc|crrrrrrr@{}}
 			\toprule
 			& \multicolumn{7}{c}{ID-Attack (non-attacked OOD)} &  & &  \multicolumn{7}{c}{OOD-Attack (non-attacked ID)} \\
 			\cmidrule{2-8}  \cmidrule{11-17}
 			Att. Rad. & 0.0 & 0.1 & 0.2 & 0.5 & 1.0 & 2.0 & 4.0 & & &
 			            0.0 & 0.1 & 0.2 & 0.5 & 1.0 & 2.0 & 4.0 \\
 			\midrule
 			& \multicolumn{16}{c}{\textbf{MNIST -- KMNIST}} \\
            \PostNet  & 94.5 &  94.1 &  93.9 &  91.1 &  77.1 &  44.0 &  31.9 & &
                      & 94.5 &  93.1 &  91.4 &  82.1 &  62.2 &  50.7 &  48.8 \\
            \PriorNet & \bf{99.6} & \bf{99.4} & \bf{99.1} & \bf{97.8} & \bf{93.8} & \bf{77.6} &  \bf{32.0} & &
                      & \bf{99.6} & \bf{99.4} & \bf{99.1} &  98.0 &  94.6 &  85.5 &  \bf{73.9} \\
            \DDNet    & 99.3 &  99.1 &  98.9 & \bf{97.8} &  93.5 &  63.3 &  30.7 & &
                      & 99.3 &  99.1 &  99.0 & \bf{98.3} & \bf{96.7} & \bf{91.3} &  73.8  \\
            \EvNet    & 69.0 &  67.1 &  65.6 &  61.8 &  57.4 &  50.9 &  43.6 & &
                      & 69.0 &  55.8 &  48.0 &  39.4 &  36.2 &  34.9 &  34.4  \\
 			\midrule
 			& \multicolumn{16}{c}{\textbf{Seg. -- Seg. class sky}} \\
            \PostNet  & \bf{99.0} & \bf{80.7} & \bf{53.5} & \bf{38.0} & \bf{34.0} & \bf{41.6} &  \bf{49.5} & &
                      & \bf{99.0} & \bf{88.4} &  69.2 &  45.1 & \bf{36.4} & \bf{42.6} &  \bf{75.4} \\
            \PriorNet & 34.8 &  31.4 &  30.9 &  30.8 &  30.8 &  30.8 &  30.8 & &
                      & 34.8 &  31.8 &  31.0 &  30.8 &  30.8 &  30.8 &  32.1 \\
            \DDNet    & 31.5 &  30.9 &  30.8 &  30.8 &  30.8 &  30.8 &  30.8 & &
                      & 31.5 &  31.0 &  30.8 &  30.8 &  30.8 &  30.8 &  30.8  \\
            \EvNet    & 92.5 &  67.2 &  43.2 &  31.6 &  30.9 &  30.9 &  31.2 & &
                      & 92.5 &  86.1 & \bf{82.7} & \bf{48.9} &  32.7 &  30.9 &  30.9  \\
 			\bottomrule
 		\end{tabular}
 	\end{small}
 	\label{tab:id_ood_attacks_part2}
\end{table*}

 \begin{table*}[htbp!]
 	\centering
 	\caption{OOD detection under PGD uncertainty attacks against differential entropy on ID data and OOD data (AUC-ROC).}
 	\begin{small}
 		\begin{tabular}{@{}rrrrrrrrc|crrrrrrr@{}}
 			\toprule
 			& \multicolumn{7}{c}{ID-Attack (non-attacked OOD)} &  & &  \multicolumn{7}{c}{OOD-Attack (non-attacked ID)} \\
 			\cmidrule{2-8}  \cmidrule{11-17}
 			Att. Rad. & 0.0 & 0.1 & 0.2 & 0.5 & 1.0 & 2.0 & 4.0 & & &
 			            0.0 & 0.1 & 0.2 & 0.5 & 1.0 & 2.0 & 4.0 \\
 			\midrule
 			& \multicolumn{16}{c}{\textbf{MNIST -- KMNIST}} \\
            \PostNet  & 91.6 &  91.3 &  91.9 &  91.5 &  80.2 &  38.8 &   9.2 & &
                      & 91.6 &  90.4 &  89.0 &  81.6 &  62.6 &  45.0 &  43.1 \\
            \PriorNet & \bf{99.8} & \bf{99.7} & \bf{99.5} & \bf{99.0} & \bf{97.1} & \bf{81.1} &   8.7 & &
                      & \bf{99.8} & \bf{99.7} & \bf{99.6} & \bf{99.1} & \bf{97.7} & \bf{93.0} &  \bf{84.9} \\
            \DDNet    & 99.2 &  98.9 &  98.6 &  97.3 &  92.1 &  58.2 &   1.2 & &
                      & 99.2 &  99.0 &  98.8 &  97.9 &  95.8 &  89.1 &  69.3 \\
            \EvNet    & 81.2 &  79.6 &  78.2 &  74.6 &  69.5 &  58.7 &  \bf{43.0} & &
                      & 81.2 &  67.2 &  54.8 &  35.4 &  25.5 &  20.7 &  18.5 \\
 			\midrule
 			& \multicolumn{16}{c}{\textbf{CIFAR10 -- SVHN}} \\
\bottomrule
 		\end{tabular}
 	\end{small}
 	\label{tab:id_ood_attacks_measure_diffe_auroc}
\end{table*}

 \begin{table*}[htbp!]
 	\centering
 	\caption{OOD detection (AU-PR) under PGD uncertainty attacks against precision~$\alpha_0$ on ID data and OOD data.}
 	\begin{small}
 		\begin{tabular}{@{}rrrrrrrrc|crrrrrrr@{}}
 			\toprule
 			& \multicolumn{7}{c}{ID-Attack (non-attacked OOD)} &  & &  \multicolumn{7}{c}{OOD-Attack (non-attacked ID)} \\
 			\cmidrule{2-8}  \cmidrule{11-17}
 			Att. Rad. & 0.0 & 0.1 & 0.2 & 0.5 & 1.0 & 2.0 & 4.0 & & &
 			            0.0 & 0.1 & 0.2 & 0.5 & 1.0 & 2.0 & 4.0 \\
 			\midrule
 			& \multicolumn{16}{c}{\textbf{MNIST -- KMNIST}} \\
            \PostNet  & 98.4 &  97.4 &  96.0 &  88.8 &  70.9 &  39.3 &  31.3 & &
                      & 98.4 &  97.2 &  95.2 &  82.8 &  52.6 &  34.3 &  32.1 \\
            \PriorNet & \bf{99.6} & \bf{99.5} & \bf{99.2} & \bf{98.0} & \bf{94.1} & \bf{76.0} &  31.1 & &
                      & \bf{99.6} & \bf{99.5} & \bf{99.2} & \bf{98.2} & \bf{95.3} & \bf{87.5} & \bf{75.6} \\
            \DDNet    & 97.2 &  96.7 &  96.1 &  93.8 &  86.4 &  53.2 &  31.0 & &
                      & 97.2 &  96.7 &  96.2 &  94.5 &  91.1 &  82.9 &  64.6 \\
            \EvNet    & 39.8 &  39.2 &  38.8 &  37.9 &  37.1 &  36.3 & \bf{35.4} & &
                      & 39.8 &  34.5 &  32.5 &  31.2 &  31.0 &  30.9 &  31.0 \\
 			\midrule
 			& \multicolumn{16}{c}{\textbf{CIFAR10 -- SVHN}} \\
\bottomrule
 		\end{tabular}
 	\end{small}
 	\label{tab:id_ood_attacks_measure_alpha0_aupr}
\end{table*}

 \begin{table*}[htbp!]
 	\centering
 	\caption{OOD detection (AUC-ROC) under PGD uncertainty attacks against precision~$\alpha_0$ on ID data and OOD data.}
 	\begin{small}
 		\begin{tabular}{@{}rrrrrrrrc|crrrrrrr@{}}
 			\toprule
 			& \multicolumn{7}{c}{ID-Attack (non-attacked OOD)} &  & &  \multicolumn{7}{c}{OOD-Attack (non-attacked ID)} \\
 			\cmidrule{2-8}  \cmidrule{11-17}
 			Att. Rad. & 0.0 & 0.1 & 0.2 & 0.5 & 1.0 & 2.0 & 4.0 & & &
 			            0.0 & 0.1 & 0.2 & 0.5 & 1.0 & 2.0 & 4.0 \\
 			\midrule
 			& \multicolumn{16}{c}{\textbf{MNIST -- KMNIST}} \\
            \PostNet  & 98.4 &  97.6 &  96.4 &  90.9 &  74.0 &  28.9 &   6.3 & &
                      & 98.4 &  97.6 &  96.3 &  89.0 &  61.3 &  19.6 &   9.7 \\
            \PriorNet & \bf{99.8} & \bf{99.7} & \bf{99.6} & \bf{99.1} & \bf{97.2} & \bf{79.4} &   4.4 & &
                      & \bf{99.8} & \bf{99.7} & \bf{99.6} & \bf{99.2} & \bf{98.0} & \bf{93.9} & \bf{85.8} \\
            \DDNet    & 96.5 &  95.9 &  95.1 &  92.0 &  82.6 &  44.3 &   3.5 & &
                      & 96.5 &  95.9 &  95.2 &  92.9 &  88.6 &  78.7 &  59.4 \\
            \EvNet    & 35.9 &  34.1 &  32.8 &  30.1 &  27.4 &  24.6 & \bf{21.4} & &
                      & 35.9 &  18.7 &  10.4 &   3.7 &   2.0 &   1.7 &   2.0 \\
 			\midrule
 			& \multicolumn{16}{c}{\textbf{CIFAR10 -- SVHN}} \\
\bottomrule
 		\end{tabular}
 	\end{small}
 	\label{tab:id_ood_attacks_measure_alpha0_auroc}
\end{table*}

 \begin{table*}[htbp!]
 	\centering
 	\caption{OOD detection (AU-PR) under PGD uncertainty attacks against distributional uncertainty on ID data and OOD data.}
 	\begin{small}
 		\begin{tabular}{@{}rrrrrrrrc|crrrrrrr@{}}
 			\toprule
 			& \multicolumn{7}{c}{ID-Attack (non-attacked OOD)} &  & &  \multicolumn{7}{c}{OOD-Attack (non-attacked ID)} \\
 			\cmidrule{2-8}  \cmidrule{11-17}
 			Att. Rad. & 0.0 & 0.1 & 0.2 & 0.5 & 1.0 & 2.0 & 4.0 & & &
 			            0.0 & 0.1 & 0.2 & 0.5 & 1.0 & 2.0 & 4.0 \\
 			\midrule
 			& \multicolumn{16}{c}{\textbf{MNIST -- KMNIST}} \\
            \PostNet  & 80.5 &  76.2 &  73.4 &  69.1 &  66.6 &  65.4 & \bf{60.2} & &
                      & 80.5 &  72.1 &  63.9 &  43.9 &  33.0 &  30.9 &  30.8 \\
            \PriorNet & \bf{99.6} & \bf{99.4} & \bf{99.2} & \bf{98.0} & \bf{94.1} & \bf{76.3} &  31.2 & &
                      & \bf{99.6} & \bf{99.4} & \bf{99.2} & \bf{98.2} & \bf{95.2} & \bf{87.2} & \bf{75.2} \\
            \DDNet    & 98.4 &  98.1 &  97.7 &  95.8 &  89.5 &  56.2 &  30.9 & &
                      & 98.4 &  98.1 &  97.8 &  96.5 &  93.8 &  86.3 &  67.7 \\
            \EvNet    & 40.1 &  39.5 &  39.1 &  38.2 &  37.3 &  36.5 &  35.6 & &
                      & 40.1 &  34.6 &  32.6 &  31.3 &  31.0 &  31.0 &  31.1 \\
 			\midrule
 			& \multicolumn{16}{c}{\textbf{CIFAR10 -- SVHN}} \\
\bottomrule
 		\end{tabular}
 	\end{small}
 	\label{tab:id_ood_attacks_measure_distU_aupr}
\end{table*}

 \begin{table*}[htbp!]
 	\centering
 	\caption{OOD detection (AUC-ROC) under PGD uncertainty attacks against distributional uncertainty on ID data and OOD data.}
 	\begin{small}
 		\begin{tabular}{@{}rrrrrrrrc|crrrrrrr@{}}
 			\toprule
 			& \multicolumn{7}{c}{ID-Attack (non-attacked OOD)} &  & &  \multicolumn{7}{c}{OOD-Attack (non-attacked ID)} \\
 			\cmidrule{2-8}  \cmidrule{11-17}
 			Att. Rad. & 0.0 & 0.1 & 0.2 & 0.5 & 1.0 & 2.0 & 4.0 & & &
 			            0.0 & 0.1 & 0.2 & 0.5 & 1.0 & 2.0 & 4.0 \\
 			\midrule
 			& \multicolumn{16}{c}{\textbf{MNIST -- KMNIST}} \\
            \PostNet  & 90.1 &  88.0 &  86.2 &  82.2 &  79.0 &  77.1 & \bf{66.1} & &
                      & 90.1 &  84.5 &  77.2 &  46.4 &  12.9 &   2.7 &   2.4 \\
            \PriorNet & \bf{99.8} & \bf{99.7} & \bf{99.6} & \bf{99.1} & \bf{97.2} & \bf{79.7} &   4.7 & &
                      & \bf{99.8} & \bf{99.7} & \bf{99.6} & \bf{99.2} & \bf{97.9} & \bf{93.7} & \bf{85.6} \\
            \DDNet    & 98.1 &  97.7 &  97.2 &  94.8 &  87.0 &  48.7 &   3.0 & &
                      & 98.1 &  97.8 &  97.3 &  95.8 &  92.3 &  83.3 &  63.3 \\
            \EvNet    & 36.8 &  35.0 &  33.7 &  30.9 &  28.2 &  25.3 &  22.1 & &
                      & 36.8 &  19.3 &  10.7 &   3.9 &   2.1 &   1.8 &   2.2 \\
 			\midrule
 			& \multicolumn{16}{c}{\textbf{CIFAR10 -- SVHN}} \\
\bottomrule
 		\end{tabular}
 	\end{small}
 	\label{tab:id_ood_attacks_measure_distU_auroc}
\end{table*}

 \begin{table*}[htbp!]
 	\centering
 	\caption{OOD detection (AU-PR) under FGSM uncertainty attacks against differential entropy on ID data and OOD data.}
 	\begin{small}
 		\begin{tabular}{@{}rrrrrrrrc|crrrrrrr@{}}
 			\toprule
 			& \multicolumn{7}{c}{ID-Attack (non-attacked OOD)} &  & &  \multicolumn{7}{c}{OOD-Attack (non-attacked ID)} \\
 			\cmidrule{2-8}  \cmidrule{11-17}
 			Att. Rad. & 0.0 & 0.1 & 0.2 & 0.5 & 1.0 & 2.0 & 4.0 & & &
 			            0.0 & 0.1 & 0.2 & 0.5 & 1.0 & 2.0 & 4.0 \\
 			\midrule
 			& \multicolumn{16}{c}{\textbf{MNIST -- KMNIST}} \\
            \PostNet  & 94.5 &  94.2 &  94.1 &  93.5 &  89.9 &  81.2 &  \bf{71.6} & &
                      & 94.5 &  93.3 &  92.0 &  87.6 &  81.1 &  75.7 &  75.7 \\
            \PriorNet & \bf{99.6} &  \bf{99.4} &  \bf{99.2} &  \bf{98.1} &  \bf{95.6} &  \bf{90.0} &  65.3 & &
                      & \bf{99.6} &  \bf{99.4} &  \bf{99.2} &  \bf{98.6} &  97.5 &  \bf{95.9} &  \bf{94.4} \\
            \DDNet    & 99.3 &  99.1 &  98.9 &  98.0 &  95.4 &  80.9 &  48.2 & &
                      & 99.3 &  99.2 &  99.0 &  98.5 &  \bf{97.6} &  95.5 &  92.0 \\
            \EvNet    & 69.0 &  67.4 &  66.2 &  64.0 &  61.9 &  59.8 &  56.70 & &
                      & 9.0 &  60.1 &  56.5 &  53.4 &  52.7 &  52.9 &  53.5 \\
 			\midrule
 			& \multicolumn{16}{c}{\textbf{CIFAR10 -- SVHN}} \\
\bottomrule
 		\end{tabular}
 	\end{small}
 	\label{tab:id_ood_attacks_measure_diffE_aupr_fgsm}
\end{table*}

 \begin{table*}[htbp!]
 	\centering
 	\caption{OOD detection (AU-PR) under Noise uncertainty attacks against differential entropy on ID data and OOD data.}
 	\begin{small}
 		\begin{tabular}{@{}rrrrrrrrc|crrrrrrr@{}}
 			\toprule
 			& \multicolumn{7}{c}{ID-Attack (non-attacked OOD)} &  & &  \multicolumn{7}{c}{OOD-Attack (non-attacked ID)} \\
 			\cmidrule{2-8}  \cmidrule{11-17}
 			Noise Std & 0.0 & 0.1 & 0.2 & 0.5 & 1.0 & 2.0 & 4.0 & & &
 			            0.0 & 0.1 & 0.2 & 0.5 & 1.0 & 2.0 & 4.0 \\
 			\midrule
 			& \multicolumn{16}{c}{\textbf{MNIST -- KMNIST}} \\
            \PostNet  & 93.0 &  94.2 &  82.3 &  34.4 &  31.6 &  31.0 &  30.9 & &
                      & 92.2 &  91.8 &  91.5 &  92.3 &   92.7 &  93.2 &  93.5 \\
            \PriorNet & \bf{99.7} &  \bf{99.6} &  \bf{96.7} &  \bf{40.0} &  \bf{40.6} &  \bf{45.7} &  \bf{55.6} & &
                      & \bf{99.5} &  97.3 &  96.5 &  99.4 &  \bf{100.0} &  99.5 &  72.4 \\
            \DDNet    & 99.1 &  97.5 &  81.2 &  31.3 &  31.0 &  30.9 &  31.2 & &
                      & 99.0 &  \bf{98.8} &  \bf{99.2} &  \bf{99.8} &   99.9 &  \bf{99.8} &  \bf{99.1} \\
            \EvNet    & 65.5 &  60.5 &  51.4 &  35.3 &  34.5 &  35.5 &  35.0 & &
                      & 62.5 &  47.2 &  40.9 &  35.1 &   34.6 &  33.5 &  34.9 \\
 			\midrule
 			& \multicolumn{16}{c}{\textbf{CIFAR10 -- SVHN}} \\
\bottomrule
 		\end{tabular}
 	\end{small}
 	\label{tab:id_ood_attacks_measure_diffE_aupr_noise}
\end{table*}

\clearpage
\subsection{How to make DBU models more robust}

To improve robustness of DBU models we perform median smoothing and adversarial training. Smoothing computes the smooth median, worst case and best case performance of DBU models for three tasks:  distinguishing between correct and wrong predictions, attack detection, distinguishing between ID data and OOD data under label attacks and under uncertainty attacks.

\begin{table*}[ht!]
	\centering
	\caption{Distinguishing between correctly and wrongly labeled inputs based on differential entropy under PGD label attacks. Smoothed DBU models on CIFAR10. Column format: guaranteed lowest performance $\cdot$ empirical performance $\cdot$ guaranteed highest performance (blue: normally/adversarially trained smooth classifier is more robust than the base model).}
	\label{tab:cifar10_smooth_confidence}
	\resizebox{\textwidth}{!}{
\begin{tabular}{llcccccc}
    \toprule
    & \textbf{Att. Rad.} & 0.0 &   0.1 &  0.2 &  0.5 &  1.0 &  2.0 \\
    \midrule
    
    \midrule
    
    \midrule
    
    \bottomrule
    \end{tabular}}
\end{table*}

\begin{table*}[ht!]
	\centering
	\caption{Distinguishing between correctly and wrongly labeled inputs based on differential entropy under PGD label attacks. Smoothed DBU models on MNIST. Column format: guaranteed lowest performance $\cdot$ empirical performance $\cdot$ guaranteed highest performance (blue: normally/adversarially trained smooth classifier is more robust than the base model).}
	\label{tab:mnist_smooth_confidence}
	\resizebox{\textwidth}{!}{
		\begin{tabular}{llccccccc}
			\toprule
			& \textbf{Att. Rad.} & 0.0 &   0.1 &  0.2 &  0.5 &  1.0 &  2.0 \\
			\midrule
              \multirow{4}{1.2cm}{Smoothed models} &   
  \textbf{PostNet} &  
  $97.2\cdot\bm{{\color{black}99.4}}\cdot100.0$ & 
  $95.9\cdot\bm{{\color{black}99.1}}\cdot99.9$ &  
  $94.7\cdot\bm{{\color{black}98.9}}\cdot99.9$ & 
  $89.3\cdot\bm{{\color{black}96.8}}\cdot\phantom{0}99.9$ & 
  $75.5\cdot\bm{{\color{blue}90.2}}\cdot100.0$ & 
  $35.5\cdot\bm{{\color{blue}56.7}}\cdot100.0$ \\
& \textbf{PriorNet} &  
$96.8\cdot\bm{{\color{black}99.2}}\cdot\phantom{0}99.3$ & 
$95.5\cdot\bm{{\color{black}99.1}}\cdot99.7$ & 
$94.6\cdot\bm{{\color{black}98.8}}\cdot99.7$ & 
$90.2\cdot\bm{{\color{black}97.2}}\cdot\phantom{0}99.9$ & 
$81.1\cdot\bm{{\color{blue}93.4}}\cdot\phantom{0}99.9$ &  
$53.9\cdot\bm{{\color{blue}75.2}}\cdot100.0$ \\
 &   \textbf{DDNet} &  
 $97.6\cdot\bm{{\color{black}99.4}}\cdot\phantom{0}99.5$ &  
 $96.8\cdot\bm{{\color{black}99.2}}\cdot99.4$ &
 $95.5\cdot\bm{{\color{black}98.8}}\cdot99.4$ & 
 $90.4\cdot\bm{{\color{black}97.2}}\cdot\phantom{0}99.8$ &  
 $77.0\cdot\bm{{\color{black}91.3}}\cdot100.0$ &
 $29.2\cdot\bm{{\color{black}48.6}}\cdot100.0$ \\
  &  \textbf{EvNet} &   
  $97.3\cdot\bm{{\color{black}99.4}}\cdot\phantom{0}99.4$ & 
  $95.4\cdot\bm{{\color{black}98.8}}\cdot99.6$ &            
  $93.9\cdot\bm{98.7}\cdot99.9$ & 
  $89.0\cdot\bm{{\color{blue}96.5}}\cdot100.0$ &  
  $78.9\cdot\bm{{\color{blue}92.9}}\cdot100.0$ & 
  $52.2\cdot\bm{{\color{blue}73.2}}\cdot100.0$ \\

			\midrule
              \multirow{4}{1.2cm}{Smoothed + adv. w. label attacks} &   
  \textbf{PostNet} &  - &  
  $94.4\cdot\bm{{\color{black}98.6}}\cdot99.5$ &  
  $90.6\cdot\bm{{\color{black}97.9}}\cdot99.9$ &  
  $83.4\cdot\bm{{\color{black}93.1}}\cdot\phantom{0}99.9$ &   
  $72.1\cdot\bm{{\color{blue}91.2}}\cdot100.0$ &   
  $41.8\cdot\bm{{\color{blue}65.0}}\cdot100.0$ \\
\bottomrule
		\end{tabular}}
\end{table*}

\begin{table*}[ht!]
	\centering
	\caption{Distinguishing between correctly and wrongly labeled inputs based on differential entropy under PGD label attacks. Smoothed DBU models on Sensorless. Column format: guaranteed lowest performance $\cdot$ empirical performance $\cdot$ guaranteed highest performance (blue: normally/adversarially trained smooth classifier is more robust than the base model).}
	\label{tab:sensorless_smooth_confidence}
	\resizebox{\textwidth}{!}{
		\begin{tabular}{llccccccc}
			\toprule
			& \textbf{Att. Rad.} & 0.0 &   0.1 &  0.2 &  0.5 &  1.0 &  2.0 \\
			\midrule
              \multirow{4}{1.2cm}{Smoothed models} &  
  \textbf{PostNet} &  
  $93.5\cdot\bm{{\color{black}98.4}}\cdot100.0$ & 
  $\phantom{0}6.7\cdot\bm{{\color{blue}12.4}}\cdot100.0$ &  
  $\phantom{0}2.9\cdot\phantom{0}\bm{{\color{blue}5.3}}\cdot100.0$ &  
  $\phantom{0}4.1\cdot\phantom{0}\bm{{\color{blue}4.1}}\cdot\phantom{0}49.1$ & 
  $\phantom{0}6.4\cdot\phantom{0}\bm{{\color{black}6.4}}\cdot\phantom{00}6.4$ &
  $10.6\cdot\bm{{\color{blue}10.6}}\cdot\phantom{0}10.6$ \\
& \textbf{PriorNet} & 
$97.1\cdot\bm{{\color{black}99.3}}\cdot100.0$ &   
$\phantom{0}8.6\cdot\bm{{\color{blue}17.6}}\cdot100.0$ & 
$\phantom{0}3.3\cdot\phantom{0}\bm{{\color{blue}7.7}}\cdot100.0$ &  
$\phantom{0}0.7\cdot\phantom{0}\bm{{\color{blue}1.5}}\cdot100.0$ & 
$\phantom{0}0.4\cdot\phantom{0}\bm{{\color{blue}0.7}}\cdot100.0$ &  
$\phantom{0}0.1\cdot\phantom{0}\bm{0.2}\cdot100.0$ \\
 &   \textbf{DDNet} &  
 $95.9\cdot\bm{{\color{black}98.9}}\cdot\phantom{0}99.7$ & 
 $\phantom{0}7.0\cdot\bm{{\color{blue}14.0}}\cdot100.0$ & 
 $\phantom{0}0.8\cdot\phantom{0}\bm{{\color{black}1.3}}\cdot100.0$ &   
 $\phantom{0}0.2\cdot\phantom{0}\bm{0.4}\cdot100.0$ &           
 $\phantom{0}0.2\cdot\phantom{0}\bm{0.2}\cdot100.0$ &   
 $\phantom{0}0.2\cdot\phantom{0}\bm{{\color{blue}0.4}}\cdot100.0$ \\
  &  \textbf{EvNet} &   
  $94.0\cdot\bm{{\color{black}99.0}}\cdot\phantom{0}99.9$ & 
  $18.1\cdot\bm{{\color{blue}34.2}}\cdot100.0$ & 
  $\phantom{0}9.6\cdot\bm{{\color{blue}17.1}}\cdot100.0$ &
  $\phantom{0}4.1\cdot\phantom{0}\bm{{\color{blue}6.8}}\cdot100.0$ &  
  $\phantom{0}2.7\cdot\phantom{0}\bm{{\color{blue}4.9}}\cdot100.0$ &
  $\phantom{0}2.4\cdot\phantom{0}\bm{{\color{blue}4.3}}\cdot100.0$ \\
\bottomrule
		\end{tabular}}
\end{table*}

\begin{table*}[ht!]
	\centering
	\caption{Distinguishing between correctly and wrongly labeled inputs based on differential entropy under PGD label attacks. Smoothed DBU models on Segment. Column format: guaranteed lowest performance $\cdot$ empirical performance $\cdot$ guaranteed highest performance (blue: normally/adversarially trained smooth classifier is more robust than the base model)..}
	\label{tab:segment_smooth_confidence}
	\resizebox{\textwidth}{!}{
		\begin{tabular}{llccccccc}
			\toprule
			& \textbf{Att. Rad.} & 0.0 &   0.1 &  0.2 &  0.5 &  1.0 &  2.0 \\
			\midrule
               \multirow{4}{1.2cm}{Smoothed models} & 
   \textbf{PostNet} &  
   $94.0\cdot\bm{{\color{black}99.1}}\cdot99.8$ & 
   $63.5\cdot\bm{{\color{blue}84.7}}\cdot100.0$ & 
   $33.2\cdot\bm{{\color{blue}56.1}}\cdot100.0$ & 
   $10.2\cdot\bm{{\color{blue}16.9}}\cdot100.0$ &  
   $\phantom{0}5.2\cdot\bm{{\color{blue}10.3}}\cdot100.0$ &  
   $\phantom{0}0.3\cdot\phantom{0}\bm{{\color{black}0.3}}\cdot\phantom{00}0.3$ \\
& \textbf{PriorNet} &  
$97.0\cdot\bm{{\color{black}99.8}}\cdot99.9$ & 
$75.6\cdot\bm{{\color{black}90.8}}\cdot100.0$ &
$31.1\cdot\bm{{\color{blue}50.8}}\cdot100.0$ & 
$\phantom{0}2.6\cdot\phantom{0}\bm{{\color{blue}4.7}}\cdot100.0$ &  
$\phantom{0}0.0\cdot\phantom{0}\bm{{\color{black}0.0}}\cdot100.0$ &      
$\phantom{0}0.0\cdot\phantom{0}\bm{0.0}\cdot100.0$ \\
 &   \textbf{DDNet} & 
 $96.2\cdot\bm{{\color{black}99.5}}\cdot99.7$ & 
 $75.7\cdot\bm{{\color{black}89.8}}\cdot\phantom{0}99.9$ &
 $28.5\cdot\bm{{\color{black}51.6}}\cdot100.0$ & 
 $3.7\cdot\phantom{0}\bm{{\color{blue}8.2}}\cdot100.0$ &        
 $\phantom{0}0.0\cdot\phantom{0}\bm{0.0}\cdot100.0$ &              
 $\phantom{0}0.0\cdot\phantom{0}\bm{0.0}\cdot100.0$ \\
  &  \textbf{EvNet} &  
  $95.8\cdot\bm{{\color{black}99.6}}\cdot99.9$ & 
  $80.2\cdot\bm{{\color{black}93.7}}\cdot100.0$ &
  $35.2\cdot\bm{{\color{blue}57.2}}\cdot100.0$ &
  $\phantom{0}6.8\cdot\bm{{\color{blue}12.0}}\cdot100.0$ &
  $\phantom{0}1.2\cdot\phantom{0}\bm{{\color{blue}2.1}}\cdot100.0$ & 
  $\phantom{0}1.1\cdot\phantom{0}\bm{{\color{blue}2.0}}\cdot100.0$ \\
\bottomrule
		\end{tabular}}
\end{table*}

\begin{table*}[ht!]
	\centering
	\caption{Distinguishing between correctly and wrongly labeled inputs based on differential entropy under FGSM label attacks. Smoothed DBU models on CIFAR10. Column format: guaranteed lowest performance $\cdot$ empirical performance $\cdot$ guaranteed highest performance (blue: normally/adversarially trained smooth classifier is more robust than the base model).}
	\label{tab:cifar10_smooth_confidence_fgsm}
	\resizebox{\textwidth}{!}{
		\begin{tabular}{llccccccc}
			\toprule
			& \textbf{Att. Rad.} & 0.0 &   0.1 &  0.2 &  0.5 &  1.0 &  2.0 \\
			\midrule
              \multirow{4}{1.2cm}{Smoothed models} &  
  \textbf{PostNet} & 
  $80.5\cdot\bm{{\color{black}91.4}}\cdot94.4$ &  
  $52.3\cdot\bm{{\color{black}73.2}}\cdot95.4$ &
  $35.8\cdot\bm{{\color{black}57.2}}\cdot97.5$ & 
  $17.0\cdot\bm{{\color{black}29.0}}\cdot100.0$ &  
  $10.2\cdot\bm{{\color{blue}18.7}}\cdot100.0$ &  
  $\phantom{0}8.1\cdot\bm{{\color{black}14.7}}\cdot100.0$ \\
& \textbf{PriorNet} & 
$81.9\cdot\bm{{\color{black}87.7}}\cdot88.8$ &  
$69.6\cdot\bm{{\color{blue}78.4}}\cdot90.3$ &   
$53.3\cdot\bm{{\color{blue}70.5}}\cdot91.7$ &  
$42.1\cdot\bm{{\color{blue}62.6}}\cdot\phantom{0}97.2$ &  
$37.5\cdot\bm{{\color{blue}55.7}}\cdot100.0$ &  
$36.0\cdot\bm{{\color{blue}59.5}}\cdot100.0$ \\
 &   \textbf{DDNet} & 
 $65.9\cdot\bm{{\color{black}84.1}}\cdot85.6$ &  
 $55.3\cdot\bm{{\color{black}69.6}}\cdot87.0$ &  
 $38.6\cdot\bm{{\color{black}55.8}}\cdot87.2$ &  
 $16.3\cdot\bm{{\color{black}28.5}}\cdot\phantom{0}94.6$ &  
 $\phantom{0}6.4\cdot\bm{{\color{black}12.0}}\cdot\phantom{0}99.9$ &     
 $\phantom{0}3.6\cdot\phantom{0}\bm{{\color{blue}7.2}}\cdot100.0$ \\
  &  \textbf{EvNet} & 
  $76.3\cdot\bm{{\color{black}90.4}}\cdot91.7$ & 
  $54.1\cdot\bm{{\color{black}74.5}}\cdot95.5$ & 
  $35.5\cdot\bm{{\color{black}54.7}}\cdot95.1$ &
  $14.6\cdot\bm{{\color{black}29.3}}\cdot\phantom{0}95.6$ & 
  $\phantom{0}8.6\cdot\bm{{\color{black}16.1}}\cdot100.0$ & 
  $\phantom{0}7.2\cdot\bm{{\color{black}13.0}}\cdot100.0$ \\

			\midrule
              \multirow{4}{1.2cm}{Smoothed + adv. w. label attacks} &  
  \textbf{PostNet} &  - &  
  $52.3\cdot\bm{{\color{black}71.6}}\cdot95.1$ &  
  $34.7\cdot\bm{{\color{black}54.8}}\cdot96.6$ &      
  $18.9\cdot\bm{{\color{blue}32.1}}\cdot\phantom{0}99.4$ &   
  $10.9\cdot\bm{{\color{blue}19.2}}\cdot100.0$ &   
  $\phantom{0}8.5\cdot\bm{{\color{blue}16.2}}\cdot100.0$ \\
\bottomrule
		\end{tabular}}
\end{table*}

\begin{table*}[ht!]
	\centering
	\caption{Distinguishing between correctly and wrongly labeled inputs based on differential entropy under FGSM label attacks. Smoothed DBU models on MNIST. Column format: guaranteed lowest performance $\cdot$ empirical performance $\cdot$ guaranteed highest performance (blue: normally/adversarially trained smooth classifier is more robust than the base model).}
	\label{tab:mnist_smooth_confidence_fgsm}
	\resizebox{\textwidth}{!}{
		\begin{tabular}{llccccccc}
			\toprule
			& \textbf{Att. Rad.} & 0.0 &   0.1 &  0.2 &  0.5 &  1.0 &  2.0 \\
			\midrule
              \multirow{4}{1.2cm}{Smoothed models} & 
  \textbf{PostNet} & 
  $97.2\cdot\bm{{\color{black}99.3}}\cdot99.9$ &
  $96.1\cdot\bm{{\color{black}99.2}}\cdot99.9$ & 
  $95.2\cdot\bm{{\color{black}98.9}}\cdot99.9$ & 
  $91.7\cdot\bm{{\color{black}98.0}}\cdot\phantom{0}99.9$ &  
  $86.1\cdot\bm{{\color{black}95.9}}\cdot100.0$ & 
  $75.7\cdot\bm{{\color{black}91.1}}\cdot100.0$ \\
&  \textbf{PriorNet} & 
$96.8\cdot\bm{{\color{black}99.2}}\cdot99.3$ & 
$95.5\cdot\bm{{\color{black}99.0}}\cdot99.6$ & 
$94.7\cdot\bm{{\color{black}98.7}}\cdot99.6$ & 
$91.3\cdot\bm{{\color{black}97.6}}\cdot\phantom{0}99.9$ &  
$85.5\cdot\bm{{\color{blue}95.6}}\cdot100.0$ & 
$78.7\cdot\bm{{\color{blue}92.4}}\cdot100.0$ \\
  &  \textbf{DDNet} & 
  $97.6\cdot\bm{{\color{black}99.3}}\cdot99.4$ & 
  $96.8\cdot\bm{{\color{black}99.2}}\cdot99.5$ & 
  $95.6\cdot\bm{{\color{black}98.7}}\cdot99.4$ & 
  $91.7\cdot\bm{{\color{black}97.7}}\cdot\phantom{0}99.9$ &  
  $83.4\cdot\bm{{\color{black}95.2}}\cdot100.0$ &
  $58.3\cdot\bm{{\color{black}79.6}}\cdot100.0$ \\
   & \textbf{EvNet} & 
   $97.3\cdot\bm{{\color{black}99.3}}\cdot99.4$ &  
   $95.5\cdot\bm{{\color{black}99.0}}\cdot99.6$ & 
   $94.3\cdot\bm{{\color{black}98.9}}\cdot99.9$ & 
   $92.0\cdot\bm{{\color{black}97.7}}\cdot100.0$ &
   $87.4\cdot\bm{{\color{blue}96.3}}\cdot100.0$ &   
   $78.8\cdot\bm{{\color{blue}92.4}}\cdot100.0$ \\

			\midrule
               \multirow{4}{1.2cm}{Smoothed + adv. w. label attacks} & 
   \textbf{PostNet} &  - &  
   $95.1\cdot\bm{{\color{black}98.9}}\cdot99.8$ & 
   $91.2\cdot\bm{{\color{black}97.2}}\cdot99.6$ &  
   $87.6\cdot\bm{{\color{black}96.3}}\cdot\phantom{0}99.9$ &  
   $81.0\cdot\bm{{\color{black}93.3}}\cdot100.0$ &
   $69.9\cdot\bm{{\color{black}87.2}}\cdot100.0$ \\
\bottomrule
		\end{tabular}}
\end{table*}

\begin{table*}[ht!]
	\centering
	\caption{Distinguishing between correctly and wrongly labeled inputs based on differential entropy under FGSM label attacks. Smoothed DBU models on Sensorless. Column format: guaranteed lowest performance $\cdot$ empirical performance $\cdot$ guaranteed highest performance (blue: normally/adversarially trained smooth classifier is more robust than the base model).}
	\label{tab:sensorless_smooth_confidence_fgsm}
	\resizebox{\textwidth}{!}{
		\begin{tabular}{llccccccc}
			\toprule
			& \textbf{Att. Rad.} & 0.0 &   0.1 &  0.2 &  0.5 &  1.0 &  2.0 \\
			\midrule
               \multirow{4}{1.2cm}{Smoothed models} &
   \textbf{PostNet} & 
   $94.5\cdot\bm{{\color{black}98.1}}\cdot100.0$ &  
   $10.3\cdot\bm{{\color{blue}19.6}}\cdot100.0$ &   
   $\phantom{0}5.1\cdot\bm{{\color{blue}11.0}}\cdot100.0$ &   
   $\phantom{0}6.4\cdot\phantom{0}\bm{{\color{black}6.4}}\cdot\phantom{00}6.4$ &  
   $10.4\cdot\bm{{\color{black}10.4}}\cdot\phantom{0}10.4$ & 
   $11.4\cdot\bm{{\color{black}11.4}}\cdot\phantom{0}11.4$ \\
& \textbf{PriorNet} & 
$97.1\cdot\bm{{\color{black}99.5}}\cdot100.0$ &  
$13.6\cdot\bm{{\color{blue}27.3}}\cdot100.0$ & 
$\phantom{0}6.3\cdot\bm{{\color{blue}12.4}}\cdot100.0$ &  
$\phantom{0}2.6\cdot\phantom{0}\bm{{\color{black}6.8}}\cdot100.0$ &  
$\phantom{0}3.1\cdot\phantom{0}\bm{{\color{black}7.3}}\cdot100.0$ & 
$\phantom{0}3.2\cdot\phantom{0}\bm{{\color{black}6.7}}\cdot100.0$ \\
 &   \textbf{DDNet} &   
 $95.9\cdot\bm{{\color{black}99.4}}\cdot\phantom{0}99.8$ & 
 $\phantom{0}8.6\cdot\bm{{\color{black}14.9}}\cdot100.0$ & 
 $\phantom{0}1.6\cdot\phantom{0}\bm{{\color{black}3.8}}\cdot100.0$ &  
 $\phantom{0}2.6\cdot\phantom{0}\bm{{\color{blue}4.5}}\cdot100.0$ &   
 $\phantom{0}3.4\cdot\phantom{0}\bm{{\color{blue}6.9}}\cdot100.0$ &  
 $\phantom{0}3.3\cdot\phantom{0}\bm{{\color{blue}6.4}}\cdot100.0$ \\
  &  \textbf{EvNet} & 
  $94.0\cdot\bm{{\color{black}98.5}}\cdot\phantom{0}99.7$ & 
  $26.0\cdot\bm{{\color{black}43.2}}\cdot100.0$ &
  $15.8\cdot\bm{{\color{blue}30.8}}\cdot100.0$ & 
  $11.7\cdot\bm{{\color{blue}20.2}}\cdot100.0$ & 
  $\phantom{0}8.1\cdot\bm{{\color{black}15.0}}\cdot100.0$ &
  $\phantom{0}7.6\cdot\bm{{\color{black}12.7}}\cdot100.0$ \\
\bottomrule
		\end{tabular}}
\end{table*}

\begin{table*}[ht!]
	\centering
	\caption{Distinguishing between correctly and wrongly labeled inputs based on differential entropy under FGSM label attacks. Smoothed DBU models on Segment. Column format: guaranteed lowest performance $\cdot$ empirical performance $\cdot$ guaranteed highest performance (blue: normally/adversarially trained smooth classifier is more robust than the base model).}
	\label{tab:segments_smooth_confidence_fgsm}
	\resizebox{\textwidth}{!}{
		\begin{tabular}{llccccccc}
			\toprule
			& \textbf{Att. Rad.} & 0.0 &   0.1 &  0.2 &  0.5 &  1.0 &  2.0 \\
			\midrule
                \multirow{4}{1.2cm}{Smoothed models} &
    \textbf{PostNet} & 
    $94.0\cdot\bm{{\color{black}99.2}}\cdot99.8$ &    
    $55.2\cdot\bm{{\color{blue}78.3}}\cdot100.0$ &   
    $40.1\cdot\bm{{\color{blue}61.4}}\cdot100.0$ &  
    $17.9\cdot\bm{{\color{blue}31.7}}\cdot100.0$ & 
    $\phantom{0}6.8\cdot\bm{{\color{black}12.7}}\cdot100.0$ &
    $17.6\cdot\bm{{\color{black}17.9}}\cdot\phantom{0}18.0$ \\
 & \textbf{PriorNet} & 
 $97.0\cdot\bm{{\color{black}99.8}}\cdot99.9$ &
 $69.2\cdot\bm{{\color{black}89.7}}\cdot100.0$ &  
 $29.7\cdot\bm{{\color{blue}45.5}}\cdot100.0$ & 
 $\phantom{0}1.7\cdot\bm{{\color{black}4.1}}\cdot100.0$ &  
 $\phantom{0}0.0\cdot\phantom{0}\bm{{\color{black}0.0}}\cdot100.0$ & 
 $\phantom{0}0.0\cdot\phantom{0}\bm{{\color{black}0.0}}\cdot100.0$ \\
   & \textbf{DDNet} & 
   $96.2\cdot\bm{{\color{black}99.5}}\cdot99.6$ &
   $70.6\cdot\bm{{\color{black}86.3}}\cdot\phantom{0}99.8$ & 
   $22.3\cdot\bm{{\color{black}38.8}}\cdot100.0$ &
   $\phantom{0}6.3\cdot\bm{{\color{blue}13.3}}\cdot100.0$ &    
   $\phantom{0}1.1\cdot\phantom{0}\bm{{\color{blue}3.0}}\cdot100.0$ &      
   $\phantom{0}0.0\cdot\phantom{0}\bm{0.0}\cdot100.0$ \\
&    \textbf{EvNet} & 
$95.8\cdot\bm{{\color{black}99.1}}\cdot99.8$ &
$78.4\cdot\bm{{\color{black}92.5}}\cdot100.0$ &  
$40.7\cdot\bm{{\color{blue}62.1}}\cdot100.0$ &  
$\phantom{0}9.8\cdot\bm{{\color{blue}17.6}}\cdot100.0$ &   
$\phantom{0}0.0\cdot\phantom{0}\bm{{\color{black}0.0}}\cdot100.0$ & 
$\phantom{0}0.0\cdot\phantom{0}\bm{{\color{black}0.0}}\cdot100.0$ \\
\bottomrule
		\end{tabular}}
\end{table*}

\begin{table*}[ht!]
	\centering
	\caption{Attack detection (PGD label attacks) based on differential entropy. Smoothed DBU models on CIFAR10. Column format: guaranteed lowest performance $\cdot$ empirical performance $\cdot$ guaranteed highest performance (blue: normally/adversarially trained smooth classifier is more robust than the base model).}
	\label{tab:cifar10_smooth_attackdetection}
	\resizebox{\textwidth}{!}{
		\begin{tabular}{llccccc}
			\toprule
			& \textbf{Att. Rad.} &   0.1 &  0.2 &  0.5 &  1.0 &  2.0 \\
			\midrule
            
			\midrule
            
			\midrule
            
			\bottomrule
		\end{tabular}}
\end{table*}

\begin{table*}[ht!]
	\centering
	\caption{Attack detection (PGD label attacks) based on differential entropy. Smoothed DBU models on MNIST. Column format: guaranteed lowest performance $\cdot$ empirical performance $\cdot$ guaranteed highest performance (blue: normally/adversarially trained smooth classifier is more robust than the base model).}
	\label{tab:mnist_smooth_attackdetection}
	\resizebox{\textwidth}{!}{
		\begin{tabular}{llcccccc}
			\toprule
			& \textbf{Att. Rad.} &   0.1 &  0.2 &  0.5 &  1.0 &  2.0 \\
			\midrule
              \multirow{4}{1.2cm}{Smoothed models} &  
  \textbf{PostNet} &  
  $30.9\cdot\bm{{\color{black}52.5}}\cdot95.6$ & 
  $31.5\cdot\bm{{\color{black}51.5}}\cdot90.9$ & 
  $31.1\cdot\bm{{\color{black}49.9}}\cdot97.1$ & 
  $30.7\cdot\bm{{\color{black}47.6}}\cdot100.0$ & 
  $30.7\cdot\bm{{\color{black}45.0}}\cdot100.0$ \\
 & \textbf{PriorNet} &  
 $38.2\cdot\bm{{\color{black}57.8}}\cdot80.9$ &  
 $36.0\cdot\bm{{\color{black}57.2}}\cdot84.3$ &  
 $31.6\cdot\bm{{\color{black}63.4}}\cdot98.4$ &
 $30.8\cdot\bm{{\color{black}61.0}}\cdot\phantom{0}99.3$ & 
 $30.7\cdot\bm{{\color{black}66.8}}\cdot100.0$ \\
   & \textbf{DDNet} &  
   $44.6\cdot\bm{{\color{black}51.9}}\cdot60.7$ &
   $39.3\cdot\bm{{\color{black}52.7}}\cdot72.2$ &
   $31.6\cdot\bm{{\color{black}50.9}}\cdot95.2$ & 
   $30.7\cdot\bm{{\color{black}47.3}}\cdot100.0$ &
   $30.7\cdot\bm{{\color{black}45.9}}\cdot100.0$ \\
&    \textbf{EvNet} &  
$36.5\cdot\bm{{\color{black}51.8}}\cdot76.1$ & 
$31.5\cdot\bm{{\color{black}51.1}}\cdot93.2$ & 
$30.7\cdot\bm{{\color{black}51.1}}\cdot99.9$ &  
$30.7\cdot\bm{{\color{black}48.7}}\cdot100.0$ &
$30.7\cdot\bm{{\color{black}43.8}}\cdot100.0$ \\

			\midrule
              \multirow{4}{1.2cm}{Smoothed + adv. w. label attacks} &  
  \textbf{PostNet} &  %
  $33.6\cdot\bm{{\color{black}52.8}}\cdot82.3$ &  
  $31.4\cdot\bm{{\color{black}51.2}}\cdot91.6$ &  
  $30.9\cdot\bm{{\color{black}49.4}}\cdot\phantom{0}99.1$ &  
  $30.7\cdot\bm{{\color{black}49.3}}\cdot100.0$ & 
  $30.7\cdot\bm{{\color{black}56.0}}\cdot100.0$ \\
& \textbf{PriorNet} &  %
$37.3\cdot\bm{{\color{black}60.5}}\cdot84.3$ & 
$34.3\cdot\bm{{\color{black}59.9}}\cdot87.9$ &  
$32.1\cdot\bm{{\color{black}61.0}}\cdot\phantom{0}97.0$ & 
$30.7\cdot\bm{{\color{black}69.3}}\cdot100.0$ & 
$30.7\cdot\bm{{\color{black}68.0}}\cdot100.0$ \\
\bottomrule
		\end{tabular}}
\end{table*}

\begin{table*}[ht!]
	\centering
	\caption{Attack detection (PGD label attacks) based on differential entropy. Smoothed DBU models on Sensorless. Column format: guaranteed lowest performance $\cdot$ empirical performance $\cdot$ guaranteed highest performance (blue: normally/adversarially trained smooth classifier is more robust than the base model).}
	\label{tab:sensorless_smooth_attackdetection}
	\resizebox{\textwidth}{!}{
		\begin{tabular}{llcccccc}
			\toprule
			& \textbf{Att. Rad.} &   0.1 &  0.2 &  0.5 &  1.0 &  2.0 \\
			\midrule
               \multirow{4}{1.2cm}{Smoothed models} & 
   \textbf{PostNet} &  %
   $30.7\cdot\bm{{\color{blue}61.9}}\cdot100.0$ & 
   $30.7\cdot\bm{{\color{blue}60.1}}\cdot100.0$ & 
   $46.5\cdot\bm{{\color{blue}50.0}}\cdot\phantom{0}75.5$ &  
   $50.0\cdot\bm{{\color{blue}50.0}}\cdot\phantom{0}50.0$ & 
   $50.0\cdot\bm{{\color{black}50.0}}\cdot\phantom{0}50.0$ \\
 & \textbf{PriorNet} &  %
 $30.7\cdot\bm{{\color{blue}50.1}}\cdot100.0$ &
 $30.7\cdot\bm{{\color{blue}46.5}}\cdot100.0$ &  
 $30.7\cdot\bm{{\color{blue}42.3}}\cdot100.0$ &  
 $30.7\cdot\bm{{\color{blue}66.7}}\cdot100.0$ &  
 $30.9\cdot\bm{{\color{blue}79.2}}\cdot100.0$ \\
   & \textbf{DDNet} &  %
   $30.7\cdot\bm{{\color{blue}57.5}}\cdot100.0$ &
   $30.7\cdot\bm{{\color{blue}49.9}}\cdot100.0$ & 
   $30.7\cdot\bm{{\color{blue}45.5}}\cdot100.0$ & 
   $30.7\cdot\bm{{\color{blue}50.0}}\cdot100.0$ &
   $30.7\cdot\bm{{\color{blue}59.3}}\cdot100.0$ \\
&    \textbf{EvNet} &  %
$30.7\cdot\bm{{\color{blue}62.0}}\cdot100.0$ & 
$30.7\cdot\bm{{\color{blue}59.6}}\cdot100.0$ &  
$30.7\cdot\bm{{\color{blue}55.8}}\cdot100.0$ &
$30.7\cdot\bm{{\color{blue}48.3}}\cdot100.0$ & 
$31.8\cdot\bm{{\color{blue}50.0}}\cdot100.0$ \\

			\midrule
               \multirow{4}{1.2cm}{Smoothed + adv. w. label attacks} & 
   \textbf{PostNet} &  %
   $30.7\cdot\bm{{\color{blue}58.8}}\cdot100.0$ &  
   $30.7\cdot\bm{{\color{blue}58.2}}\cdot100.0$ &   
   $50.0\cdot\bm{{\color{blue}50.0}}\cdot\phantom{0}50.0$ &   
   $50.0\cdot\bm{{\color{blue}50.0}}\cdot\phantom{0}50.0$ &  
   $50.0\cdot\bm{{\color{black}50.0}}\cdot\phantom{0}50.0$ \\
& \textbf{PriorNet} &  %
$30.7\cdot\bm{{\color{blue}60.2}}\cdot100.0$ &  
$30.7\cdot\bm{{\color{blue}54.6}}\cdot100.0$ &  
$30.7\cdot\bm{{\color{blue}45.0}}\cdot100.0$ & 
$30.7\cdot\bm{{\color{blue}38.0}}\cdot100.0$ &  
$33.9\cdot\bm{{\color{blue}49.9}}\cdot100.0$ \\
\bottomrule
		\end{tabular}}
\end{table*}

\begin{table*}[ht!]
	\centering
	\caption{Attack detection (PGD label attacks) based on differential entropy. Smoothed DBU models on Segment. Column format: guaranteed lowest performance $\cdot$ empirical performance $\cdot$ guaranteed highest performance (blue: normally/adversarially trained smooth classifier is more robust than the base model).}
	\label{tab:segment_smooth_attackdetection}
	\resizebox{\textwidth}{!}{
		\begin{tabular}{llcccccc}
			\toprule
			& \textbf{Att. Rad.} &   0.1 &  0.2 &  0.5 &  1.0 &  2.0 \\
			\midrule
                \multirow{4}{1.2cm}{Smoothed models} & 
    \textbf{PostNet} &  %
    $30.8\cdot\bm{{\color{black}73.5}}\cdot100.0$ &  
    $30.8\cdot\bm{{\color{black}59.9}}\cdot100.0$ & 
    $30.8\cdot\bm{{\color{blue}60.3}}\cdot100.0$ &  
    $30.8\cdot\bm{{\color{blue}50.2}}\cdot100.0$ &  
    $49.5\cdot\bm{{\color{black}50.0}}\cdot\phantom{0}50.0$ \\
&  \textbf{PriorNet} &  %
$30.9\cdot\bm{{\color{black}77.1}}\cdot\phantom{0}99.9$ &  
$30.8\cdot\bm{{\color{black}78.1}}\cdot100.0$ &
$30.8\cdot\bm{{\color{blue}39.5}}\cdot100.0$ & 
$30.8\cdot\bm{{\color{blue}35.2}}\cdot100.0$ &  
$30.8\cdot\bm{{\color{blue}41.4}}\cdot100.0$ \\
  &  \textbf{DDNet} &  %
  $31.4\cdot\bm{{\color{black}69.6}}\cdot\phantom{0}99.5$ &  
  $30.8\cdot\bm{{\color{black}71.2}}\cdot100.0$ & 
  $30.8\cdot\bm{{\color{blue}54.3}}\cdot100.0$ &  
  $30.8\cdot\bm{{\color{blue}35.5}}\cdot100.0$ & 
  $30.8\cdot\bm{{\color{blue}35.7}}\cdot100.0$ \\
   & \textbf{EvNet} &  %
   $30.8\cdot\bm{{\color{black}86.2}}\cdot100.0$ &  
   $30.8\cdot\bm{{\color{black}80.3}}\cdot100.0$ & 
   $30.8\cdot\bm{{\color{blue}54.0}}\cdot100.0$ &  
   $30.8\cdot\bm{{\color{blue}43.3}}\cdot100.0$ &
   $30.8\cdot\bm{{\color{blue}40.5}}\cdot100.0$ \\

			\midrule
               \multirow{4}{1.2cm}{Smoothed + adv. w. label attacks} & 
   \textbf{PostNet} &  %
   $30.8\cdot\bm{{\color{black}75.6}}\cdot100.0$ &  
   $30.8\cdot\bm{{\color{black}69.7}}\cdot100.0$ & 
   $30.8\cdot\bm{{\color{blue}66.5}}\cdot100.0$ &   
   $30.8\cdot\bm{{\color{blue}50.0}}\cdot100.0$ & 
   $50.0\cdot\bm{{\color{black}50.0}}\cdot\phantom{0}50.0$ \\
 & \textbf{PriorNet} &  %
 $31.0\cdot\bm{{\color{black}74.4}}\cdot\phantom{0}99.2$ &  
 $30.8\cdot\bm{{\color{black}74.0}}\cdot100.0$ & 
 $30.8\cdot\bm{{\color{blue}59.8}}\cdot100.0$ &  
 $30.8\cdot\bm{{\color{blue}56.0}}\cdot100.0$ &  
 $30.8\cdot\bm{{\color{blue}38.8}}\cdot100.0$ \\
\bottomrule
		\end{tabular}}
\end{table*}

\begin{table*}[ht!]
	\centering
	\caption{Attack detection (FGSM label attacks) based on differential entropy. Smoothed DBU models on CIFAR10. Column format: guaranteed lowest performance $\cdot$ empirical performance $\cdot$ guaranteed highest performance (blue: normally/adversarially trained smooth classifier is more robust than the base model).}
	\label{tab:cifar10_smooth_attackdetection_fgsm}
	\resizebox{\textwidth}{!}{
		\begin{tabular}{llcccccc}
			\toprule
			& \textbf{Att. Rad.} &   0.1 &  0.2 &  0.5 &  1.0 &  2.0 \\
			\midrule
               \multirow{4}{1.2cm}{Smoothed models} & \textbf{PostNet} &  %
   $33.1\cdot\bm{{\color{black}50.3}}\cdot89.9$ & 
   $31.0\cdot\bm{{\color{black}50.2}}\cdot96.9$ & 
   $30.7\cdot\bm{{\color{black}50.1}}\cdot100.0$ &  
   $30.7\cdot\bm{{\color{black}49.5}}\cdot100.0$ &  
   $30.7\cdot\bm{{\color{black}50.2}}\cdot100.0$ \\
 & \textbf{PriorNet} &  
 $36.0\cdot\bm{{\color{black}50.8}}\cdot74.6$ &  
 $33.0\cdot\bm{{\color{black}50.4}}\cdot82.8$ &  
 $31.2\cdot\bm{{\color{black}50.2}}\cdot\phantom{0}\phantom{0}95.6$ &  
 $30.7\cdot\bm{{\color{black}50.7}}\cdot\phantom{0}99.9$ &  
 $30.7\cdot\bm{{\color{black}51.4}}\cdot100.0$ \\
   & \textbf{DDNet} &  
   $36.4\cdot\bm{{\color{black}50.4}}\cdot76.4$ &  
   $32.8\cdot\bm{{\color{black}49.9}}\cdot84.6$ &  
   $30.8\cdot\bm{{\color{black}50.1}}\cdot\phantom{0}97.9$ &  
   $30.7\cdot\bm{{\color{black}50.2}}\cdot100.0$ &
   $30.7\cdot\bm{{\color{black}49.9}}\cdot100.0$ \\
&    \textbf{EvNet} &  
$32.9\cdot\bm{{\color{black}50.3}}\cdot89.7$ & 
$31.4\cdot\bm{{\color{black}50.2}}\cdot94.0$ & 
$30.8\cdot\bm{{\color{black}50.1}}\cdot\phantom{0}98.0$ & 
$30.7\cdot\bm{{\color{black}49.7}}\cdot100.0$ & 
$30.7\cdot\bm{{\color{black}49.7}}\cdot100.0$ \\

			\midrule
              \multirow{4}{1.2cm}{Smoothed + adv. w. label attacks} &  
  \textbf{PostNet} &  %
  $32.7\cdot\bm{{\color{black}50.1}}\cdot90.3$ &  
  $31.1\cdot\bm{{\color{black}50.3}}\cdot96.4$ &   
  $30.7\cdot\bm{{\color{black}50.1}}\cdot\phantom{0}99.7$ &  
  $30.7\cdot\bm{{\color{black}49.8}}\cdot100.0$ &  
  $30.7\cdot\bm{{\color{black}50.5}}\cdot100.0$ \\
 & \textbf{PriorNet} &  %
 $35.4\cdot\bm{{\color{black}52.3}}\cdot78.9$ &  
 $32.9\cdot\bm{{\color{black}51.3}}\cdot84.5$ &   
 $30.7\cdot\bm{{\color{black}50.3}}\cdot\phantom{0}98.7$ &  
 $30.7\cdot\bm{{\color{black}50.7}}\cdot100.0$ &   
 $30.8\cdot\bm{{\color{black}50.2}}\cdot\phantom{0}98.2$ \\
\bottomrule
		\end{tabular}}
\end{table*}

\begin{table*}[ht!]
	\centering
	\caption{Attack detection (FGSM label attacks) based on differential entropy. Smoothed DBU models on MNIST. Column format: guaranteed lowest performance $\cdot$ empirical performance $\cdot$ guaranteed highest performance (blue: normally/adversarially trained smooth classifier is more robust than the base model).}
	\label{tab:mnist_smooth_attackdetection_fgsm}
	\resizebox{\textwidth}{!}{
		\begin{tabular}{llcccccc}
			\toprule
			& \textbf{Att. Rad.} &   0.1 &  0.2 &  0.5 &  1.0 &  2.0 \\
			\midrule
               \multirow{4}{1.2cm}{Smoothed models} & 
   \textbf{PostNet} & 
   $30.9\cdot\bm{{\color{black}52.3}}\cdot95.6$ &
   $31.5\cdot\bm{{\color{black}51.2}}\cdot90.8$ &  
   $31.1\cdot\bm{{\color{black}49.8}}\cdot97.0$ &  
   $30.7\cdot\bm{{\color{black}48.3}}\cdot100.0$ & 
   $30.7\cdot\bm{{\color{black}46.5}}\cdot100.0$ \\
& \textbf{PriorNet} & 
$38.1\cdot\bm{{\color{black}57.7}}\cdot80.8$ & 
$35.8\cdot\bm{{\color{black}56.6}}\cdot84.0$ &  
$31.5\cdot\bm{{\color{black}61.7}}\cdot98.3$ & 
$30.8\cdot\bm{{\color{black}58.9}}\cdot\phantom{0}99.2$ &  
$30.7\cdot\bm{{\color{black}62.3}}\cdot100.0$ \\
 &   \textbf{DDNet} &  
 $44.7\cdot\bm{{\color{black}52.0}}\cdot60.9$ &
 $39.4\cdot\bm{{\color{black}52.9}}\cdot72.5$ &
 $31.6\cdot\bm{{\color{black}50.8}}\cdot95.2$ & 
 $30.7\cdot\bm{{\color{black}47.5}}\cdot100.0$ &
 $30.7\cdot\bm{{\color{black}46.8}}\cdot100.0$ \\
  &  \textbf{EvNet} &  
  $36.5\cdot\bm{{\color{black}51.7}}\cdot76.0$ & 
  $31.5\cdot\bm{{\color{black}51.1}}\cdot93.2$ & 
  $30.7\cdot\bm{{\color{black}50.9}}\cdot99.9$ & 
  $30.7\cdot\bm{{\color{black}48.9}}\cdot100.0$ & 
  $30.7\cdot\bm{{\color{black}46.2}}\cdot100.0$ \\

			\midrule
              \multirow{4}{1.2cm}{Smoothed + adv. w. label attacks} &  
  \textbf{PostNet} &  %
  $33.5\cdot\bm{{\color{black}52.6}}\cdot82.2$ &  
  $31.4\cdot\bm{{\color{black}51.0}}\cdot91.5$ &  
  $30.9\cdot\bm{{\color{black}49.8}}\cdot\phantom{0}99.0$ &  
  $30.7\cdot\bm{{\color{black}50.1}}\cdot100.0$ & 
  $30.7\cdot\bm{{\color{black}54.4}}\cdot100.0$ \\
 & \textbf{PriorNet} &  %
 $37.3\cdot\bm{{\color{black}60.6}}\cdot84.3$ &  
 $34.2\cdot\bm{{\color{black}59.5}}\cdot87.8$ &  
 $32.1\cdot\bm{{\color{black}60.0}}\cdot\phantom{0}96.9$ & 
 $30.7\cdot\bm{{\color{black}66.3}}\cdot100.0$ & 
 $30.7\cdot\bm{{\color{black}63.3}}\cdot100.0$ \\
\bottomrule
		\end{tabular}}
\end{table*}

\begin{table*}[ht!]
	\centering
	\caption{Attack detection (FGSM label attacks) based on differential entropy. Smoothed DBU models on Sensorless. Column format: guaranteed lowest performance $\cdot$ empirical performance $\cdot$ guaranteed highest performance (blue: normally/adversarially trained smooth classifier is more robust than the base model).}
	\label{tab:sensorless_smooth_attackdetection_fgsm}
	\resizebox{\textwidth}{!}{
		\begin{tabular}{llcccccc}
			\toprule
			& \textbf{Att. Rad.} &    0.1 &  0.2 &  0.5 &  1.0 &  2.0 \\
			\midrule
               \multirow{4}{1.2cm}{Smoothed models} & 
   \textbf{PostNet} &  %
   $30.7\cdot\bm{{\color{black}82.0}}\cdot100.0$ & 
   $30.7\cdot\bm{{\color{black}88.6}}\cdot100.0$ & 
   $50.0\cdot\bm{{\color{black}50.0}}\cdot\phantom{0}50.1$ & 
   $50.0\cdot\bm{{\color{black}50.0}}\cdot\phantom{0}50.0$ &
   $50.0\cdot\bm{{\color{black}50.0}}\cdot\phantom{0}50.0$ \\
& \textbf{PriorNet} &  %
$30.7\cdot\bm{{\color{blue}51.7}}\cdot100.0$ & 
$30.7\cdot\bm{{\color{blue}48.2}}\cdot100.0$ & 
$30.7\cdot\bm{{\color{blue}48.6}}\cdot100.0$ & 
$30.7\cdot\bm{{\color{blue}68.6}}\cdot100.0$ & 
$31.4\cdot\bm{{\color{blue}63.7}}\cdot100.0$ \\
 &   \textbf{DDNet} &  %
 $30.7\cdot\bm{{\color{blue}67.1}}\cdot100.0$ &    
 $30.7\cdot\bm{{\color{blue}58.2}}\cdot100.0$ &   
 $30.7\cdot\bm{{\color{blue}51.7}}\cdot100.0$ &  
 $30.7\cdot\bm{{\color{blue}69.8}}\cdot100.0$ & 
 $30.7\cdot\bm{{\color{blue}73.7}}\cdot100.0$ \\
  &  \textbf{EvNet} &  %
  $30.7\cdot\bm{{\color{blue}77.9}}\cdot100.0$ &    
  $30.7\cdot\bm{{\color{blue}85.3}}\cdot100.0$ &   
  $30.7\cdot\bm{{\color{blue}90.5}}\cdot100.0$ &  
  $30.8\cdot\bm{{\color{blue}84.3}}\cdot100.0$ & 
  $34.0\cdot\bm{{\color{black}50.0}}\cdot100.0$ \\

			\midrule
              \multirow{4}{1.2cm}{Smoothed + adv. w. label attacks} &  
  \textbf{PostNet} &  %
  $30.7\cdot\bm{{\color{black}76.7}}\cdot100.0$ & 
  $30.7\cdot\bm{{\color{black}78.7}}\cdot100.0$ & 
  $50.0\cdot\bm{{\color{black}50.0}}\cdot\phantom{0}50.0$ & 
  $50.0\cdot\bm{{\color{black}50.0}}\cdot\phantom{0}50.0$ & 
  $50.0\cdot\bm{{\color{black}50.0}}\cdot\phantom{0}50.0$ \\
 & \textbf{PriorNet} & %
 $30.7\cdot\bm{{\color{blue}63.9}}\cdot100.0$ &   
 $30.7\cdot\bm{{\color{blue}58.4}}\cdot100.0$ &   
 $30.7\cdot\bm{{\color{blue}45.2}}\cdot100.0$ &
 $30.7\cdot\bm{{\color{blue}43.3}}\cdot100.0$ &
 $32.9\cdot\bm{{\color{black}35.5}}\cdot100.0$ \\
\bottomrule
		\end{tabular}}
\end{table*}

\begin{table*}[ht!]
	\centering
	\caption{Attack detection (FGSM label attacks) based on differential entropy. Smoothed DBU models on Segment. Column format: guaranteed lowest performance $\cdot$ empirical performance $\cdot$ guaranteed highest performance (blue: normally/adversarially trained smooth classifier is more robust than the base model)..}
	\label{tab:segment_smooth_attackdetection_fgsm}
	\resizebox{\textwidth}{!}{
		\begin{tabular}{llcccccc}
			\toprule
			& \textbf{Att. Rad.} &   0.1 &  0.2 &  0.5 &  1.0 &  2.0 \\
			\midrule
               \multirow{4}{1.2cm}{Smoothed models} & 
   \textbf{PostNet} &  %
   $30.8\cdot\bm{{\color{black}76.9}}\cdot100.0$ &  
   $30.8\cdot\bm{{\color{black}62.5}}\cdot100.0$ &  
   $30.8\cdot\bm{{\color{black}59.2}}\cdot100.0$ &  
   $30.8\cdot\bm{{\color{black}48.7}}\cdot100.0$ & 
   $49.7\cdot\bm{{\color{black}50.0}}\cdot\phantom{0}50.0$ \\
 & \textbf{PriorNet} &  %
 $30.9\cdot\bm{{\color{black}81.3}}\cdot\phantom{0}99.9$ &    
 $30.8\cdot\bm{{\color{blue}85.0}}\cdot100.0$ &   
 $30.8\cdot\bm{{\color{blue}48.7}}\cdot100.0$ &  
 $30.8\cdot\bm{{\color{black}37.1}}\cdot100.0$ &
 $30.8\cdot\bm{{\color{blue}43.7}}\cdot100.0$ \\
  &  \textbf{DDNet} &  %
  $31.7\cdot\bm{{\color{black}73.8}}\cdot\phantom{0}99.7$ & 
  $30.8\cdot\bm{{\color{black}80.5}}\cdot100.0$ &
  $30.8\cdot\bm{{\color{blue}80.4}}\cdot100.0$ &  
  $30.8\cdot\bm{{\color{blue}72.7}}\cdot100.0$ & 
  $30.8\cdot\bm{{\color{blue}70.6}}\cdot100.0$ \\
   & \textbf{EvNet} &  %
   $30.8\cdot\bm{{\color{black}89.1}}\cdot100.0$ &    
   $30.8\cdot\bm{{\color{blue}89.5}}\cdot100.0$ &    
   $30.8\cdot\bm{{\color{blue}75.3}}\cdot100.0$ &    
   $30.8\cdot\bm{{\color{blue}73.1}}\cdot100.0$ &  
   $30.8\cdot\bm{{\color{blue}83.1}}\cdot100.0$ \\

			\midrule
              \multirow{4}{1.2cm}{Smoothed + adv. w. label attacks} &  
  \textbf{PostNet} &  %
  $30.8\cdot\bm{{\color{black}81.0}}\cdot100.0$ & 
  $30.8\cdot\bm{{\color{black}75.6}}\cdot100.0$ &  
  $30.8\cdot\bm{{\color{black}56.3}}\cdot100.0$ &  
  $30.8\cdot\bm{{\color{black}50.0}}\cdot100.0$ &  
  $50.0\cdot\bm{{\color{black}50.0}}\cdot\phantom{0}50.0$ \\
 & \textbf{PriorNet} &  %
 $31.1\cdot\bm{{\color{black}77.9}}\cdot\phantom{0}99.4$ &  
 $30.8\cdot\bm{{\color{black}76.1}}\cdot100.0$ &  
 $30.8\cdot\bm{{\color{blue}62.4}}\cdot100.0$ &    
 $30.8\cdot\bm{{\color{blue}65.5}}\cdot100.0$ &   
 $30.8\cdot\bm{{\color{blue}53.3}}\cdot100.0$ \\
\bottomrule
		\end{tabular}}
\end{table*}

\begin{table*}[ht!]
	\centering
	\caption{OOD detection based on differential entropy under PGD uncertainty attacks against differential entorpy on ID data and OOD data. Smoothed DBU models on CIFAR10. Column format: guaranteed lowest performance $\cdot$ empirical performance $\cdot$ guaranteed highest performance (blue: normally/adversarially trained smooth classifier is more robust than the base model).}
	\label{tab:cifar10_smooth_ooddetection}
	\resizebox{\textwidth}{!}{
		\begin{tabular}{llcccccc}
			\toprule
			& \textbf{Att. Rad.} & 0.0 &   0.1 &  0.2 &  0.5 &  1.0 &  2.0 \\
			\midrule
			& & \multicolumn{6}{c}{\textbf{ID-Attack}} \\
            
			\midrule
            
			\midrule
            
			\midrule
			\midrule
			& & \multicolumn{6}{c}{\textbf{OOD-Attack}} \\
            
			\midrule
            
			\midrule
            
			\bottomrule
		\end{tabular}}
\end{table*}

\begin{table*}[ht!]
	\centering
	\caption{OOD detection based on differential entropy under PGD uncertainty attacks against differential entropy on ID data and OOD data. Smoothed DBU models  on MNIST. Column format: guaranteed lowest performance $\cdot$ empirical performance $\cdot$ guaranteed highest performance (blue: normally/adversarially trained smooth classifier is more robust than the base model).}
	\label{tab:mnist_smooth_ooddetection}
	\resizebox{\textwidth}{!}{
		\begin{tabular}{llccccccc}
			\toprule
			&\textbf{Att. Rad.} & 0.0 &   0.1 &  0.2 &  0.5 &  1.0 &  2.0 \\
			\midrule
			& & \multicolumn{6}{c}{\textbf{ID-Attack}} \\
               \multirow{4}{1.2cm}{Smoothed models} & 
   \textbf{PostNet} & 
   $59.9\cdot\bm{{\color{black}91.1}}\cdot98.6$ &    
   $61.2\cdot\bm{{\color{blue}97.7}}\cdot99.6$ &   
   $64.8\cdot\bm{{\color{blue}94.7}}\cdot99.7$ & 
   $31.6\cdot\bm{{\color{black}64.9}}\cdot\phantom{0}99.7$ & 
   $30.7\cdot\bm{{\color{black}63.2}}\cdot100.0$ & 
   $30.7\cdot\bm{{\color{blue}70.5}}\cdot100.0$ \\
& \textbf{PriorNet} &     
$99.8\cdot\bm{{\color{blue}99.8}}\cdot99.8$ &    
$99.4\cdot\bm{{\color{blue}99.8}}\cdot99.9$ &  
$98.3\cdot\bm{{\color{blue}99.6}}\cdot99.9$ & 
$48.5\cdot\bm{{\color{black}91.9}}\cdot\phantom{0}99.9$ & 
$31.1\cdot\bm{{\color{black}74.6}}\cdot\phantom{0}99.8$ & 
$30.7\cdot\bm{{\color{black}67.3}}\cdot100.0$ \\
 &   \textbf{DDNet} & 
 $98.5\cdot\bm{{\color{black}98.6}}\cdot98.7$ & 
 $95.0\cdot\bm{{\color{black}97.6}}\cdot98.9$ & 
 $74.7\cdot\bm{{\color{black}92.0}}\cdot98.2$ & 
 $31.4\cdot\bm{{\color{black}52.0}}\cdot\phantom{0}98.5$ &  
 $30.7\cdot\bm{{\color{black}52.0}}\cdot100.0$ & 
 $30.7\cdot\bm{{\color{black}41.1}}\cdot100.0$ \\
  &  \textbf{EvNet} &     
  $85.7\cdot\bm{{\color{blue}87.5}}\cdot89.2$ &   
  $68.9\cdot\bm{{\color{blue}90.4}}\cdot97.7$ &   
  $42.5\cdot\bm{{\color{blue}90.2}}\cdot99.6$ &
  $30.7\cdot\bm{{\color{blue}69.8}}\cdot100.0$ &  
  $30.7\cdot\bm{{\color{black}50.3}}\cdot100.0$ & 
  $30.7\cdot\bm{{\color{black}45.6}}\cdot100.0$ \\

			\midrule
              \multirow{4}{1.2cm}{Smoothed + adv. w. label attacks} &  
  \textbf{PostNet} &  - &     
  $84.3\cdot\bm{{\color{blue}96.2}}\cdot\phantom{0}99.3$ &  
  $50.4\cdot\bm{{\color{black}89.2}}\cdot\phantom{0}99.5$ &  
  $30.9\cdot\bm{{\color{black}46.2}}\cdot\phantom{0}99.4$ & 
  $30.7\cdot\bm{{\color{black}46.9}}\cdot100.0$ &  
  $30.7\cdot\bm{{\color{blue}62.2}}\cdot100.0$ \\
\bottomrule
		\end{tabular}}
\end{table*}

\begin{table*}[ht!]
	\centering
	\caption{OOD detection based on differential entropy under PGD uncertainty attacks against differential entropy on ID data and OOD data. Smoothed DBU models  on Sensorless. Column format: guaranteed lowest performance $\cdot$ empirical performance $\cdot$ guaranteed highest performance (blue: normally/adversarially trained smooth classifier is more robust than the base model).}
	\label{tab:sensorless_smooth_ooddetection}
	\resizebox{\textwidth}{!}{
		\begin{tabular}{llccccccc}
			\toprule
			& \textbf{Att. Rad.} & 0.0 &   0.1 &  0.2 &  0.5 &  1.0 &  2.0 \\
			\midrule
			& & \multicolumn{6}{c}{\textbf{ID-Attack}} \\
               \multirow{4}{1.2cm}{Smoothed models} & 
   \textbf{PostNet} &    
   $49.3\cdot\bm{{\color{blue}90.4}}\cdot99.8$ & 
   $30.7\cdot\bm{{\color{blue}49.2}}\cdot100.0$ & 
   $30.7\cdot\bm{{\color{black}36.0}}\cdot100.0$ &  
   $49.2\cdot\bm{{\color{blue}50.0}}\cdot\phantom{0}74.9$ & 
   $50.0\cdot\bm{{\color{blue}50.0}}\cdot\phantom{0}50.0$ &   
   $50.0\cdot\bm{{\color{blue}50.0}}\cdot\phantom{0}50.0$ \\
 & \textbf{PriorNet} &     
 $31.2\cdot\bm{{\color{blue}39.0}}\cdot66.9$ & 
 $30.7\cdot\bm{{\color{blue}35.5}}\cdot100.0$ &    
 $30.7\cdot\bm{{\color{blue}38.9}}\cdot100.0$ &
 $30.7\cdot\bm{{\color{blue}46.2}}\cdot100.0$ &
 $30.7\cdot\bm{{\color{blue}62.7}}\cdot100.0$ &
 $30.7\cdot\bm{{\color{blue}51.3}}\cdot100.0$ \\
   & \textbf{DDNet} & 
   $31.0\cdot\bm{{\color{black}31.5}}\cdot32.7$ & 
   $30.7\cdot\bm{{\color{blue}30.8}}\cdot100.0$ & 
   $30.7\cdot\bm{{\color{blue}31.8}}\cdot100.0$ &  
   $30.7\cdot\bm{{\color{blue}53.6}}\cdot100.0$ &
   $30.7\cdot\bm{{\color{blue}43.9}}\cdot100.0$ &
   $30.7\cdot\bm{{\color{blue}40.5}}\cdot100.0$ \\
&    \textbf{EvNet} &
$33.6\cdot\bm{{\color{black}55.2}}\cdot91.3$ &
$30.7\cdot\bm{{\color{blue}44.2}}\cdot100.0$ &   
$30.7\cdot\bm{{\color{blue}43.8}}\cdot100.0$ & 
$30.7\cdot\bm{{\color{blue}39.3}}\cdot100.0$ &  
$30.8\cdot\bm{{\color{blue}51.6}}\cdot100.0$ & 
$32.4\cdot\bm{{\color{blue}50.0}}\cdot100.0$ \\

			\midrule
               \multirow{4}{1.2cm}{Smoothed + adv. w. label attacks} &
   \textbf{PostNet} &  - &
   $30.7\cdot\bm{{\color{blue}62.4}}\cdot100.0$ & 
   $30.7\cdot\bm{{\color{blue}39.2}}\cdot100.0$ & 
   $50.0\cdot\bm{{\color{blue}50.0}}\cdot\phantom{0}50.0$ &  
   $50.0\cdot\bm{{\color{blue}50.0}}\cdot\phantom{0}50.0$ & 
   $50.0\cdot\bm{{\color{blue}50.0}}\cdot\phantom{0}50.0$ \\
\bottomrule
		\end{tabular}}
\end{table*}

\begin{table*}[ht!]
	\centering
	\caption{OOD detection based on differential entropy under PGD uncertainty attacks against differential entropy on ID data and OOD data. Smoothed DBU models  on Segment. Column format: guaranteed lowest performance $\cdot$ empirical performance $\cdot$ guaranteed highest performance (blue: normally/adversarially trained smooth classifier is more robust than the base model).}
	\label{tab:segment_smooth_ooddetection}
	\resizebox{\textwidth}{!}{
		%
}
\end{table*}

\begin{table*}[ht!]
	\centering
	\caption{OOD detection based on differential entropy under FGSM uncertainty attacks against differential entropy on ID data and OOD data. Smoothed DBU models  on CIFAR10. Column format: guaranteed lowest performance $\cdot$ empirical performance $\cdot$ guaranteed highest performance (blue: normally/adversarially trained smooth classifier is more robust than the base model).}
	\label{tab:cifar10_smooth_ooddetection_fgsm}
	\resizebox{\textwidth}{!}{
		%
}
\end{table*}

\begin{table*}[ht!]
	\centering
	\caption{OOD detection based on differential entropy under FGSM uncertainty attacks against differential entropy on ID data and OOD data. Smoothed DBU models  on MNIST. Column format: guaranteed lowest performance $\cdot$ empirical performance $\cdot$ guaranteed highest performance (blue: normally/adversarially trained smooth classifier is more robust than the base model).}
	\label{tab:mnist_smooth_ooddetection_fgsm}
	\resizebox{\textwidth}{!}{
		%
}
\end{table*}

\begin{table*}[ht!]
	\centering
	\caption{OOD detection based on differential entropy under FGSM uncertainty attacks against differential entropy on ID data and OOD data. Smoothed DBU models  on Sensorless. Column format: guaranteed lowest performance $\cdot$ empirical performance $\cdot$ guaranteed highest performance (blue: normally/adversarially trained smooth classifier is more robust than the base model)..}
	\label{tab:sensorless_smooth_ooddetection_fgsm}
	\resizebox{\textwidth}{!}{
		%
}
\end{table*}

\begin{table*}[ht!]
	\centering
	\caption{OOD detection based on differential entropy under FGSM uncertainty attacks against differential entropy on ID data and OOD data. Smoothed DBU models  on Segment. Column format: guaranteed lowest performance $\cdot$ empirical performance $\cdot$ guaranteed highest performance (blue: normally/adversarially trained smooth classifier is more robust than the base model).}
	\label{tab:segment_smooth_ooddetection_fgsm}
	\resizebox{\textwidth}{!}{
		%
}
\end{table*}

\clearpage
\subsection{Visualization of differential entropy distributions on ID data and OOD data}

The following Figures visualize the differential entropy distribution for ID data and OOD data for all models with standard training. We used label attacks and uncertainty attacks for CIFAR10 and MNIST. Thus, they show how well the DBU models separate on clean and perturbed ID data and OOD data. 

\begin{figure*}[ht!]
    \centering
        \begin{subfigure}[t]{1.0\textwidth}
        \centering
        \includegraphics[width=0.99 \textwidth]{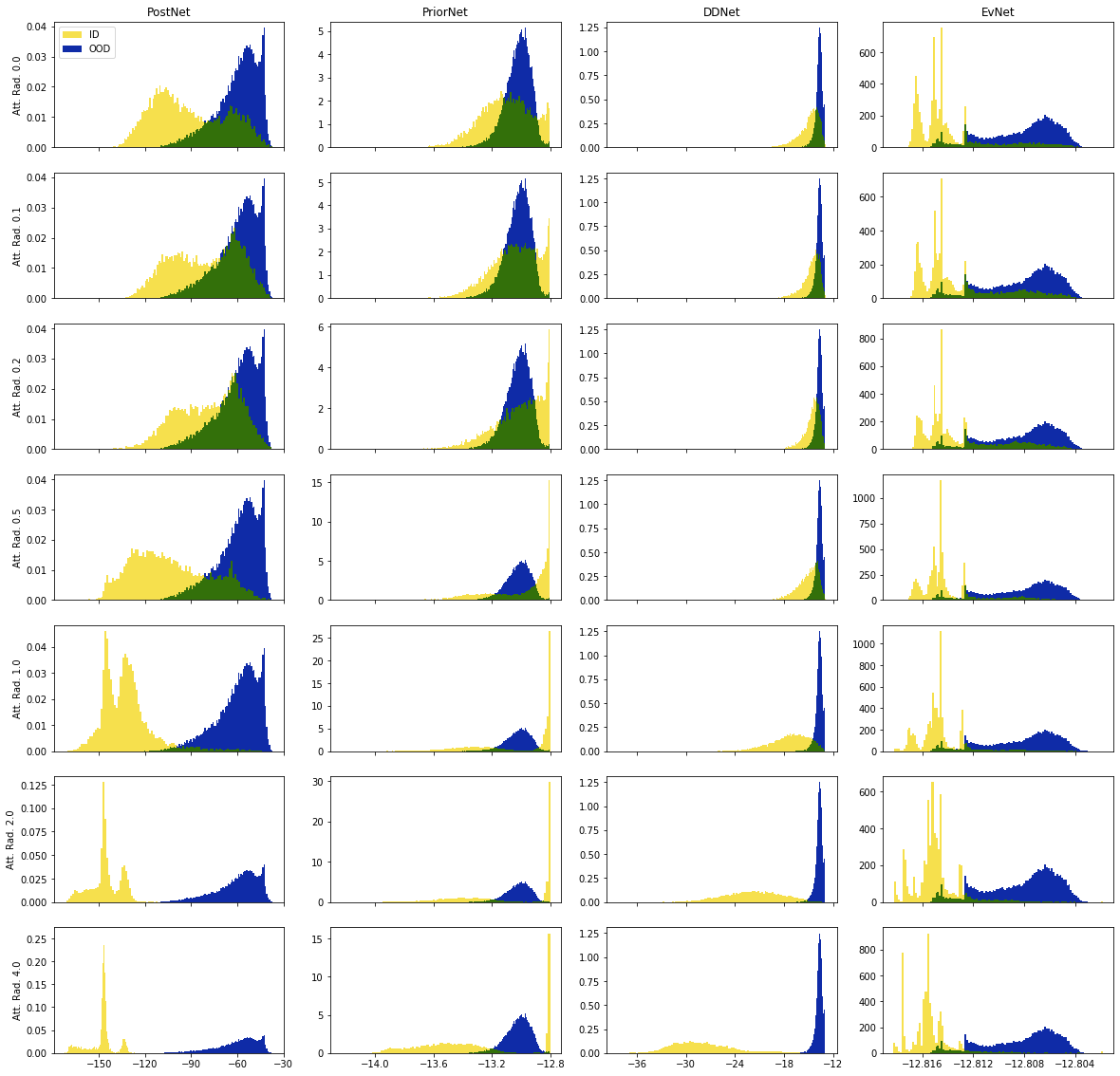}
    \end{subfigure}%
    \caption{Visualization of the differential entropy distribution of ID data (CIFAR10) and OOD data (SVHN) under label attack. The first row corresponds to no attack. The other rows correspond do increasingly stronger attack strength.}
    \label{fig:attaked_samples_idood_label_attacks_2}
	\vspace{-.5cm}
\end{figure*}

\begin{figure*}[ht!]
    \centering
        \begin{subfigure}[t]{1.0\textwidth}
        \centering
        \includegraphics[width=0.99 \textwidth]{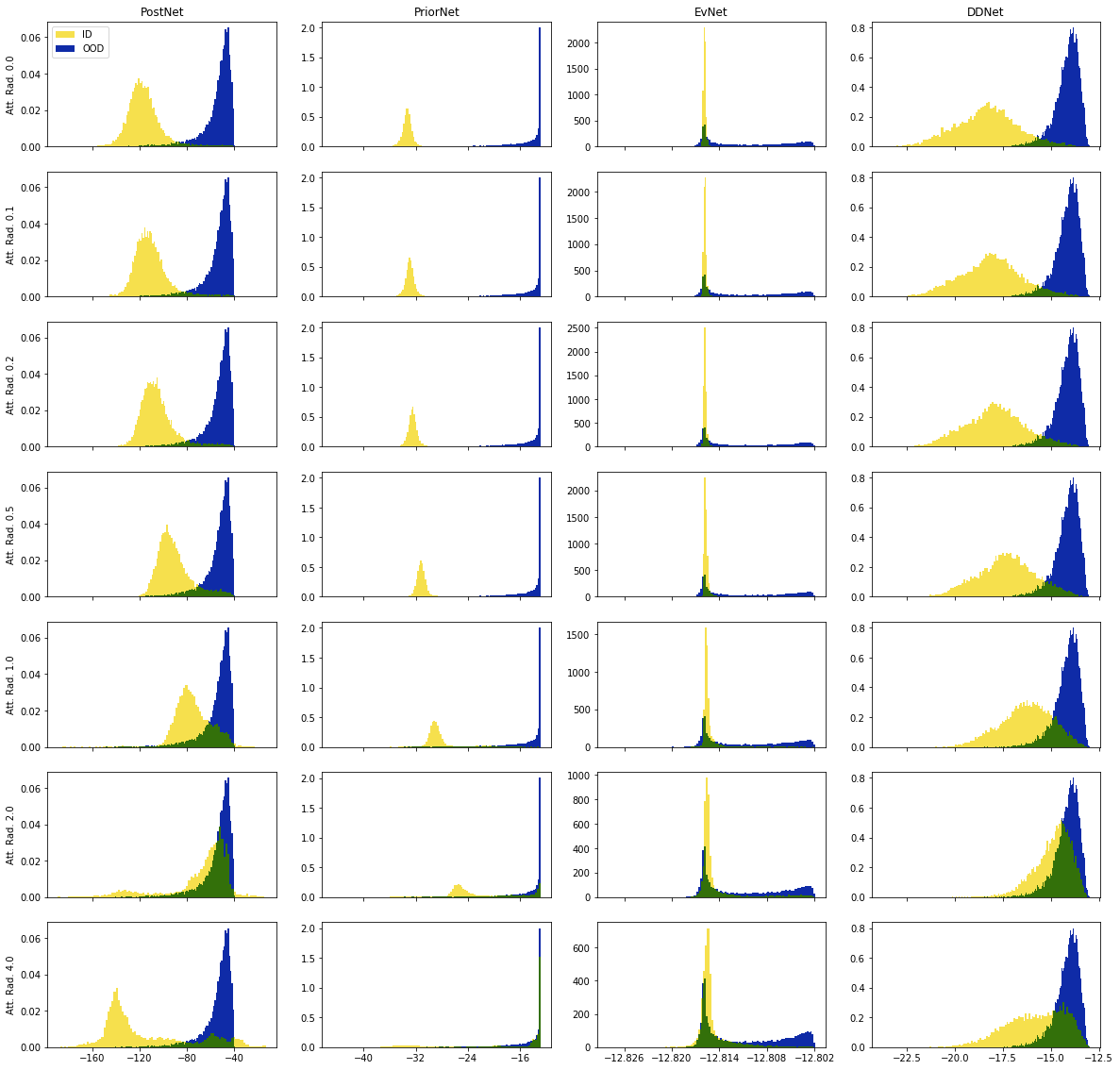}
    \end{subfigure}%
    \caption{Visualization of the differential entropy distribution of ID data (MNIST) and OOD data (KMNIST) under label attack. The first row corresponds to no attack. The other rows correspond do increasingly stronger attack strength.}
    \label{fig:attaked_samples_idood_label_attacks_3}
	\vspace{-.5cm}
\end{figure*}

Figures~\ref{fig:attaked_samples_idood_label_attacks_2} and \ref{fig:attaked_samples_idood_label_attacks_3} visualizes the differential entropy distribution of ID data and OOD data under label attacks. On CIFAR10, \PriorNet and \DDNet can barely distinguish between clean ID and OOD data. We observe a better ID/OOD distinction for \PostNet and \EvNet for clean data. However, we do not observe for any model an increase of the uncertainty estimates on label attacked data. Even worse, \PostNet, \PriorNet and \DDNet seem to assign higher confidence on class label attacks. On MNIST, models show a slightly better behavior. They are capable to assign a higher uncertainty to label attacks up to some attack radius.

\begin{figure*}[ht!]
    \centering
        \begin{subfigure}[t]{1.0\textwidth}
        \centering
        \includegraphics[width=0.99 \textwidth]{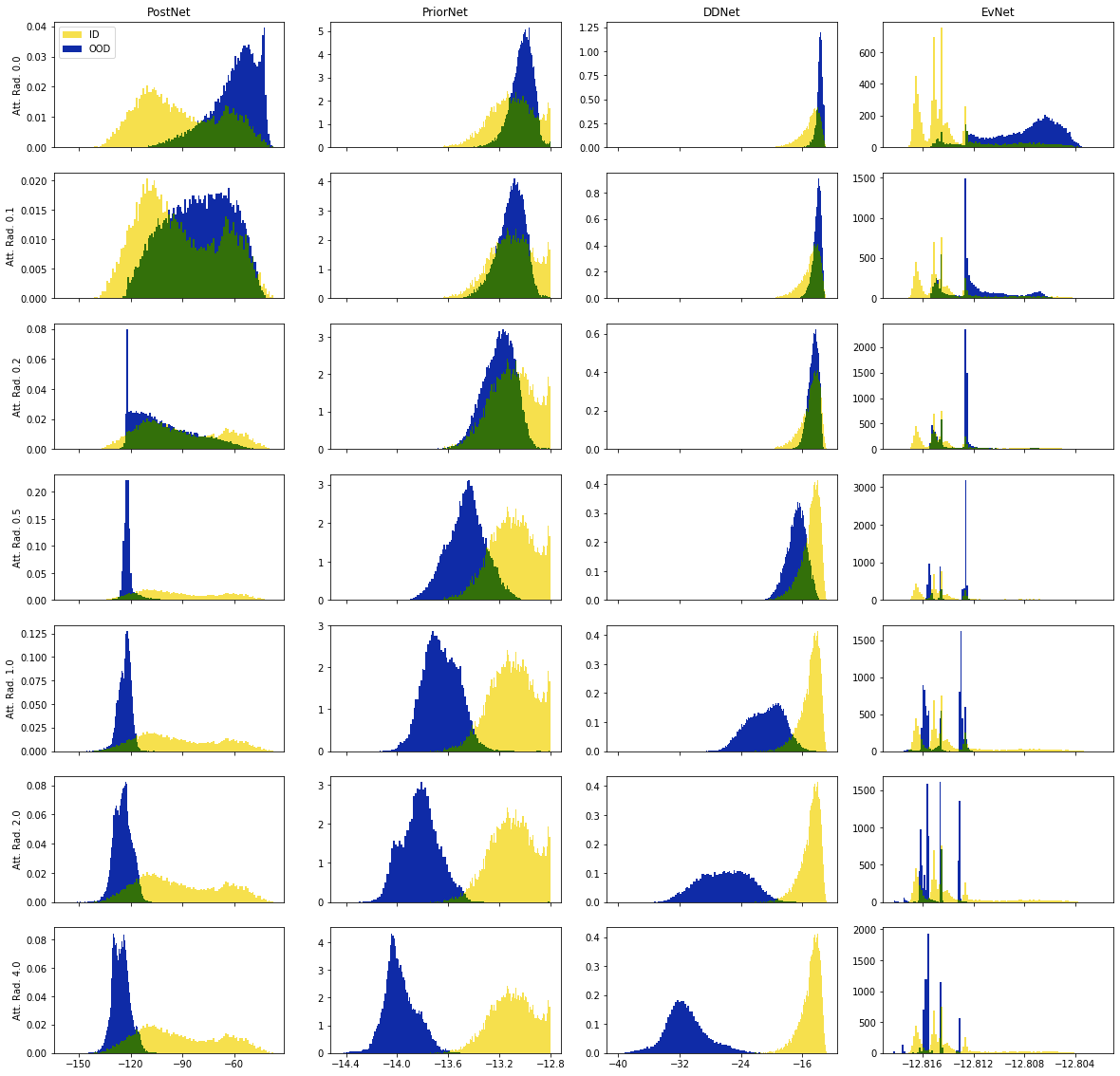}
    \end{subfigure}%
    \caption{Visualization of the differential entropy distribution of ID data (CIFAR10) and OOD data (SVHN) under OOD uncertainty attack. The first row corresponds to no attack. The other rows correspond do increasingly stronger attack strength.}
    \label{fig:attaked_samples_idood_2}
	\vspace{-.5cm}
\end{figure*}

\begin{figure*}[ht!]
    \centering
        \begin{subfigure}[t]{1.0\textwidth}
        \centering
        \includegraphics[width=0.99 \textwidth]{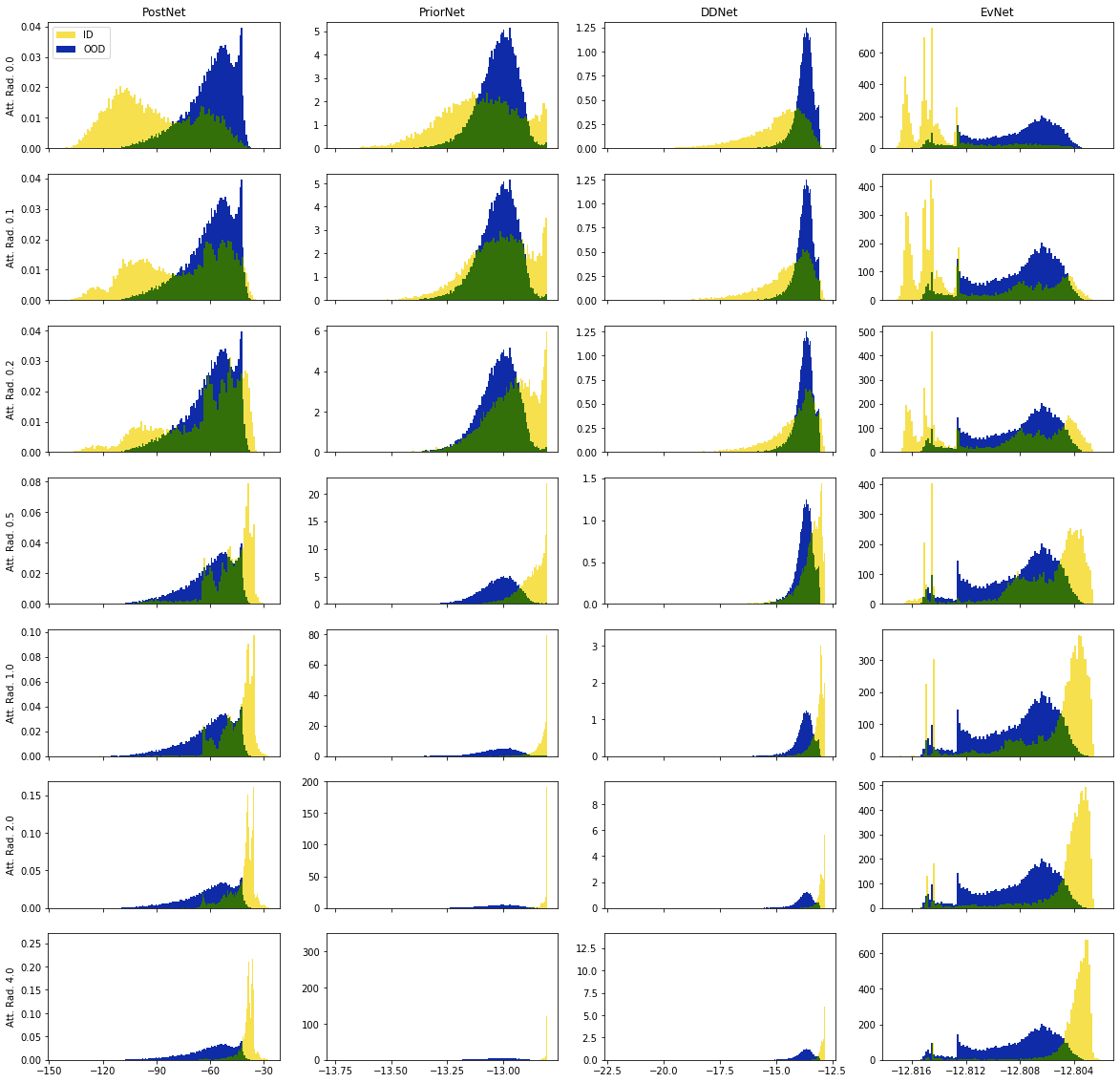}
    \end{subfigure}%
    \caption{Visualization of the differential entropy distribution of ID data (CIFAR10) and OOD data (SVHN) under ID uncertainty attack. The first row corresponds to no attack. The other rows correspond do increasingly stronger attack strength.}
    \label{fig:attaked_samples_idood_3}
	\vspace{-.5cm}
\end{figure*}

\begin{figure*}[ht!]
    \centering
        \begin{subfigure}[t]{1.0\textwidth}
        \centering
        \includegraphics[width=0.99 \textwidth]{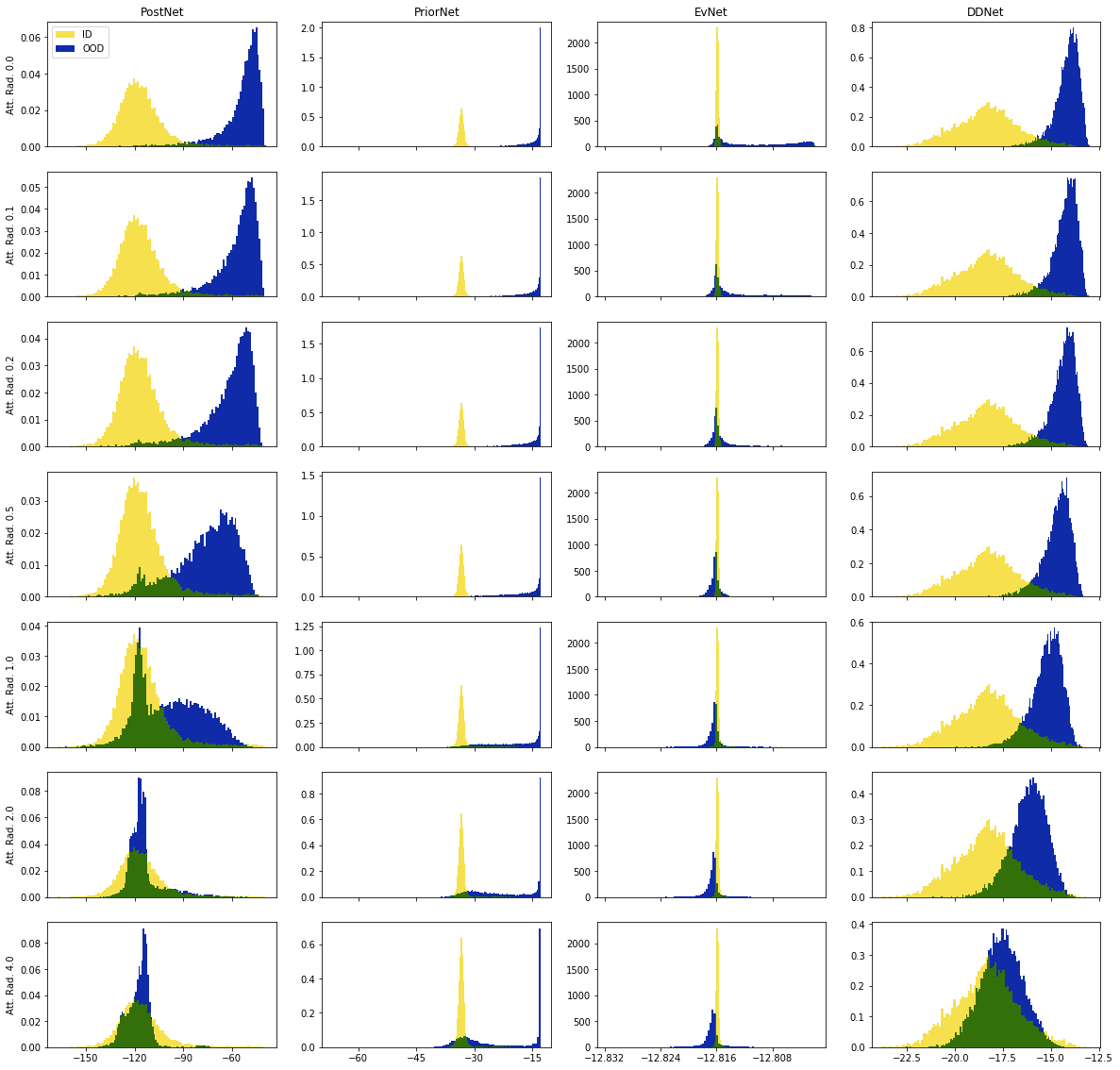}
    \end{subfigure}%
    \caption{Visualization of the differential entropy distribution of ID data (MNIST) and OOD data (KMNIST) under OOD uncertainty attack. The first row corresponds to no attack. The other rows correspond do increasingly stronger attack strength.}
    \label{fig:attaked_samples_idood_mnist}
	\vspace{-.5cm}
\end{figure*}

\begin{figure*}[ht!]
    \centering
        \begin{subfigure}[t]{1.0\textwidth}
        \centering
        \includegraphics[width=0.99 \textwidth]{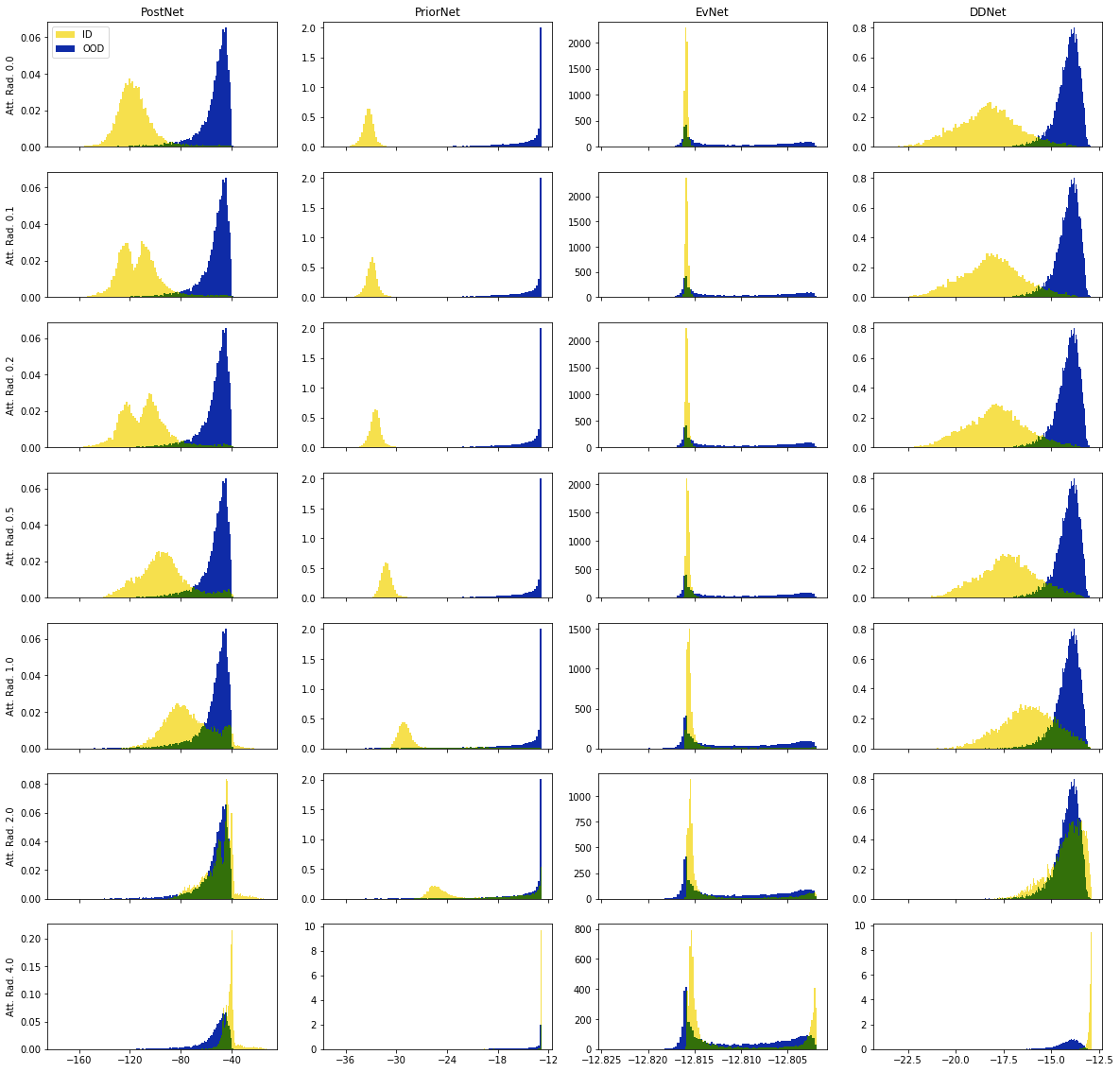}
    \end{subfigure}%
    \caption{Visualization of the differential entropy distribution of ID data (MNIST) and OOD data (KMNIST) under ID uncertainty attack. The first row corresponds to no attack. The other rows correspond do increasingly stronger attack strength.}
    \label{fig:attaked_samples_idood_mnist_2}
	\vspace{-.5cm}
\end{figure*}

Figures~\ref{fig:attaked_samples_idood_2}, \ref{fig:attaked_samples_idood_3}, \ref{fig:attaked_samples_idood_mnist} and \ref{fig:attaked_samples_idood_mnist_2} visualizes the differential entropy distribution of ID data and OOD data under uncertainty attacks. For both CIFAR10 and MNIST data sets, we observed that uncertainty estimations of all models can be manipulated. That is, OOD uncertainty attacks can shift the OOD uncertainty distribution to more certain predictions, and ID uncertainty attacks can shift the ID uncertainty distribution to less certain predictions.

\newpage

\newpage

\end{document}